\documentclass{article}

     \PassOptionsToPackage{numbers, compress}{natbib}
\usepackage[preprint]{neurips_2026}

\usepackage[utf8]{inputenc} 
\usepackage[T1]{fontenc}    
\usepackage{hyperref}       
\usepackage{url}            
\usepackage{booktabs}       
\usepackage{amsfonts}       
\usepackage{nicefrac}       
\usepackage{microtype}      
\usepackage{xcolor}         

\usepackage{float}  
\usepackage{graphicx}
\usepackage{booktabs} 
\usepackage{bm}

\usepackage{wrapfig}
\usepackage{booktabs}

\usepackage{color}
\usepackage{amsthm}
\usepackage{cite}
\usepackage{enumerate}
\usepackage{bm}
\usepackage{multirow}
\usepackage{mathrsfs}
\usepackage{algorithm}
\usepackage[noend]{algpseudocode}
\usepackage{caption}
\usepackage{multirow}
\usepackage{wrapfig}
\usepackage{tikz}
\usepackage{nicefrac,xfrac}
\usepackage{amsmath,amssymb}
\usepackage{amsfonts}
\usepackage{subcaption}
\usepackage{todonotes}
\usepackage{mathtools}

\usepackage{amssymb}
\usepackage{pifont}
\newcommand{\xmark}{\ding{55}}%

\title{\textit{MedGym}: A Unified Continuous-Time Benchmark for Dynamic Medical Treatment Reinforcement Learning
}

\author{
\textbf{
Yuepeng Wang$^{1*}$ \quad
Ken Kawano$^{2}$\thanks{Y. Wang and K. Kawano contributed equally to this work.} \quad
Yoshihiko Fujisawa$^{2}$ \quad
Yongqi Zhou$^{3}$
}\\
\textbf{
Akifumi Wachi$^{4}$ \quad
Mehrshad Sadria$^{5}$ \quad
Lei Zhou$^{2}$ \quad
Richard Weiss$^{3}$ 
}\\
\textbf{
 \quad
Katsuki Fujisawa$^{2}$ \quad
Ying Chen$^{3}$ \quad
Xin Liu$^{6}$
}\\
\textbf{
Kyoung\mbox{-}Sook Kim$^{6}$ \quad
Xiao Hu$^{7}$ \quad
Sebastien Gros$^{8}$\quad
Xun Shen$^{1}$
}\\
$^{1}$Tokyo University of Agriculture and Technology, 
$^{2}$Institute of Science Tokyo\\
$^{3}$National University of Singapore, 
$^{4}$LY Corporation,
$^{5}$University of Toronto, \\
$^{6}$National Institute of Advanced Industrial Science and Technology (AIST), \\
$^{7}$Emory University,
$^{8}$Norwegian University of Science and Technology,\\
\texttt{shen@go.tuat.ac.jp}
}

\begin{document}

\maketitle

\begin{abstract}

Medical treatment recommendation poses several challenges to reinforcement learning (RL): patient physiology evolves in continuous time, measurements and interventions are performed at irregular intervals, and treatment effects vary substantially across individuals. 
Existing RL formulations and simulated environments, however, are based on discrete-time MDPs with fixed decision intervals. 
Thus, it remains difficult to evaluate whether RL methods can handle time-interval-dependent disease progression, personalized treatment response, and safety between consecutive measurement points. To address this gap, we introduce \textit{MedGym}, a benchmark environment for dynamic treatment recommendation. 
\textit{MedGym} models longitudinal patient evolution in a continuous-time framework and constructs a configurable medical RL benchmark from clinical data by using Physics-Informed Neural Networks. 
The resulting benchmark enables direct comparison between discrete-time and continuous-time methods under irregular treatment timing and patient-specific dynamics. Furthermore, \textit{MedGym} supports evaluation from clinically important perspectives, such as personalization and trajectory-level safety. 
By providing a standardized and configurable benchmark for continuous-time dynamic treatment, \textit{MedGym} enables more realistic and informative evaluation of medical RL methods.

\end{abstract}

\section{Introduction}
\label{sec:introduction}

Medical treatment recommendation can naturally be formulated as a reinforcement learning (RL) problem, as treatment decisions must be dynamically adapted to a patient's clinical state, including vital signs, laboratory test results, symptoms, comorbidities, and prior treatment responses \citep{Chakraborty2014DTR,  Yu2021RLHealthcareSurvey}.
Compared with standard RL benchmark settings, RL for medical treatment recommendation has two distinctive properties.
First, a patient's clinical state evolves continuously, while observations are collected and treatments are updated at irregular, decision-dependent times. 
In contrast, standard RL benchmarks model the environment as a discrete-time Markov decision process (MDP) with uniformly spaced decision steps.
Second, even under similar observed conditions and identical interventions, different patients may exhibit distinct latent clinical-state trajectories \citep{Kravitz2004HTE}. In contrast, standard RL benchmarks typically learn a single population-level policy, without explicitly accounting for patient-specific latent dynamics.

\begin{table}[htbp]
\centering
\scriptsize
\caption{
Comparison of \textit{MedGym} with representative medical RL benchmarks.
\textit{MedGym} jointly exposes continuous-time patient evolution,
adaptive intervention timing, personalization at multiple levels,
online and offline policy learning, and trajectory-level safety under
a common evaluation framework.
(\checkmark~explicitly supported as a benchmark dimension,
\xmark~not supported,
$\triangle$~partially supported or a related property is modeled but
not evaluated as a controlled benchmark dimension.)
}
\label{tab:benchmark_comparison}

\renewcommand{\arraystretch}{1.15}
\setlength{\tabcolsep}{3.5pt}

\resizebox{\linewidth}{!}{%
\begin{tabular}{lcccccc}
\toprule
\textbf{Benchmark}
&
\shortstack{\textbf{RL for}\\\textbf{Treatment}}
&
\shortstack{\textbf{Continuous-Time}\\\textbf{Dynamics}}
&
\shortstack{\textbf{Adaptive}\\\textbf{Timing}}
&
\shortstack{\textbf{Population / Cluster /}\\\textbf{Individual Evaluation}}
&
\shortstack{\textbf{Online + Offline}\\\textbf{Policy Learning}}
&
\shortstack{\textbf{Trajectory-Level}\\\textbf{Safety}}
\\
\midrule

MedAgentGym \citep{Xu_ICLR2026}
& \xmark
& \xmark
& \xmark
& \xmark
& \xmark
& \xmark
\\

EpiCare \citep{EpiCare_NeurIPS2024}
& \checkmark
& \xmark
& \xmark
& $\triangle$
& $\triangle$
& $\triangle$
\\

ICU-Sepsis \citep{Choudhary2024ICUSepsis}
& \checkmark
& \xmark
& \xmark
& $\triangle$
& \xmark
& \xmark
\\

DTR-Bench \citep{Luo2024DTRBench}
& \checkmark
& \xmark
& \xmark
& $\triangle$
& \xmark
& \xmark
\\

\textbf{MedGym (ours)}
& \checkmark
& \checkmark
& \checkmark
& \checkmark
& \checkmark
& \checkmark
\\

\bottomrule
\end{tabular}%
}
\end{table}

Despite these characteristics, existing medical RL benchmarks largely simplify dynamic treatment through discrete-time formulations with fixed or predefined decision intervals and primarily evaluate population-level policies. 
MedAgentGym \citep{Xu_ICLR2026} focuses on biomedical coding and reasoning rather than treatment-policy learning, whereas treatment-oriented benchmarks, such as EpiCare \citep{EpiCare_NeurIPS2024}, ICU-Sepsis \citep{Choudhary2024ICUSepsis}, and DTR-Bench \citep{Luo2024DTRBench}, retain discrete-time formulations. 
To the best of our knowledge, no existing benchmark jointly supports continuous-time patient evolution, irregular intervention timing, and explicit comparison between population-level and individualized treatment policies. Table~\ref{tab:benchmark_comparison} summarizes this gap. 
Consequently, existing benchmarks cannot reliably determine whether an RL method addresses the defining challenges of dynamic treatment or merely performs well under a simplified discrete-time approximation \citep{Gottesman2018ObservationalHealthRL, Gottesman2019RLHealthcareGuidelines}.


To address this benchmark gap, we introduce \textit{MedGym}, a unified evaluation platform for dynamic medical treatment RL with two complementary layers. 
The first is a configurable environment-construction layer: a physiology-inspired ODE is combined with a learned neural residual, and a physics-informed neural network (PINN) approximates the resulting solution as a function of the current state, treatment action, and elapsed time, while clinical trajectories are used to construct patient-, cluster-, and population-level dynamics models.
The second is a standardized evaluation layer that uses this environment family to make clinically relevant benchmark dimensions directly comparable, such as fixed versus adaptive intervention timing, population versus increasingly personalized policies, and treatment effectiveness versus trajectory-level safety. 
Thus, the continuous-time patient simulator is an enabling component of \textit{MedGym}, while the benchmark novelty lies in the controlled evaluation capabilities supported by the complete platform.

Our contributions are threefold: (1) We develop a modular pipeline that converts irregular clinical trajectories from different patients into individualized continuous-time treatment simulators. (2) We establish a unified protocol for comparing fixed-interval and time-adaptive methods, population-level, cluster-level, and patient-specific (individual) policies, and offline and online RL. Population- and cluster-level policies are evaluated on held-out patient simulators, while all learned policies are assessed through matched closed-loop rollouts. (3) We instantiate \textit{MedGym} in two clinically distinct treatment settings: sepsis using MIMIC-III, with six physiological state variables, three continuous treatment controls, a SOFA-based objective, and lactate-based trajectory safety; and acute hypotension using the Synthetic Acute Hypotension Dataset, with a MAP-centered treatment objective and distinct physiological safety constraints. Experiments show that intervention timing and the degree of personalization materially affect treatment performance across these settings, demonstrating the importance of evaluating these dimensions explicitly in dynamic treatment recommendation.
The benchmark implementation, including the environment-construction pipeline, the policy-training and evaluation protocols, and scripts to reproduce all reported results, is released at \url{https://github.com/kawanoKen/Medgym}.

\section{Dynamic Medical Treatment and Prior Works}
\label{sec:prior_works}

Let $\bm{\mathrm{x}}\in\mathcal{X}\subseteq\mathbb{R}^{d_{\mathsf{s}}}$ and $\bm{\mathrm{u}}\in\mathcal{U}\subseteq\mathbb{R}^{d_{\mathsf{a}}}$ denote the patient state and treatment action, respectively, where both $\mathcal{X}$ and $\mathcal{U}$ are continuous spaces. 
The patient state evolves in continuous time according to $\bm{\mathrm{x}}_{t}=\bm{\mathrm{x}}_0 + \int_{0}^{t} f_{\xi}(\bm{\mathrm{x}}_{\tau},\bm{\mathrm{u}}_{\tau},\bm{\mathrm{w}}_{\tau})\mathsf{d}\tau.$
Here, $\bm{\mathrm{w}}_{t}\in\mathcal{W}$ denotes exogenous uncertainty, accounting for unpredictable factors such as latent physiological variation, heterogeneous treatment sensitivity, and clinical noise. 
Moreover, $\xi$ characterizes patient-specific features, so that the state transition model may differ across individuals. 
To assess treatment quality, a bounded reward function $b:\mathcal{X}\times\mathcal{U}\rightarrow\mathbb{R}$ is defined from clinical health indicators, measuring the utility of treatment decisions over a finite and clinically meaningful range. 
The goal of dynamic medical treatment is to find a feedback policy $\pi:\mathcal{X}\rightarrow\mathcal{U}$ that maximizes the cumulative reward $b(\bm{\mathrm{x}}_{t},\bm{\mathrm{u}}_{t})$ over a fixed horizon $\mathcal{T}=[0,T]$, that is,
\begin{equation}
    \label{eq:problem_original}
\tag{$\mathcal{P}_{\xi}$}
    \max_{\pi\in\Pi}\ \mathbb{E}\left[\int_{t\in\mathcal{T}} b(\bm{\mathrm{x}}_{t},\pi(\bm{\mathrm{x}}_{t})) \mathsf{d}t\right]\quad \mathsf{s.t.}\quad \bm{\mathrm{x}}_{t}=\bm{\mathrm{x}}_0 + \int_{0}^{t} f_{\xi}(\bm{\mathrm{x}}_{\tau},\bm{\mathrm{u}}_{\tau},\bm{\mathrm{w}}_{\tau})\mathsf{d}\tau.
\end{equation}
Here, $\Pi$ denotes the policy class. 
Patient states are observed, and treatment actions are updated at discrete decision times, which are irregular and depend on clinical judgment \citep{Chakraborty2014DTR, Shortreed2011SequentialClinicalRL, Yu2021RLHealthcareSurvey}. 
Dynamic medical treatment exhibits three key characteristics: (a) Patient states' evolution admits a \textit{continuous-time Markov Decision Process} (CTMDP). (b) Clinical measurements and treatment actions are \textit{generated at irregular time intervals}, as their timing depends on clinicians' judgments of the current patient condition rather than on a fixed decision schedule. (c) Patient responses are heterogeneous, so each individual may follow a distinct dynamics and thus \textit{requires personalized prediction and treatment} \citep{Kravitz2004HTE}.

Recent reinforcement learning approaches to dynamic treatment commonly cast the problem as a discrete-time Markov Decision Process (MDP) with fixed or pre-specified decision intervals \citep{Komorowski,Shen_NeurIPS2025,Peine2021VentilationRL}. 
Accordingly, existing simulated benchmarks are also built on the discrete-time MDP formulation \citep{EpiCare_NeurIPS2024}. 
In this setting, $\bm{\mathrm{s}}_k=\bm{\mathrm{x}}_{t_k}$ and $\bm{\mathrm{a}}_k=\bm{\mathrm{u}}_{t_k}$ denote the measured state and implemented treatment at the $k$-th decision time, respectively. 
The subsequent state $\bm{\mathrm{s}}_{k+1}=\bm{\mathrm{x}}_{t_{k+1}}$ ($t_{k+1}>t_k$) is assumed to follow the transition dynamics 
$\mathcal{T}(\bm{\mathrm{s}}_{k+1} \mid \bm{\mathrm{s}}_k, \bm{\mathrm{a}}_k)$, which characterizes the distribution of the next state $\bm{\mathrm{s}}_{{k+1}}$ conditioned on the current state $\bm{\mathrm{s}}_k$ and action $\bm{\mathrm{a}}_k$. 
The reward function $r: \mathcal{X} \times \mathcal{U} \rightarrow [0, r_{\mathrm{max}}]$ is defined by $r(\bm{\mathrm{s}}_k,\bm{\mathrm{a}}_k):=b(\bm{\mathrm{s}}_k,\bm{\mathrm{a}}_k)$, that is, it coincides with the instantaneous reward evaluated at time $t_k$.
Then, dynamic medical treatment is represented by the discrete-time MDP $\mathcal{M}:=\langle \mathcal{X},\mathcal{U},\mathcal{T},\gamma, \rho_0 \rangle,$ 
where $\gamma\in(0,1]$ is the discount factor, and $\rho_0$ denotes the probability density of the initial patient state $\bm{\mathrm{s}}_0$, typically reflecting the diversity of patient conditions at ICU admission or treatment initiation. 
A treatment policy is modeled as a stochastic mapping from each state to a probability density over admissible actions. 
Let $\tau := \{\bm{\mathrm{s}}_0, \bm{\mathrm{a}}_0, \dots, \bm{\mathrm{s}}_k, \bm{\mathrm{a}}_k, \dots\}$ denote a trajectory generated by a policy $\pi \in \Pi$. 
The value function under policy $\pi$ and transition model $\mathcal{T}$ is defined as $V^{\pi}_{r, \mathcal{T}}(\bm{\mathrm{s}}) = \mathbb{E}_{\pi}[\sum_{k=0}^{\infty} \gamma^{k} r(\bm{\mathrm{s}}_k, \bm{\mathrm{a}}_k) \mid \bm{\mathrm{s}}_0 = \bm{\mathrm{s}}]$. 
The corresponding population-level performance is given by $V^{\pi}_{r, \mathcal{T}}(\rho_0) := \mathbb{E}_{\bm{\mathrm{s}} \sim \rho_0}[V^{\pi}_{r, \mathcal{T}}(\bm{\mathrm{s}})]$.
The resulting RL problem is formulated as
\begin{equation}
\label{eq:prob_DTRL} \tag{$\mathsf{DTRL}$}
\begin{aligned}
\max_{\pi\in\Pi} V^{\pi}_{r,\mathcal{T}}(\rho_0) 
\end{aligned}
\end{equation}
Existing simulated environments for medical RL are built in discrete time. 
For example, EpiCare in \citep{EpiCare_NeurIPS2024} simulates longitudinal treatment through timestep-wise state transitions, where the patient evolves from one decision step to the next until remission or termination.

\noindent
\textbf{Gaps.} 
By comparing the above two formulations, two important gaps emerge. 
First, the continuous-time treatment process in \ref{eq:problem_original} is reduced to the discrete-time formulation in \ref{eq:prob_DTRL}. 
Because clinical decision times $t_k$ and $t_{k+1}$ are generally irregular and condition-dependent \citep{Chakraborty2014DTR,Murphy2005AdaptiveTreatmentStrategies,Kidwell2023SMART}, the evolution from $\bm{\mathrm{s}}_k=\bm{\mathrm{x}}_{t_k}$ to $\bm{\mathrm{s}}_{k+1}=\bm{\mathrm{x}}_{t_{k+1}}$ should depend on $t_{k+1}-t_k$. 
However, the transition model $\mathcal{T}(\bm{\mathrm{s}}_{k+1} \mid \bm{\mathrm{s}}_k, \bm{\mathrm{a}}_k)$ omits this dependence and therefore cannot distinguish transitions induced by the same state--action pair over different durations, limiting its fidelity under sparse or unevenly spaced observations \citep{Gottesman2018ObservationalHealthRL,Gottesman2019RLHealthcareGuidelines}. 
Second, \ref{eq:prob_DTRL} insufficiently represents patient heterogeneity. The patient-specific parameter $\xi$ induces dynamics $f_{\xi}$ under which the same intervention may produce different responses across patients \citep{Kravitz2004HTE}. 
Yet existing benchmarks generally lack individual patient models, preventing patient-level policy evaluation, cross-patient comparison, and assessment of whether population-average optimality translates into optimality for each patient. 
For a patient model characterized by $\xi$, the individualized optimal policy can be denoted by $\pi_{\xi}^{\star}$. 
In contrast, solving \ref{eq:prob_DTRL} yields a shared policy $\pi(\bm{\mathrm{s}})$ that assigns the same treatment to various patients with identical observed states, regardless of $\xi$. 
Overall, evaluation settings that jointly capture continuous-time disease progression, irregular decision timing, and patient-specific treatment dynamics remain lacking.

\section{Environment}
\label{sec:environment}

\begin{figure}[t]
    \includegraphics[width=1\textwidth]{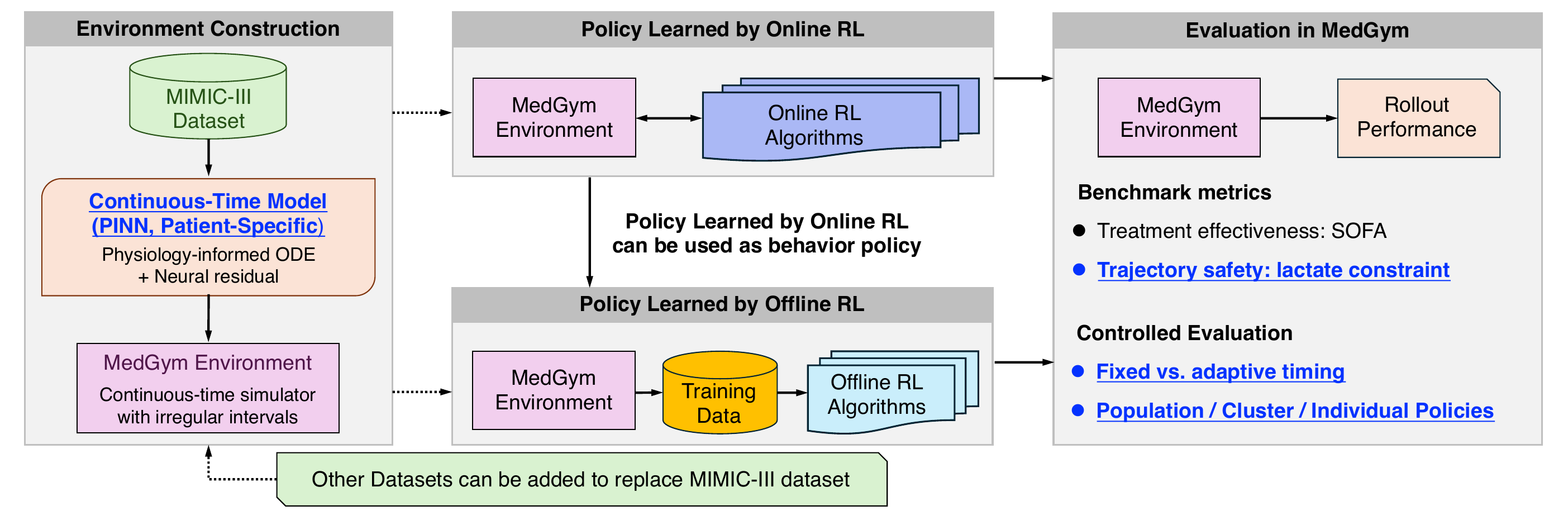}
    \caption{Benchmarking pipeline of \textit{MedGym}. Clinical trajectories are used to construct a PINN-based continuous-time treatment environment, which supports both online RL through environment interaction and offline RL through generated training data. Learned policies are evaluated in \textit{MedGym} under controlled comparisons of fixed versus adaptive timing and population-, cluster-, and individual-level policies, using SOFA-based treatment effectiveness and lactate-based trajectory safety metrics.}
   \label{fig:benchmarking_pipeline}
\end{figure}

\noindent
\textbf{Benchmarking pipeline.}
Figure~\ref{fig:benchmarking_pipeline} summarizes how to use \textit{MedGym} as a benchmarking platform. 
Clinical trajectories, instantiated here by MIMIC-III, are first converted into the continuous-time simulator described later; the same construction pipeline can be re-instantiated with other clinical datasets. 
The environment supports two policy-learning routes. 
Online RL methods learn through direct interaction with \textit{MedGym}, whereas offline RL methods learn from fixed trajectory datasets generated from the environment; when desired, a policy obtained by online RL can be used as the behavior policy for constructing such offline datasets. 
The resulting policies are then assessed through closed-loop rollouts in \textit{MedGym} under a common evaluation interface.
Importantly, the evaluation stage is organized around the benchmark dimensions rather than a single aggregate performance score. 
Policies can be compared under fixed or adaptive intervention timing and across population-, cluster-, and patient-level training settings, while treatment effectiveness and trajectory safety are reported separately through SOFA- and lactate-based metrics, respectively. 
This separation of environment construction, policy learning, and controlled rollout evaluation allows the same benchmark instance to support consistent comparisons across different RL paradigms and treatment-policy settings.

\begin{figure}[t]
    \includegraphics[width=1\textwidth]{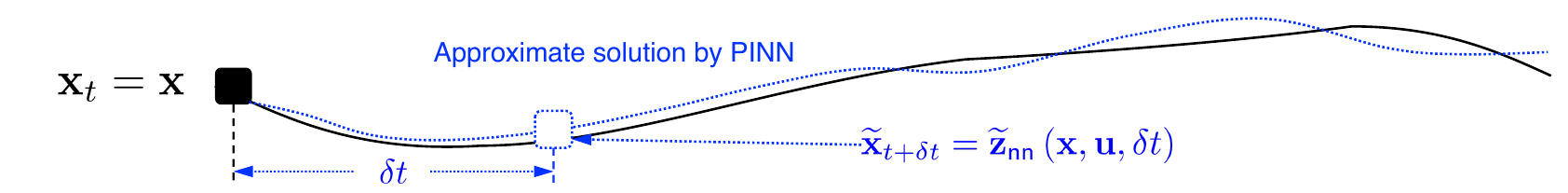}
    \caption{Schematic illustration of the PINN-based state transition in \textit{MedGym}. Starting from the current state $\bm{\mathrm{x}}_t=\bm{\mathrm{x}}$, the patient state evolves over an elapsed interval $\delta t$ under treatment action $\bm{\mathrm{u}}$, and the future state at $t+\delta t$ is approximated by $\widetilde{\bm{\mathrm{x}}}_{t+\delta t}=\widetilde{\bm{\mathrm{z}}}_{\mathsf{nn}}(\bm{\mathrm{x}},\bm{\mathrm{u}},\delta t)$ from \eqref{eq:pinn_ckd_approximate_solution_main}.}
   \label{fig:CTMDP_framework}
\end{figure}

\textbf{PINN-based framework.}
\textit{MedGym} approximates the patient CTMDP using a Physics-Informed Neural Networks (PINN)-based framework as illustrated in Figure~\ref{fig:CTMDP_framework}. 
The patient evolution is modeled as a stochastic Ordinary Differential Equation (ODE) together with treatment interventions. 
We use a PINN to yield the following approximate solution:
\begin{equation}
    \label{eq:pinn_ckd_approximate_solution_main}
    \widetilde{\bm{\mathrm{z}}}_{\mathsf{nn}}\left(\bm{\mathrm{x}},\bm{\mathrm{u}},\delta t\right)=\bm{\mathrm{F}}^{\mathsf{nn}}_{\xi}\left(\bm{\mathrm{x}},\bm{\mathrm{u}},\delta t\right) + \bm{\mathrm{W}}_{\mathsf{c}},\ \bm{\mathrm{W}}_{\mathsf{c}}\sim \widetilde{p}\left(\cdot|\bm{\mathrm{x}},\delta t\right).
\end{equation}
For any given current state $\bm{\mathrm{x}}_t=\bm{\mathrm{x}}$ at $t$, the estimated state $\widetilde{\bm{\mathrm{x}}}_{t+\delta t}$ is given by $\widetilde{\bm{\mathrm{x}}}_{t+\delta t}=\widetilde{\bm{\mathrm{z}}}_{\mathsf{nn}}\left(\bm{\mathrm{x}},\bm{\mathrm{u}},\delta t\right)$. 
Here, $\xi$ characterizes patient-specific factors, so that the deterministic component $\bm{\mathrm{F}}^{\mathsf{nn}}_{\xi}\left(\bm{\mathrm{x}},\bm{\mathrm{u}},\delta t\right)$ captures individualized disease progression and treatment effects. 
This approximation consists of a deterministic component $\bm{\mathrm{F}}^{\mathsf{nn}}_{\xi}\left(\bm{\mathrm{x}},\bm{\mathrm{u}},\delta t\right)$ learned by PINN and a stochastic component $\bm{\mathrm{W}}_{\mathsf{c}}\sim \widetilde{p}\left(\cdot|\bm{\mathrm{x}},\delta t\right)$, whose conditional distribution depends on both the current state and the elapsed time interval $\delta t$. 
In our implementation, the residual term $\widetilde{p}(\cdot\mid \bm{\mathrm{x}}_{t},\delta t)$ is instantiated as an isotropic Gaussian distribution, i.e., $\bm{\mathrm{W}}_{\mathsf{c}} \sim \mathcal{N}(\bm{0}, \delta t\cdot\sigma^2 \bm{I})$, providing a stochastic perturbation to the deterministic PINN dynamics. 
It is reasonable to incorporate $\delta t$ into the variance scale, since a longer interaction interval may lead to greater uncertainty.
The details of training $\bm{\mathrm{F}}^{\mathsf{nn}}_{\xi}\left(\bm{\mathrm{x}},\bm{\mathrm{u}},\delta t\right)$ are presented in Appendix \ref{appendix:PINN}. 

\textbf{Patient-specific models.}
Patient specificity is implemented at the dynamics-model level. 
For patient $i$, we train $\bm{\mathrm{F}}^{\mathsf{nn}}_{\xi_i}$ using only that patient's longitudinal trajectory, where $\xi_i$ serves as an abstract index of the resulting individualized dynamics rather than an explicitly inferred patient representation. 
Thus, different realizations of $\xi$ are operationalized in the current benchmark through different patient-specific datasets and learned dynamics models. 
This construction is sufficient for the benchmark objective of examining how patient-specific dynamics affect prediction and decision making, while separating it from the distinct problem of identifying patient-specific factors from clinical information. 
Learning an explicit and transferable representation of $\xi$ is therefore left as an open problem, and future personalization methods can be evaluated in \textit{MedGym} by how well their inferred patient representations recover individualized dynamics and treatment performance.

\textbf{Reason for using PINN.}
We adopt a PINN to exploit available physiological structure without assuming that the simplified ODE is exact. 
The physiology-inspired component provides a clinically meaningful inductive bias, while the neural residual absorbs model mismatch and patient-specific effects from data, offering a practical balance between purely mechanistic and fully data-driven dynamics models. 
Moreover, learning the resulting solution map as a function of $(\bm{\mathrm{x}},\bm{\mathrm{u}},\delta t)$ enables direct evaluation at arbitrary elapsed times without repeatedly solving the ODE during RL rollouts, which is particularly suitable for the irregular-time and simulation-intensive setting of \textit{MedGym}.

\textbf{Sepsis instantiation.}
The reward is defined from a smooth SOFA-based surrogate with a penalty on vasopressor usage, while blood lactate is monitored separately as a trajectory-level safety quantity; details are given in Appendix~\ref{appendix:spesis_treatment_formulation}. 
These components are application-specific rather than intrinsic to \textit{MedGym}: for another disease, the state variables, treatment actions, reward, and safety quantities can be replaced while retaining the same continuous-time environment interface.

\noindent
\textbf{Online policy learning algorithms.}  
To evaluate online reinforcement learning methods in \textit{MedGym}, we consider both standard discrete-time RL algorithms and recent continuous-time RL approaches. 
For the discrete-time setting, representative policy-learning baselines include Soft Actor-Critic (SAC), Proximal Policy Optimization (PPO) \citep{Schulman2017PPO} and Trust Region Policy Optimization (TRPO), together with safety-aware variants such as PPO Lagrangian (PPO Lag) and TRPO Lagrangian (TRPO Lag). 
These methods provide suitable baselines for benchmarking treatment performance under the \textit{MedGym} environment. 
For the continuous-time setting, we follow the bounded-interaction
formulation of TACOS \citep{Treven_NeurIPS2024}, in which the policy
jointly selects a treatment action and the duration for which that action
is applied.
TACOS reformulates the continuous-time control problem as an equivalent
extended discrete-time MDP by accumulating the running reward over each
selected physical-time interval.
This formulation is particularly suitable for medical treatment, where
measurements and treatment updates occur irregularly and the number of
interventions is limited.
The corresponding reformulation used in \textit{MedGym} is detailed in
Appendix~\ref{appendix:online_policy_learning}. 
Furthermore, as discussed in \citep{Shen_NeurIPS2025}, online policy learning methods can be naturally adapted to offline policy learning once a predictive environment model is available, thereby connecting online RL and model-based offline RL within the same benchmark framework.

\noindent
\textbf{Offline policy learning algorithms.}  
To evaluate offline reinforcement learning methods in \textit{MedGym}, we consider several representative value-based algorithms, including Deep Q-Network (DQN) \citep{Mnih_nature, Raghu, Komorowski}, Conservative Q-Learning (CQL) \citep{Kumar}, and Guarded CQL \citep{Shen_NeurIPS2025, Tumay2025GuardianRL}. 
These methods learn treatment policies from pre-collected trajectories without requiring additional interaction with the environment during training, which is particularly relevant in medical settings where data are often retrospective and online exploration is limited \citep{Levine2020OfflineRLTutorial, Thomas2016DataEfficientOPE, Voloshin2021EmpiricalOPE, Prudencio2023OfflineRLSurvey, Uehara2022OPEReview}. 
DQN provides a standard value-based baseline for policy learning from offline data. 
CQL mitigates extrapolation error by constraining learned actions to remain close to those supported by the dataset, making it particularly suitable for offline decision-making. 
Guarded CQL further improves robustness by penalizing overestimation on states and actions insufficiently supported by the data. 
Together, these methods cover a representative spectrum of commonly used offline RL approaches and enable a meaningful comparison of policy learning performance in longitudinal medical treatment settings.

\textbf{Patient-model split.}
Let $N_{\mathsf{psm}}$ and $n_{\mathsf{clu}}$ denote the number of patient-specific models and the desired number of clusters, respectively. As detailed in Appendix~\ref{appendix:PINN}, patient models with low mutual estimation error are assigned to the same cluster. For two patient models $M_i$ and $M_j$, this error evaluates $M_j$ on transitions generated by $M_i$ and $M_i$ on transitions generated by $M_j$, providing a symmetric measure of their predictive compatibility. The cluster number $n_{\mathsf{clu}}$ is configurable, allowing the benchmark to control the trade-off between data sharing and personalization. Each cluster $\mathcal{C}_k,k=1,...,N_{\mathsf{psm}}/n_{\mathsf{clu}}$ is partitioned into two disjoint subsets: the training patient set $\mathcal{C}_k^{\mathsf{train}}$ and the test patient set $\mathcal{C}_k^{\mathsf{test}}$, satisfying $\mathcal{C}_k^{\mathsf{train}}\bigcap\mathcal{C}_k^{\mathsf{test}}=\emptyset$ and $\mathcal{C}_k^{\mathsf{train}}\bigcup\mathcal{C}_k^{\mathsf{test}}=\mathcal{C}_k$. All test sets $\mathcal{C}_k^{\mathsf{test}}$ contain the same number of patient models.

\noindent
\textbf{Cluster- and population-level PINN models.}
For any given patient model $M_{i,k}\in\mathcal{C}_k^{\mathsf{train}}$, the training dataset $D_{i,k}^{\mathsf{tp}}$ used for PINN regression is generated through system identification. 
Starting states, admissible treatment actions, and elapsed time intervals are sampled, after which the patient model is simulated under each treatment action over the corresponding interval. 
Each generated transition records the current state, applied action, elapsed interval, and resulting successor state. 
We then form the $k-$th cluster training dataset $\mathcal{D}_{k,\mathsf{clu}}^{\mathsf{tp}}=\bigcup D_{i,k}^{\mathsf{tp}}$ by aggregating the datasets of all patient models in $\mathcal{C}_k^{\mathsf{train}}$. Similarly, the population training dataset is defined by $\mathcal{D}_{\mathsf{pop}}^{\mathsf{tp}}=\bigcup_{k}\mathcal{D}_{k,\mathsf{clu}}^{\mathsf{tp}}$. The PINNs trained on $\mathcal{D}_{k,\mathsf{clu}}^{\mathsf{tp}}$ and $\mathcal{D}_{\mathsf{pop}}^{\mathsf{tp}}$ are referred to as the $k$-th cluster PINN model and the population PINN model, respectively.  
Note that the test patients are not used for population or cluster model construction. 

\textbf{Individual-, Cluster-, and population-level policies.}
A patient-specific policy $\pi_{i,\mathsf{ind}}$ is trained either online through direct interaction with a given patient model $M_i$ or offline using the fixed dataset $D_{i,k}^{\mathsf{pl}}$. 
The latter is collected by rolling out a behavior policy on the patient model $M_i$. 
The $k$-th cluster-level policy $\pi_{k,\mathsf{clu}}$ is trained online by
interacting with the $k$-th cluster PINN, and the population policy
$\pi_{\mathsf{pop}}$ by interacting with the population PINN, so that each
policy is optimized against the dynamics model of its own personalization
level. For offline learning, $\pi_{k,\mathsf{clu}}$ is trained on the aggregated
dataset $\mathcal{D}_{k}^{\mathsf{pl}}=\bigcup_{i}D_{i,k}^{\mathsf{pl}}$ and
$\pi_{\mathsf{pop}}$ on the population-level dataset
$\mathcal{D}_{\mathsf{pop}}^{\mathsf{pl}}=\bigcup_{k}\mathcal{D}_{k}^{\mathsf{pl}}$.
For offline learning, $\pi_{\mathsf{pop}}$ is trained on the population-level dataset $\mathcal{D}_{\mathsf{pop}}^{\mathsf{pl}}=\bigcup_{k}\mathcal{D}_{k}^{\mathsf{pl}}$.

\section{Evaluation Experiments}
\label{sec:experiments}

\noindent
\textbf{Evaluated Policies.} 
We compare $\mathsf{SAC}$ \citep{haarnoja18b_sac}, $\mathsf{TRPO}$ \citep{schulman15_trpo}, and $\mathsf{TRPOLag}$ \citep{Pasula2020} under fixed and adaptive interaction schedules. 
The adaptive interaction mechanism follows \citep{Treven_NeurIPS2024}. 
We denote each method by $\mathsf{ALG}$-$\mathsf{T}$-$\mathsf{G}$, where $\mathsf{ALG}\in{\mathsf{SAC},\mathsf{TRPO},\mathsf{TRPOLag}}$ denotes the RL algorithm, $\mathsf{T}\in{\mathsf{F},\mathsf{A}}$ denotes fixed or adaptive timing, and $\mathsf{G}\in{\mathsf{Pop},\mathsf{Clu},\mathsf{Ind}}$ denotes the population, cluster, or individual model or policy.

\noindent
\textbf{Model and Dataset.}
The settings used to construct the individual-, cluster-, and population-level PINN model and policies are summarized as follows: The number of cohorts is set as $N_{\mathsf{coh}}=7$; In each cohort, the total number of patient-specific models is $N_{\mathsf{psm}}=110$; Clusters per cohort is $N_{\mathsf{clu}}=10$; 
Number of training models per cluster is $|\mathcal{C}_k^{\mathsf{train}}|=10$;
Number of test models per cluster is $|\mathcal{C}_k^{\mathsf{test}}|=1$;
Number of training models per cohort is
$\sum_{k=1}^{N_{\mathsf{clu}}}|\mathcal{C}_k^{\mathsf{train}}|=100$;
Number of test models per cohort is
$\sum_{k=1}^{N_{\mathsf{clu}}}|\mathcal{C}_k^{\mathsf{test}}|=10$.
Across the $N_{\mathsf{coh}}=7$ cohorts this gives $770$ patient-specific
models and $70$ held-out test patients, which is the evaluation set used in
Table~\ref{tab:policy_evaluation_selected}.
For the offline setting, we adopt the same construction with a single cohort ($N_{\mathsf{coh}}=1$), while all other settings remain unchanged.
Complete optimization hyperparameters are provided in Appendix Table~\ref{tab:hparam_rl}.

\noindent
\textbf{Benchmark questions.}
The evaluation is organized around four questions corresponding to the main benchmark dimensions of \textit{MedGym}.
\textbf{(Q1) Intervention timing:} Does allowing the policy to adapt the intervention interval improve treatment performance relative to a fixed interaction schedule?
\textbf{(Q2) Personalization:} How does policy performance change from population-level to cluster-level and patient-level treatment, and how much of the individual-oracle benefit can be recovered by cluster-based policies?
\textbf{(Q3) Learning setting:} Do the conclusions observed under online policy learning remain valid when policies must instead be learned from fixed offline datasets?
\textbf{(Q4) Treatment effectiveness versus safety:} Do improvements in treatment effectiveness also translate into safer physiological trajectories, or can the two evaluation criteria lead to different conclusions?

\subsection{Results and Discussions}
\begin{figure}[h]
\centering
\includegraphics[width=\linewidth]
{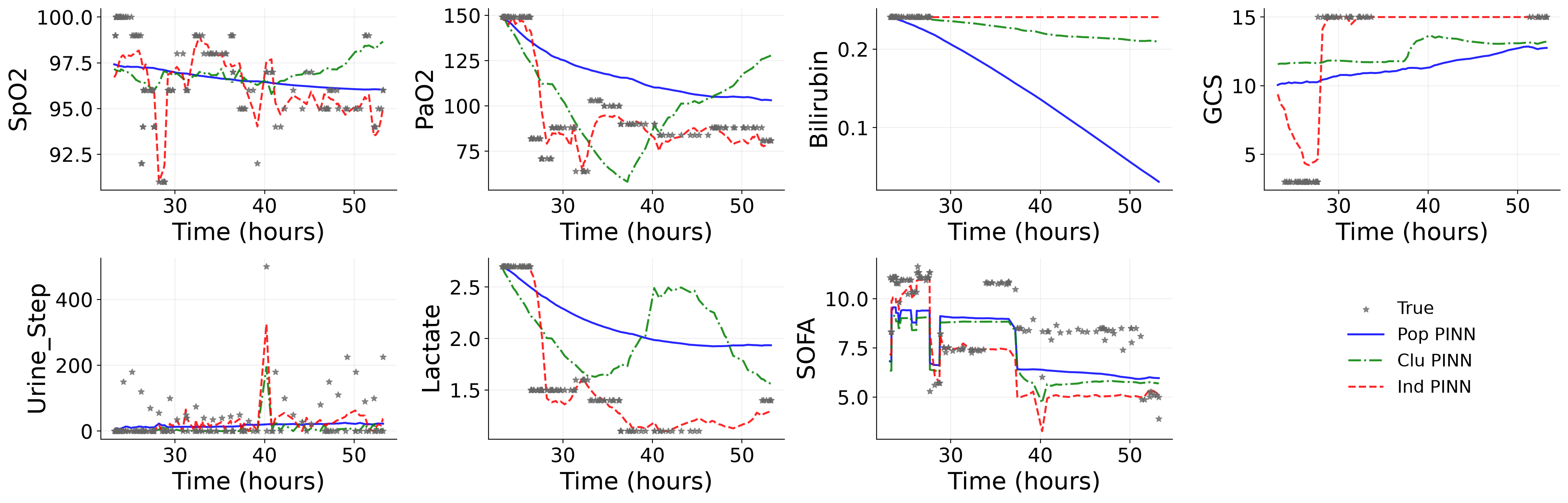}

\caption{
PINN rollout comparison for a representative held-out test patient ($T=130$ h).
The observed clinical trajectory is compared with full-horizon
deterministic rollouts generated by the patient-specific
($\mathsf{Ind}$), cluster-level ($\mathsf{Clu}$), and
population-level ($\mathsf{Pop}$) PINNs across six physiological
variables and the smoothed $\mathsf{SOFA}$ score. The test patient
is excluded from the construction of the cluster- and population-level
PINNs, while the patient-specific PINN serves as an oracle reference.
NMSE values are $0.314$, $0.784$, and $0.846$ for
$\mathsf{Ind}$, $\mathsf{Clu}$, and $\mathsf{Pop}$, respectively.
The $\mathsf{Ind}$ PINN is fitted to this patient's full trajectory and is
therefore evaluated in-sample.
As an additional check on generalization, we retrain the
$\mathsf{Ind}$ PINN for the same patient using only the first $70\%$ of the
trajectory and evaluate it on the remaining unseen $30\%$; the held-out
rollout still captures the main trends
(Appendix~\ref{fig:pinn-case-heldout}).
A separate temporal split is unnecessary for the $\mathsf{Clu}$ and
$\mathsf{Pop}$ models, since the test patient is already
excluded entirely from their construction and their rollouts are
therefore out-of-patient evaluations.}
\label{fig:pinn-case}
\end{figure}

\begin{figure}[h]
\centering
\includegraphics[width=\linewidth, height=0.5\linewidth]
{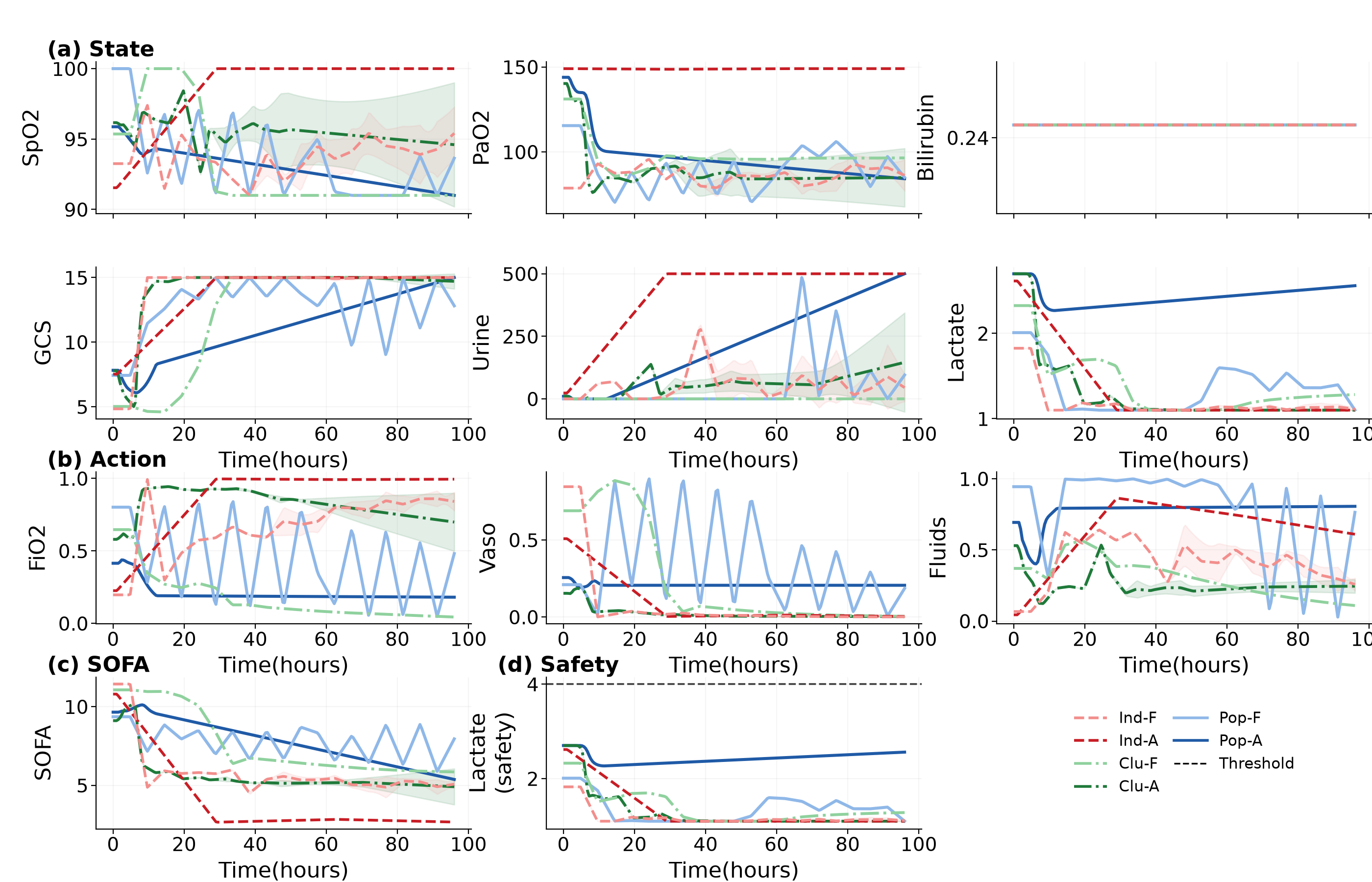}

\caption{
Closed-loop evaluation of $\mathsf{Ind}$, $\mathsf{Clu}$, and $\mathsf{Pop}$ policies on the patient-specific environment of the same representative test patient, using $\mathsf{TRPOLag}$ with $K=20$. Fixed and adaptive interaction timing are compared in terms of (a) physiological state trajectories, (b) treatment actions, (c) lactate trajectories and the safety
threshold of $4$ mmol/L, and (d) $\mathsf{SOFA}$ trajectories. Under adaptive timing, the mean lactate values are $1.289$, $1.338$, and $1.357$ mmol/L, and the mean $\mathsf{SOFA}$ scores are $3.082$, $5.563$, and $8.560$ for $\mathsf{Ind}$, $\mathsf{Clu}$, and
$\mathsf{Pop}$, respectively. Results are averaged over five
evaluation runs.
}
\label{fig:rl-case-202347}
\end{figure}

Figures~\ref{fig:pinn-case} and~\ref{fig:rl-case-202347} jointly illustrate how the degree of personalization in dynamics modeling and policy learning affects prediction and control for the same test patient. 
The patient-specific setting serves as an oracle reference: its PINN attains the highest prediction fidelity and its policy the most favorable control outcomes, and both degrade as the level of personalization coarsens from cluster-level to population-level. The degradation in control is not a fidelity artefact, since all three policies are evaluated on the same patient-specific simulator and only the training regime changes.
This consistent ordering indicates that patient heterogeneity should be considered in both environment construction and policy learning. An important direction for future medical RL is therefore to infer explicit
and transferable patient representations, such as $\xi$, from limited clinical histories and use them for patient-adaptive policy optimization on unseen patients. \textit{MedGym} provides a unified continuous-time
benchmark for evaluating how closely such methods recover individualized dynamics and treatment performance relative to cluster-level, population-level, and individual-oracle references. 

\begin{figure}[htbp]
  \centering

  \begin{subfigure}[t]{0.5\linewidth}
    \centering
    \includegraphics[width=\linewidth]{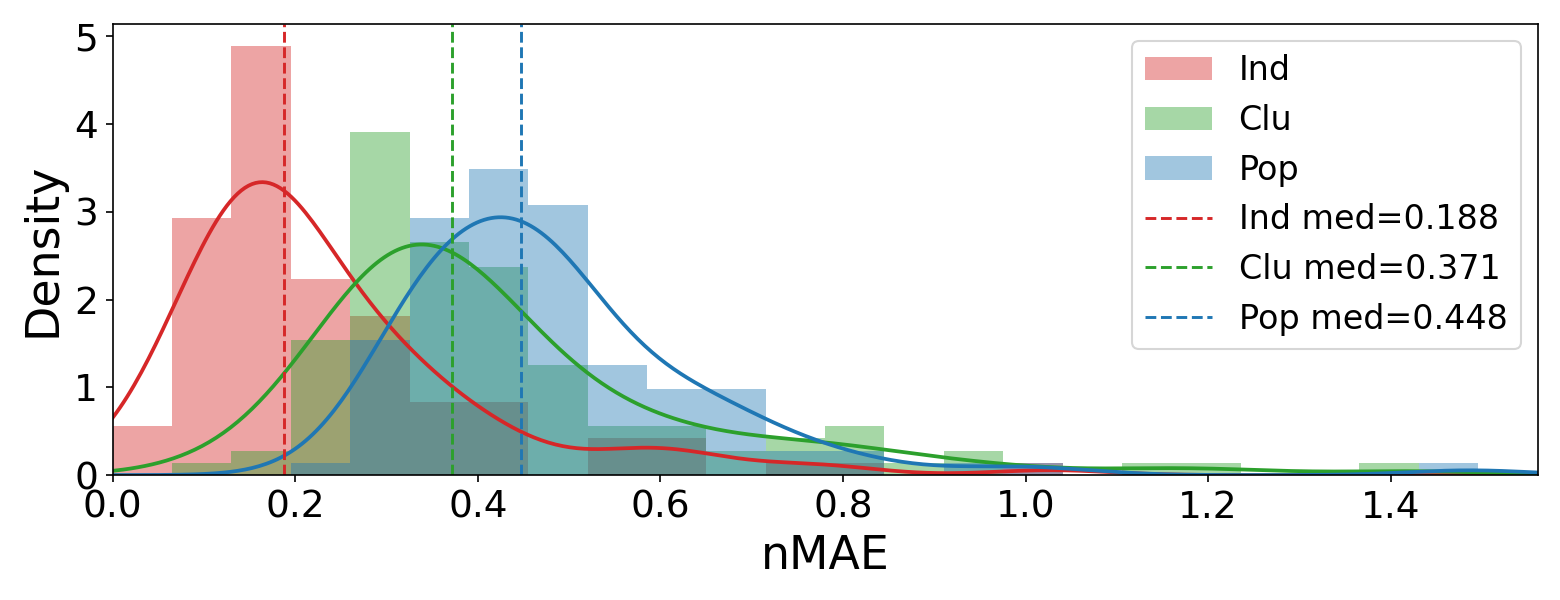}
    \caption{$\mathrm{nMAE}$}
    \label{fig:pinn_nmae}
  \end{subfigure}\hfill
  \begin{subfigure}[t]{0.5\linewidth}
    \centering
    \includegraphics[width=\linewidth]{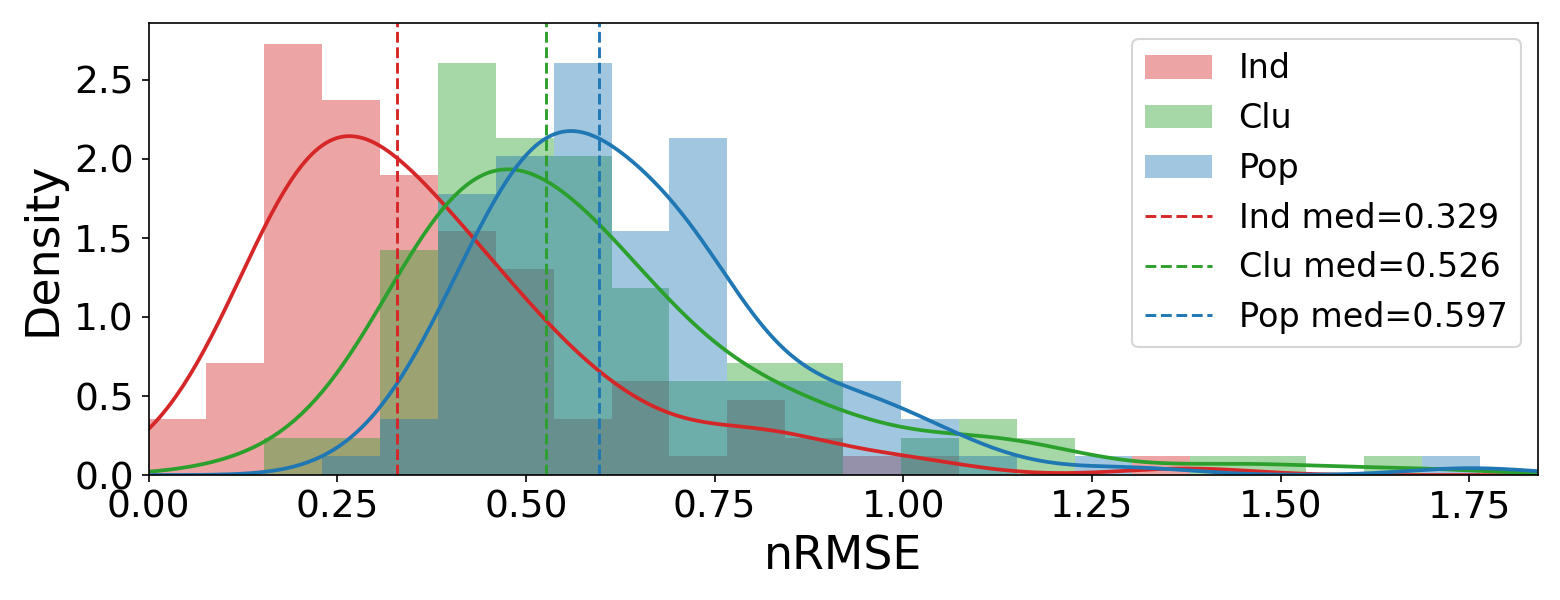}
    \caption{$\mathrm{nRMSE}$}
    \label{fig:pinn_nrmse}
  \end{subfigure}

  \caption{
    Simulator fidelity evaluation.
    Distributions of (a) normalized MAE and (b) normalized RMSE for the
    individual, cluster, and population PINNs.
  }
  \label{fig:comprehensive_evaluation}
\end{figure}

\begin{figure}[htbp]
\centering

\begin{subfigure}[t]{0.25\linewidth}
\centering
\includegraphics[width=\linewidth]
{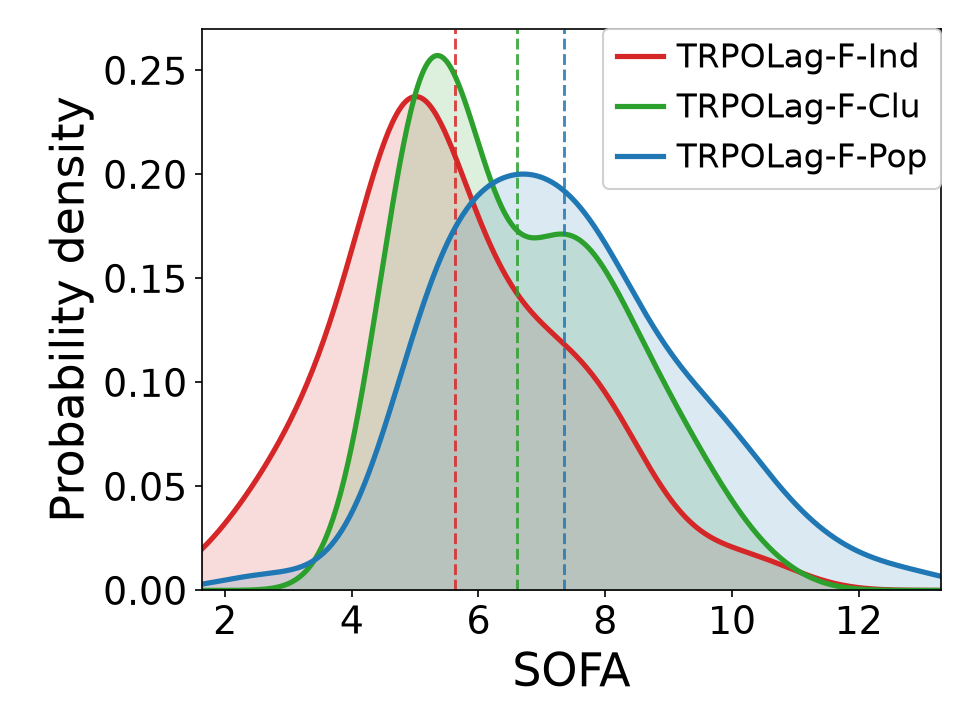}
\caption{Fixed}
\label{fig:final_sofa_dist_trpolag_fixed}
\end{subfigure}\hfill
\begin{subfigure}[t]{0.25\linewidth}
\centering
\includegraphics[width=\linewidth]
{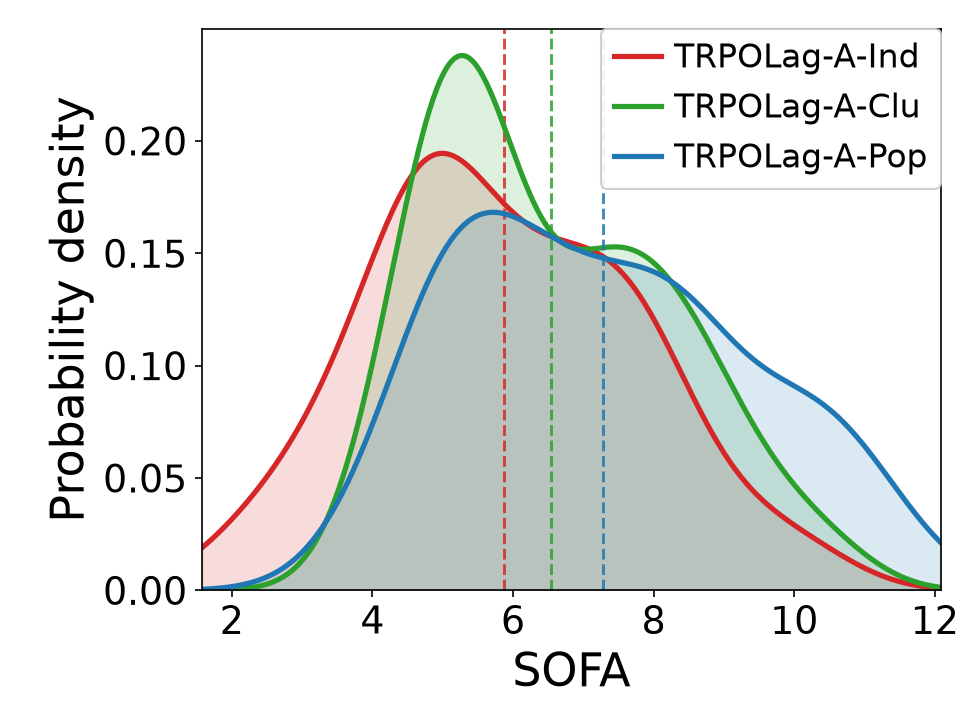}
\caption{Adaptive}
\label{fig:final_sofa_dist_trpolag_adaptive}
\end{subfigure}\hfill
\begin{subfigure}[t]{0.25\linewidth}
\centering
\includegraphics[width=\linewidth]
{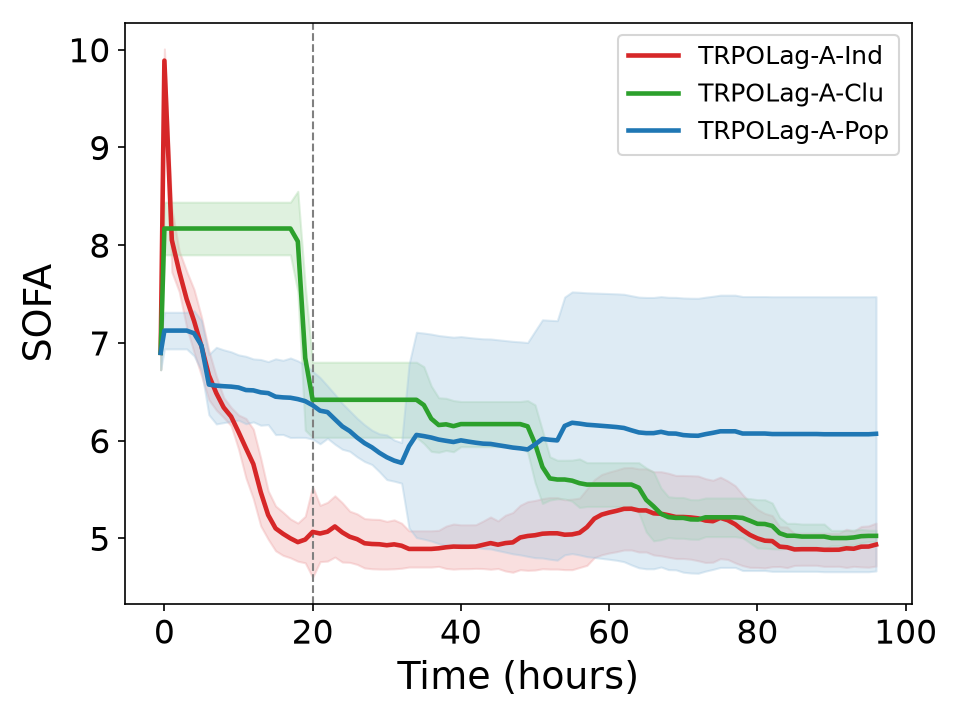}
\caption{$\mathsf{SOFA}$}
\label{fig:safety_main_a}
\end{subfigure}\hfill
\begin{subfigure}[t]{0.25\linewidth}
\centering
\includegraphics[width=\linewidth]
{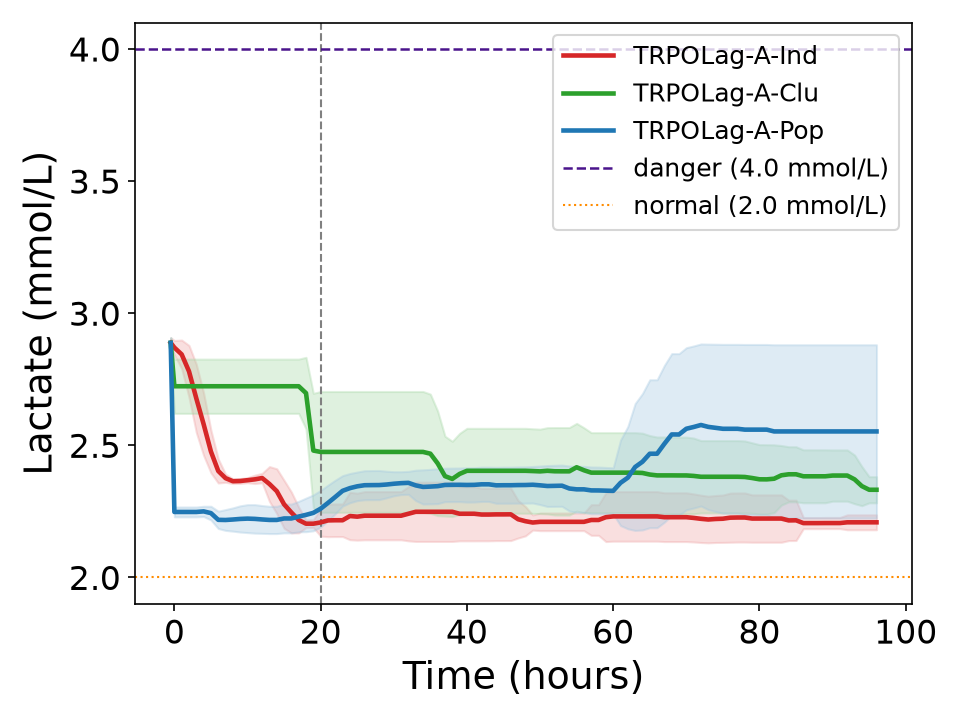}
\caption{Lactate}
\label{fig:safety_main_b}
\end{subfigure}

\caption{
Cohort- and trajectory-level online evaluation of the
individual-oracle ($\mathsf{Ind}$), cluster-level ($\mathsf{Clu}$),
and population-level ($\mathsf{Pop}$) $\mathsf{TRPOLag}$ policies.
(a,b) Distributions of final $\mathsf{SOFA}$ scores at 96 hours
across 70 held-out patients.
Each patient-level score is averaged over ten noise realizations.
(c,d) Representative $\mathsf{SOFA}$ and lactate trajectories (adaptive
timing).
}
\label{fig:sofa_distribution_over_cluster}
\end{figure}

We next examine whether the patient-level observations persist across
the held-out cohort. 
The normalized-error distributions in
Figure~\ref{fig:comprehensive_evaluation} show that the
patient-specific PINNs generally provide more accurate rollouts than the
population-level PINN. 
The cohort-level results in
Figure~\ref{fig:sofa_distribution_over_cluster} and
Table~\ref{tab:policy_evaluation_selected} are consistent with the
preceding patient-level case study. Under both timing settings, the mean
trajectory-wide and final $\mathsf{SOFA}$ scores follow the ordering
$\mathsf{Ind}<\mathsf{Clu}<\mathsf{Pop}$. The cluster policy outperforms
its matched population policy for $71.4\%$ and $65.7\%$ of patients under
fixed and adaptive timing, respectively, showing that cluster-based
training recovers part of the individual-oracle benefit. Adaptive timing
reduces the trajectory-wide $\mathsf{SOFA}$ score for the population,
cluster, and individual policy classes, although its effects on final
$\mathsf{SOFA}$ and lactate safety are not uniform
(Table~\ref{tab:policy_evaluation_selected}). Additional results
for $\mathsf{SAC}$, $\mathsf{PPO}$, $\mathsf{TRPO}$, and
$\mathsf{PPOLag}$ are presented in
Appendix~\ref{appendix:online_policy_evaluations} and exhibit the same
overall personalization trend.

\begin{table}[H]
  \centering
  \small
  \setlength{\tabcolsep}{3pt}
  \renewcommand{\arraystretch}{0.9}
  \captionsetup{skip=3pt}

  \resizebox{0.85\linewidth}{!}{%
  \begin{tabular}{@{}lcccc@{}}
    \toprule
    Policy
    & SOFA(all) ($\downarrow$)
    & Lower SOFA vs.\ Pop (\%)
    & SOFA(96h) ($\downarrow$)
    & SafeTime\% ($\uparrow$) \\
    \midrule

    $\mathsf{TRPOLag}$-$\mathsf{F}$-$\mathsf{Ind}$
    & $6.41 \pm 1.72$
    & $91.4$
    & $\mathbf{5.63 \pm 1.76}$
    & $\mathbf{97.46 \pm 12.65}$ \\

    $\mathsf{TRPOLag}$-$\mathsf{A}$-$\mathsf{Ind}$
    & $\mathbf{6.20 \pm 1.87}$
    & $81.4$
    & $5.87 \pm 1.86$
    & $95.99 \pm 13.76$ \\

    $\mathsf{TRPOLag}$-$\mathsf{F}$-$\mathsf{Clu}$
    & $7.23 \pm 1.59$
    & $71.4$
    & $6.61 \pm 1.54$
    & $91.45 \pm 23.62$ \\

    $\mathsf{TRPOLag}$-$\mathsf{A}$-$\mathsf{Clu}$
    & $7.16 \pm 1.68$
    & $65.7$
    & $6.55 \pm 1.67$
    & $88.37 \pm 25.61$ \\

    $\mathsf{TRPOLag}$-$\mathsf{F}$-$\mathsf{Pop}$
    & $7.96 \pm 1.90$
    & --
    & $7.34 \pm 1.87$
    & $89.15 \pm 29.45$ \\

    $\mathsf{TRPOLag}$-$\mathsf{A}$-$\mathsf{Pop}$
    & $7.45 \pm 1.99$
    & --
    & $7.28 \pm 2.03$
    & $89.40 \pm 24.14$ \\

    \bottomrule
  \end{tabular}%
  }

  \caption{
    Online evaluation of $\mathsf{TRPOLag}$ on patient-specific simulators
    corresponding to 70 held-out patients. SOFA(all) and SOFA(96h) denote
    the trajectory-wide and final $\mathsf{SOFA}$ scores, respectively.
    Results are reported as mean and standard deviation across patients, each policy being trained with a single seed.
    Lower SOFA vs.\ Pop denotes the percentage of matched patients for
    which the individual or cluster policy achieves a lower final
    $\mathsf{SOFA}$ score than the population policy under the same timing
    setting. SafeTime\% is the fraction of physical time with lactate below $4.0$ mmol/L. F and A denote fixed and adaptive timing. Boldface indicates the best mean in each SOFA or safety column and does not imply statistical significance.
  }
  \label{tab:policy_evaluation_selected}
\end{table}

Besides, we examine whether the personalization benefits also exist in the offline
setting, where policies must be learned from fixed behavior data.
Table~\ref{tab:offline_evaluation_dqn_cql_gcql} evaluates two representative
offline RL methods, $\mathsf{DQN}$ and Guarded Conservative Q-Learning
($\mathsf{GCQL}$), and compares their individual-, cluster- and population-level
policies under the same patient-specific evaluation environments. The online
$\mathsf{TRPOLag}$ policies are evaluated on the same simulators as a reference;
the individual one is also the behavior policy that generated the offline
datasets. Both offline methods improve on their population-level counterparts,
but the level of personalization at which the gain appears differs:
$\mathsf{GCQL}$ benefits only from full individualization, whereas for
$\mathsf{DQN}$ the cluster policy already recovers the gain. $\mathsf{GCQL}$ also
remains closest to the behavior policy while staying nearer to the support of the
behavior data. The advantage is not automatic, however: for $\mathsf{CQL}$ the
cluster policy falls behind the population policy rather than lying between it and
the individual one. These results show that access to patient-specific data alone
is not sufficient to guarantee improved offline policies; the level of
personalization at which the benefit appears depends on the learning algorithm's
ability to control distribution shift from the fixed dataset. This distinction
illustrates an important role of \textit{MedGym}: the benchmark can reveal when
clinically relevant structure such as patient heterogeneity is successfully
exploited by an RL method and when it is not. The $\mathsf{CQL}$ results, the
alternative patient grouping, and the support-distance analysis are reported in
Appendix~\ref{appendix:offline_policy_evaluations}.

\begin{table}[h]
\centering
\fontsize{8pt}{8pt}\selectfont
\setlength{\tabcolsep}{4pt}
\renewcommand{\arraystretch}{0.82}
\setlength{\aboverulesep}{0.3ex}
\setlength{\belowrulesep}{0.3ex}
\captionsetup{skip=2pt}
\begin{tabular}{lcccc}
\toprule
Policy & SOFA(all) ($\downarrow$) & Lower SOFA vs. Pop (\%) & SOFA(96h) ($\downarrow$) & SafeTime\% ($\uparrow$) \\
\midrule
$\mathsf{TRPOLag}$-$\mathsf{F}$-$\mathsf{Ind}$ (behavior)
  & $5.89 \pm 0.66$ & $90.0$ & $5.16 \pm 0.85$ & $100.00 \pm 0.00$ \\
\midrule
$\mathsf{DQN}$-$\mathsf{F}$-$\mathsf{Ind}$ & $7.88 \pm 1.02$ & $90.0$ & $7.81 \pm 1.01$ & $83.70 \pm 33.63$ \\
$\mathsf{DQN}$-$\mathsf{F}$-$\mathsf{Clu}$ & $7.84 \pm 0.93$ & $100.0$ & $7.74 \pm 1.01$ & $89.06 \pm 29.24$ \\
$\mathsf{DQN}$-$\mathsf{F}$-$\mathsf{Pop}$ & $9.20 \pm 0.97$ & --     & $9.34 \pm 0.92$ & $78.80 \pm 38.91$ \\
\midrule
$\mathsf{GCQL}$-$\mathsf{F}$-$\mathsf{Ind}$ & $\mathbf{6.27 \pm 0.72}$ & $\mathbf{80.0}$ & $\mathbf{5.82 \pm 1.02}$ & $\mathbf{99.94 \pm 0.18}$ \\
$\mathsf{GCQL}$-$\mathsf{F}$-$\mathsf{Clu}$ & $7.15 \pm 0.75$ & $40.0$ & $7.00 \pm 0.79$ & $89.68 \pm 20.67$ \\
$\mathsf{GCQL}$-$\mathsf{F}$-$\mathsf{Pop}$ & $7.21 \pm 0.88$ & --     & $7.06 \pm 0.97$ & $89.59 \pm 21.85$ \\
\bottomrule
\end{tabular}
\caption{Offline-policy results for DQN, and GCQL on
10 held-out test patients. Metrics and abbreviations follow
Table~\ref{tab:policy_evaluation_selected}; dashes indicate that the
population policy is not compared with itself.}
\label{tab:offline_evaluation_dqn_cql_gcql}
\end{table}

\begin{table}[htbp]
\centering
\fontsize{8pt}{8pt}\selectfont
\setlength{\tabcolsep}{4pt}
\renewcommand{\arraystretch}{0.82}
\setlength{\aboverulesep}{0.3ex}
\setlength{\belowrulesep}{0.3ex}
\captionsetup{skip=2pt}

  \resizebox{0.92\linewidth}{!}{%
  \begin{tabular}{@{}lccccc@{}}
    \toprule
    Policy
      & MAP(all) [mmHg]
      & Higher Safe\% vs.\ Pop (\%) ($\uparrow$)
      & MAP(48h) [mmHg]
      & SafeTime\% ($\uparrow$)
      & Lac(all) [mmol/L] ($\downarrow$) \\
    \midrule

       \midrule

    $\mathsf{TRPOLag}$-$\mathsf{F}$-$\mathsf{Ind}$
      & $69.47 \pm 5.52$  & $50.0$ & $68.90 \pm 4.97$
      & $77.8 \pm 26.8$   & $1.948 \pm 1.346$ \\

    $\mathsf{TRPOLag}$-$\mathsf{A}$-$\mathsf{Ind}$
      & $\mathbf{72.89 \pm 3.52}$ & $\mathbf{90.0}$ & $\mathbf{72.53 \pm 4.21}$
      & $\mathbf{96.3 \pm 3.4}$   & $\mathbf{1.932 \pm 1.788}$ \\

    \addlinespace[2pt]

    $\mathsf{TRPOLag}$-$\mathsf{F}$-$\mathsf{Clu}$
      & $81.01 \pm 20.75$ & $50.0$ & $79.77 \pm 24.86$
      & $76.6 \pm 36.7$   & $3.227 \pm 2.073$ \\

    $\mathsf{TRPOLag}$-$\mathsf{A}$-$\mathsf{Clu}$
      & $73.25 \pm 18.03$ & $60.0$ & $72.27 \pm 19.78$
      & $61.8 \pm 37.1$   & $3.954 \pm 2.853$ \\

    \addlinespace[2pt]

    $\mathsf{TRPOLag}$-$\mathsf{F}$-$\mathsf{Pop}$
      & $77.99 \pm 21.74$ & -- & $79.09 \pm 29.14$
      & $52.7 \pm 34.7$   & $3.648 \pm 2.068$ \\

    $\mathsf{TRPOLag}$-$\mathsf{A}$-$\mathsf{Pop}$
      & $74.98 \pm 25.06$ & -- & $76.84 \pm 28.81$
      & $53.6 \pm 38.8$   & $3.565 \pm 2.465$ \\
    \midrule

    $\mathsf{GCQL}$-$\mathsf{F}$-$\mathsf{Ind}$
      & $\mathbf{70.46 \pm 5.32}$ & $\mathbf{70.0}$ & $\mathbf{70.97 \pm 5.52}$
      & $\mathbf{86.1 \pm 19.5}$  & $\mathbf{2.005 \pm 1.338}$ \\

    $\mathsf{GCQL}$-$\mathsf{F}$-$\mathsf{Clu}$
      & $72.52 \pm 9.23$ & $40.0$ & $82.17 \pm 27.92$
      & $61.5 \pm 31.3$  & $3.298 \pm 1.907$ \\

    \addlinespace[2pt]

    $\mathsf{GCQL}$-$\mathsf{F}$-$\mathsf{Pop}$
      & $77.56 \pm 20.69$ & -- & $78.18 \pm 26.76$
      & $65.1 \pm 30.8$   & $3.587 \pm 2.025$ \\

    \bottomrule
  \end{tabular}%
  }
  \caption{%
    Evaluation of $\mathsf{TRPOLag}$ and $\mathsf{GCQL}$ policies on the acute hypotension cohort, reported as mean~$\pm$ standard deviation across the 10 held-out test patients.
    MAP(all) and MAP(48h) are the trajectory-wide and the final mean arterial pressure, with clinical target zone $[65,80]$~mmHg.
    SafeTime\% is the fraction of physical time with MAP~$\ge 65$~mmHg, that is, it counts only the lower bound of the target zone, and Higher Safe\% vs.\ Pop the percentage of patients for which Ind or Clu exceeds the population policy.
    Lac(all) is mean blood lactate, safe below $2.0$~mmol/L.
  }
  \label{tab:hypo_policy_evaluation}
\end{table}

For a more comprehensive analysis of the trajectory-level safety, additional schedule-independent trajectory-level safety analyses for the online policies are reported in Appendix~\ref{appendix:online_safety_evaluation}.
To assess whether the benchmark construction and its principal evaluation
dimensions are specific to MIMIC-III, we additionally instantiate
\textit{MedGym} using \textit{the Synthetic Acute Hypotension Dataset}.
The same environment-construction and evaluation protocol is applied with
dataset-specific physiological variables and treatment definitions in Appendix~\ref{appendix:hypotension_dataset}.
The resulting evaluation again exhibits meaningful differences across
fixed and adaptive intervention timing and across population-, cluster-,
and individual-level policies, indicating that these benchmark dimensions
are not artifacts of the MIMIC-III instantiation.
Table~\ref{tab:hypo_policy_evaluation} summarizes the representative results. 
Overall, \textit{MedGym} enables controlled evaluation of medical RL algorithms in terms of patient personalization, trajectory-level safety, and continuous-time decision-making. The results consistently show that these dimensions affect policy performance and should be considered explicitly.

\section{Conclusion}
\label{sec:conclusion}

We introduced \textit{MedGym}, a unified benchmark for dynamic medical treatment reinforcement learning. 
Built from clinical data, \textit{MedGym} provides continuous-time patient environments in which elapsed treatment intervals and patient-specific dynamics explicitly affect state evolution. 
On top of this environment family, the benchmark supports controlled evaluation across fixed and adaptive intervention timing, different levels of personalization, online and offline policy learning, and treatment effectiveness and trajectory-level safety.
Our experiments demonstrate that these evaluation dimensions materially affect the conclusions drawn about treatment policies. 
In particular, policy performance depends on intervention timing and the degree of personalization, while treatment effectiveness and safety need not improve simultaneously. 
These results highlight the importance of evaluating dynamic medical treatment RL beyond a single fixed-step, population-level setting.

\bibliographystyle{iclr2027_conference}
\bibliography{references}

\appendix
\section{Limitations}
\label{appendix:limitations}

\textit{MedGym} has several limitations that should be taken into account when interpreting the benchmark results.

First, the individual- and cluster-level personalization settings in the current benchmark should be interpreted as retrospective reference settings rather than deployable personalization procedures for previously unseen patients. 
The individual setting has access to a patient-specific dynamics model learned from that patient's trajectory and therefore serves as an oracle upper bound on the potential benefit of personalization.
Likewise, the current cluster assignment is determined from similarities between learned patient-specific dynamics, which requires substantially more patient-specific information than would generally be available at the beginning of treatment. 
The present experiments therefore evaluate how
policy performance changes when increasingly patient-specific dynamics are available, but they do not solve the separate problem of inferring an individual representation or assigning a new patient to an appropriate cluster from a short partial history. 
Learning such transferable representations and online assignment mechanisms is an important direction for future work.

Second, while \textit{MedGym} supports trajectory-level safety evaluation in continuous time, the current benchmark does not include a dedicated continuous-time safe RL method. 
As a result, the present study mainly evaluates safety through policy outcomes and safety-aware reward or constraint formulations rather than through algorithms specifically designed for continuous-time safe exploration or safe policy improvement. 
Future work can extend \textit{MedGym} by incorporating such methods once they become available, which would further strengthen the benchmark for safety-critical treatment recommendations.

Third, the current experiments do not include POMDP-specific algorithms, even though partial observability is highly relevant in medical treatment. 
In practice, the true patient condition is only indirectly revealed through clinical measurements, and therefore many treatment problems are naturally partially observable. 
Because \textit{MedGym} is built on a continuous-time latent state evolution model, extending the benchmark to a POMDP interface is conceptually straightforward: one can define an observation mapping from the latent physiological state to a possibly noisy and incomplete observation vector, and then let policies act on the observation history or belief state rather than on the full latent state. 
This extension is left for future work.

Fourth, \textit{MedGym} is constructed from observational clinical trajectories and therefore cannot establish the correctness of arbitrary counterfactual treatment responses. 
Each patient contributes observations under the treatment sequence that was actually administered, while the outcomes that would have occurred under alternative interventions are unobserved. Consequently, agreement with observed trajectories evaluates factual predictive fidelity but does not by itself validate simulator behavior under all state--action combinations that may be visited by a learned RL policy. 
This limitation cannot be eliminated by additional validation on another observational dataset, since such data likewise do not provide patient-level counterfactual ground truth. 
Establishing stronger counterfactual validity would require additional interventional evidence, richer treatment coverage together with appropriate causal assumptions, or independent physiological validation. 
Accordingly, \textit{MedGym} should be interpreted as a clinically grounded research benchmark for controlled algorithm comparison rather than as a validated digital twin for predicting real-world treatment effects.

Finally, the current paper primarily focuses on the sepsis cohort as the main clinically motivated instantiation, while also evaluating a second clinical setting based on the Synthetic Acute Hypotension Dataset, together with a limited set of representative RL methods.
While this is sufficient to demonstrate the key benchmark dimensions of continuous-time evolution, irregular intervention timing, personalization, and safety, broader validation across additional diseases, datasets, and algorithm classes would further improve the scope and robustness of the benchmark.

\section{Details of Environment Construction}
\label{appendix:details_environment_construction}

\subsection{MIMIC-III Dataset and Cohort Construction}
\label{appendix:cohort}
MIMIC-III (Medical Information Mart for Intensive Care) is a large-scale, publicly available clinical database containing detailed records from over $40{,}000$ intensive care admissions~\citep{Alistair}. 
The database includes longitudinal information such as demographics, vital signs, laboratory tests, medications, procedures, and other time-stamped clinical events, thereby providing a rich foundation for modeling dynamic treatment processes in critical care. 
For \textit{MedGym}, MIMIC-III is particularly valuable because it reflects the key characteristics emphasized in this work: patient conditions evolve continuously over time, while measurements and interventions are collected at irregular intervals determined by clinical practice rather than by a fixed sampling schedule. 
This temporal structure makes the dataset well suited for learning continuous-time patient evolution models and for constructing benchmark environments in which the elapsed time between observations and treatments meaningfully affects state transitions. 
In addition, MIMIC-III contains substantial heterogeneity across patients in terms of disease severity, physiological trajectories, intervention histories, and outcomes. 
Such variability is essential for building a benchmark that supports the evaluation of personalized treatment policies rather than only population-level decision rules. 
By grounding \textit{MedGym} in this real-world clinical diversity, we aim to construct an environment in which learned policies must account for both irregular treatment timing and individualized treatment response. 
Moreover, the widespread use of MIMIC-III in healthcare machine learning research, together with its extensive documentation and established preprocessing practices, facilitates reproducibility and comparability with prior studies. 
These properties make MIMIC-III an appropriate data foundation for \textit{MedGym} and for the evaluation of reinforcement learning methods in continuous-time dynamic treatment recommendation.

From this database we construct the sepsis cohort used throughout the paper,
yielding $110$ patients in each of seven disjoint cohorts and hence the $770$
patient-specific models used in Section~\ref{sec:experiments}. Because
measurements arrive at clinically determined times rather than on a fixed
schedule, the six state variables of Table~\ref{tab:state_selection} are
resampled onto a common grid and the remaining gaps are filled by forward
filling. This matters most for the laboratory variables, which are measured far
less frequently than the vital signs; bilirubin in particular is near-constant
within many stays, as the trajectories in Figure~\ref{fig:rl-case-202347} show,
so the hepatic component of the reward varies little for those patients.
Treatment actions are continuous and are clipped to fixed admissible ranges,
both when offline datasets are generated and when policies are evaluated, and
all state variables are standardised to zero mean and unit variance using statistics
computed from each patient's own record, while the action ranges are fixed across the
cohort; both are reused unchanged for every policy and evaluation rollout. The numerical values of the extraction window, the resampling interval, the action
bounds and the standardisation statistics are given in the released code repository.

\subsection{Sepsis Treatment: State, Reward, Safety, and Evaluation Quantities}
\label{appendix:spesis_treatment_formulation}

\noindent
\textbf{Overview.}
For the sepsis instantiation of \textit{MedGym}, we construct the treatment environment from the MIMIC-III sepsis cohort by extracting patient trajectories over a clinically relevant window around sepsis onset. 
Our goal is not to reproduce the full ICU record space, but to define a compact continuous-time treatment model that remains clinically meaningful, computationally tractable, and suitable for benchmarking both offline and online RL methods. 
To this end, we select a subset of physiological variables that are strongly associated with organ dysfunction and are routinely monitored in sepsis management. 
Treatment actions are modeled as continuous-valued clinical interventions, and the reward is designed from a smooth surrogate of organ failure severity. 
A key feature of \textit{MedGym} is personalization: each patient trajectory is associated with an individual-specific variable $\xi$, which captures heterogeneity in disease progression and treatment response. 
Therefore, the environment does not represent a single generic sepsis process, but rather a family of related patient-specific treatment dynamics. 
Table~\ref{tab:state_selection} summarizes the state variables, interventions, and reward components used in this sepsis setting.

\begin{table}[h]
\centering
\caption{Summary of State Variables, Treatment Actions, and Reward Design in the Sepsis Setting of \textit{MedGym}}
\renewcommand{\arraystretch}{1.3}
\begin{tabular}{|p{2.7cm}|p{4.8cm}|p{5.2cm}|}
\hline
\textbf{Category} & \textbf{Variables} & \textbf{Description} \\
\hline
\textbf{Dynamic Features} & 
SpO\textsubscript{2}, PaO\textsubscript{2}, Total bilirubin, GCS, Urine output, Lactate & 
Physiological variables describing respiratory, hepatic, neurological, renal, and metabolic status, selected to reflect major dimensions of organ dysfunction in sepsis. \\
\hline
\textbf{Actions} & 
FiO\textsubscript{2}, Vasopressor rate, Intravenous fluids & 
Continuous treatment controls corresponding to oxygen support, vasopressor administration, and fluid resuscitation in ICU care. \\
\hline
\textbf{Reward} & 
SOFA-based surrogate with vasopressor penalty & 
A smooth approximation of organ failure severity used as dense feedback, together with a regularization term discouraging unnecessarily aggressive vasopressor usage. \\
\hline
\end{tabular}
\label{tab:state_selection}
\end{table}

\textbf{Reward Function.}
We construct the reward from the Sequential Organ Failure Assessment (SOFA) score, since SOFA provides a clinically interpretable summary of multi-organ dysfunction and is more suitable for sequential treatment evaluation than a purely terminal outcome. 

The original SOFA scoring rule is threshold-based and assigns discrete integer
levels to each physiological component:
\begin{equation}
\label{eq:sofa_discrete}
\mathrm{SOFA}(\bm{\mathrm{x}},\bm{\mathrm{u}})
=
\sum_i \sum_k w_{i,k}\,\mathbf{1}\!\left(d_i\,(v_i-\tau_{i,k})\ge 0\right),
\end{equation}
where $\mathbf{1}(\cdot)$ is the indicator function, $i$ indexes the components,
$k$ indexes the thresholds $\tau_{i,k}$ of component $i$, and $w_{i,k}$ is the
weight of the corresponding level. The direction coefficient is $d_i=-1$ for
components scored when they fall below a threshold ($\mathrm{SpO_2}$,
$\mathrm{GCS}$, $\mathrm{Urine}$) and $d_i=+1$ for those scored when they exceed
one ($\mathrm{Bili}$, $\mathrm{Vaso}$), and all $v_i$ are evaluated in raw
clinical units. In \textit{MedGym} each indicator is replaced by a smooth
sigmoid step so that the reward remains informative between thresholds.
Following the clinical SOFA definition, the
cardiovascular component is scored through vasopressor dose, so the score
depends on the treatment action as well as on the patient state.
In our implementation, these thresholds are specified separately for each component according to the available variables in the dataset. 
For SpO$_2$, the cutoffs are 94, 90, 85, and 80 in percent. 
For bilirubin, the cutoffs are 1.2, 2.0, 6.0, and 12.0 mg/dL. 
For GCS, the cutoffs are 15, 13, 10, and 6. 
For urine output, the cutoffs are 500 and 200 mL/day. 
For vasopressor dosage, the cutoffs are 0.0, 0.05, 0.1, and 0.25 $\mu$g/kg/min. 
These choices follow the standard SOFA structure where possible, with modest adjustments to match the variables consistently available in our data.

In \textit{MedGym}, this reward construction is based on the subset of organ systems that can be modeled reliably from the available data, namely respiratory, hepatic, neurological, renal, and cardiovascular status, corresponding to SpO$_2$, bilirubin, GCS, urine output, and vasopressor usage. 
The coagulation component is not included because of substantial missingness and inconsistent temporal coverage. 
Thus, the adopted score preserves the additive threshold-based logic of SOFA while adapting it to the continuous-time benchmark setting and the practical constraints of the dataset.
We define the continuous-time running reward as
\begin{equation}
\label{eq:reward}
b(\bm{\mathrm{x}}_t,\bm{\mathrm{u}}_t)
=
-\,\mathrm{SOFA}(\bm{\mathrm{x}}_t,\bm{\mathrm{u}}_t)
-\lambda_{\mathsf{v}}\,\mathrm{Vaso}(\bm{\mathrm{u}}_t),
\end{equation}
which measures treatment utility per unit of physical time.
For a treatment action applied over an interval of duration $\delta t$,
the reward used by the RL algorithm is obtained by integrating this
running reward over the corresponding interval, following the TACOS
reformulation described in
Appendix~\ref{appendix:online_policy_learning}.

\textbf{State Space.}
The state variables are selected to balance clinical relevance and modeling tractability. 
Rather than retaining a very high-dimensional ICU representation, we focus on variables that directly reflect major aspects of sepsis-related organ dysfunction and are closely aligned with the reward definition. 
The resulting state includes respiratory, hepatic, neurological, renal, and metabolic indicators that are routinely monitored in critical care and sufficiently informative for treatment optimization. 
This choice yields a compact state representation that remains clinically meaningful while supporting stable learning and continuous-time trajectory modeling. 
In addition, \textit{MedGym} incorporates personalization through $\xi$, so the same observed state may evolve differently across patients under the same intervention, thereby enabling evaluation of individualized treatment policies rather than only population-level strategies.

\textbf{Explicit Clinical Safety Constraints.}
In addition to reward-based optimization, \textit{MedGym} includes an explicit safety cost built from a small set of clinical violation criteria, with blood lactate as the primary one. 
Lactate is widely used as an indicator of impaired perfusion and physiological instability in sepsis, and sustained elevation is strongly associated with clinical deterioration. 
For this reason, it provides an interpretable and operationally meaningful quantity for monitoring safety in the benchmark. 
The role of the safety constraint is to prevent policies from driving the patient into hazardous regions that may not be sufficiently penalized by the reward alone. 
This is particularly important in \textit{MedGym}, because treatments are applied over irregular intervals and unsafe evolution may arise between two consecutive decision times. 
By incorporating lactate-based safety monitoring, the benchmark supports evaluation of not only treatment effectiveness but also whether a policy maintains clinically acceptable continuous-time trajectories. 
Concretely, the safety cost has the same threshold form as \eqref{eq:sofa_discrete},
with a single threshold per criterion and unit weights,
\begin{equation}
c(\bm{\mathrm{x}},\bm{\mathrm{u}})
=
\sum_{j\in\mathcal{J}}
\mathbf{1}\!\left(d_j\,(v_j-\theta_j)\ge 0\right),
\label{eq:sepsis_cost}
\end{equation}
where $\mathcal{J}$ comprises vasopressor dose, lactate and oxygen saturation, with
$d_j=+1$, $\theta_j=\bar v$ for $\mathrm{Vaso}(\bm{\mathrm{u}})$; $d_j=+1$,
$\theta_j=4.0$ mmol/L for lactate; and $d_j=-1$, $\theta_j=88\%$ for
$\mathrm{SpO}_2$.
Thus $c\in\{0,1,2,3\}$ counts the clinical criteria simultaneously violated at a
decision point, weighted equally by default.cost limit

\textbf{Evaluation Quantities.}
Treatment effectiveness is reported by two quantities computed from the same
rollout. The trajectory-wide score is the physical-time average
\begin{equation}
\label{eq:sofa_all}
\mathrm{SOFA(all)}
=
\frac{1}{T}\int_{0}^{T}
\mathrm{SOFA}\!\left(\bm{\mathrm{x}}_t,\bm{\mathrm{u}}_t\right)\mathsf{d}t ,
\end{equation}
and $\mathrm{SOFA(96h)}=\mathrm{SOFA}(\bm{\mathrm{x}}_T,\bm{\mathrm{u}}_T)$ is
the score at the end of the horizon. Both are evaluated on the same
physical-time grid as the safety metrics of
Appendix~\ref{appendix:online_safety_evaluation}, so neither depends on the
number or the spacing of intervention times, and fixed- and adaptive-timing
policies are compared on identical quantities.

\subsection{Training Deterministic Part by PINN}
\label{appendix:PINN}

To evaluate reinforcement learning methods in a clinically grounded continuous-time benchmark, \textit{MedGym} constructs a family of patient-specific simulation environments from learned physiological dynamics. 
Each virtual patient is characterized by a personalization variable $\xi$, introduced in the main text, which modulates disease progression and treatment response. 
Consequently, \textit{MedGym} defines not a single simulator but a collection of related environments indexed by $\xi$, enabling systematic evaluation of personalization in dynamic treatment recommendation. 
The environment is based on a continuous-time state evolution model of the form
\begin{equation}
\label{eq:appendix_env_dynamics}
\bm{\mathrm{x}}_{t+\delta t}
=
\bm{\mathrm{F}}_{\xi}(\bm{\mathrm{x}}_{t},\bm{\mathrm{u}}_{t},\delta t)
+
\bm{\mathrm{w}}_{t,\delta t,\xi},
\end{equation}
where $\bm{\mathrm{x}}_{t}$ is the patient state, $\bm{\mathrm{u}}_{t}$ is the treatment action, $\delta t$ is the elapsed interaction interval, and $\bm{\mathrm{w}}_{t,\delta t,\xi}$ denotes residual uncertainty. 
The dependence on $\delta t$ allows the transition model to capture irregular treatment and measurement intervals, while the dependence on $\xi$ allows the dynamics to vary across patients.
To obtain a data-driven simulator, we approximate the deterministic part of the dynamics by a Physics-Informed Neural Network (PINN) \citep{RAISSI2019686}, yielding
\begin{equation}
\label{eq:appendix_env_pinn}
\widetilde{\bm{\mathrm{x}}}_{t+\delta t}
=
\widetilde{\bm{\mathrm{z}}}_{\mathsf{nn}}(\bm{\mathrm{x}}_{t},\bm{\mathrm{u}}_{t},\delta t,\xi)
=
\bm{\mathrm{F}}^{\mathsf{nn}}_{\xi}(\bm{\mathrm{x}}_{t},\bm{\mathrm{u}}_{t},\delta t)
+
\bm{\mathrm{W}}_{\mathsf{c}},
\qquad
\bm{\mathrm{W}}_{\mathsf{c}}\sim \widetilde{p}(\cdot\mid \bm{\mathrm{x}}_{t},\xi,\delta t).
\end{equation}
Here, $\bm{\mathrm{F}}^{\mathsf{nn}}_{\xi}(\bm{\mathrm{x}}_{t},\bm{\mathrm{u}}_{t},\delta t)$ is the personalized PINN approximation of the deterministic dynamics, while $\widetilde{p}(\cdot\mid \bm{\mathrm{x}}_{t},\xi,\delta t)$ models the residual stochasticity conditioned on the current state, the personalization variable, and the elapsed interval. 
Therefore, the transition mechanism in \textit{MedGym} is not tied to a fixed discrete step, but adapts to irregular interaction intervals and individualized patient dynamics in a principled manner.
In \textit{MedGym}, each episode is generated by sampling an initial patient condition together with a corresponding personalization variable $\xi$, and then rolling out the learned continuous-time dynamics under the chosen treatment policy. 
By varying both $\xi$ and the stochastic realization of $\bm{\mathrm{W}}_{\mathsf{c}}$, the benchmark can generate a diverse collection of patient trajectories that reflect heterogeneous disease progression, treatment sensitivity, and uncertainty accumulation over irregular decision intervals. 
This construction is particularly important for the intended benchmark usage of \textit{MedGym}. 
First, it preserves the continuous-time nature of patient evolution and therefore avoids the coarse approximation induced by fixed time discretization. 
Second, it explicitly incorporates personalization, allowing the benchmark to evaluate whether a method can adapt to patient-specific dynamics rather than relying only on population-level regularities. 
Third, because the same environment family can be used both to generate offline datasets and to evaluate learned policies through online rollouts, \textit{MedGym} provides a unified platform for comparing offline and online reinforcement learning methods under realistic dynamic treatment settings.

\noindent
\textbf{Physical Model for PINN.}
We describe the patient dynamics by a continuous-time ODE that captures coarse interactions among respiratory, metabolic, neurological, renal, and hepatic functions. 
The purpose of this model is not to provide a fully mechanistic physiological simulator, but to encode a clinically interpretable structural prior that guides learning toward plausible patient evolution. 
Accordingly, \textit{MedGym} combines a simplified physiology-inspired component with a flexible residual term learned from data. 
Personalization is introduced through $\xi$, so that both the disease progression law and treatment sensitivity are allowed to vary across patients.
The physiological state is $\mathbf{x} = [\mathrm{SpO_2},\,
\mathrm{PaO_2},\, \mathrm{Bili},\, \mathrm{GCS},\, \mathrm{Urine},\,
\mathrm{Lac}]^\top$ and the treatment action is $\mathbf{u} =
[\mathrm{FiO_2},\, \mathrm{Vaso},\, \mathrm{Fluid}]^\top$, where
$\mathrm{SpO_2}$ is peripheral oxygen saturation,
$\mathrm{PaO_2}$ arterial oxygen tension,
$\mathrm{Bili}$ serum bilirubin,
$\mathrm{GCS}$ the Glasgow Coma Scale,
$\mathrm{Urine}$ urine output,
$\mathrm{Lac}$ blood lactate,
$\mathrm{FiO_2}$ the fraction of inspired oxygen,
$\mathrm{Vaso}$ vasopressor dose, and
$\mathrm{Fluid}$ intravenous fluid rate.
All variables are standardised to zero mean and unit variance;
$[\cdot]^{+} = \max(\cdot,0)$ denotes rectification. 
The ODE is therefore defined in a normalized state space, where variables should be interpreted as relative deviations from nominal levels rather than absolute physiological quantities. 
For this reason, the terms below should be viewed as encoding directional and structural dependencies among variables, rather than exact physical laws.
The oxygenation block describes a first-order tendency toward an $\mathrm{FiO_2}$-dependent equilibrium, together with modulation by metabolic stress and fluid administration:
\begin{align}
\label{eq:PaO2_SpO2}
  \frac{d\,\mathrm{PaO}_2}{dt}
    &= k_1(\xi)(\mathrm{FiO}_2 - \mathrm{PaO}_2)
       - k_2(\xi)\,[\mathrm{Lac}]^+
       + k_3(\xi)\,\mathrm{Fluid}, \\[3pt]
  \frac{d\,\mathrm{SpO}_2}{dt}
    &= k_4(\xi)(\mathrm{PaO}_2 - \mathrm{SpO}_2).
\end{align}
The first equation models how inspired oxygen and supportive treatment affect arterial oxygenation, while the second links peripheral saturation to arterial oxygen tension.
Neurological status is modeled as deteriorating under metabolic burden and hypoxic stress:
\begin{align}
  \frac{d\,\mathrm{GCS}}{dt}
    = -k_5(\xi)\,[\mathrm{Lac}]^+ - k_6(\xi)\,h(\mathrm{SpO}_2),
\end{align}
where $h(\mathrm{SpO}_2)$ is a nonnegative function that increases when oxygen saturation falls below normal levels. 
In normalized coordinates, this function captures relative hypoxic deviation rather than a hard clinical threshold.
Lactate is driven upward by vasopressor-related hypoperfusion and oxygen deficit, and downward through clearance and dilution effects:
\begin{align}
  \frac{d\,\mathrm{Lac}}{dt}
    = k_7(\xi)\,\mathrm{Vaso}
      + k_8(\xi)\,h(\mathrm{SpO}_2)
      - k_9(\xi)\,[\mathrm{Lac}]^+
      - k_{10}(\xi)\,\mathrm{Fluid}.
\end{align}
Urine output reflects renal response to fluid input and is suppressed by metabolic and vasomotor stress:
\begin{align}
  \frac{d\,\mathrm{Urine}}{dt}
    = k_{11}(\xi)\,\mathrm{Fluid}
      - k_{12}(\xi)\,[\mathrm{Lac}]^+
      - k_{13}(\xi)\,\mathrm{Vaso}.
\end{align}
Bilirubin is used to represent slowly varying hepatic dysfunction:
\begin{align}
\label{eq:lac}
  \frac{d\,\mathrm{Bili}}{dt}
    = k_{14}(\xi)\,[\mathrm{Lac}]^+
      - k_{15}(\xi)\,[\mathrm{Bili}]^+.
\end{align}
The coefficients $\{k_i(\xi)\}$ determine the strength of organ-level couplings and treatment effects for each patient. 
They should be understood as patient-dependent effective parameters rather than directly measurable physiological constants. 
In the current implementation of \textit{MedGym}, personalization is realized by indexing patient-specific models, so that $\xi$ serves as an abstract label of individualized dynamics rather than an explicitly inferred latent representation. 
Thus, the present benchmark evaluates whether patient-specific dynamics improve prediction and control, while leaving the separate problem of learning an explicit transferable representation of $\xi$ to future work.
Since the true physiological dynamics are not fully known, the mechanistic component above is treated as an inductive bias rather than a complete model. 
There may exist substantial mismatch between the real dynamics and the simplified relations in \eqref{eq:PaO2_SpO2}--\eqref{eq:lac}. 
To capture this discrepancy, we use the following hybrid model:
\begin{equation}
\label{eq:Med_NN}
   \frac{d\bm{\mathrm{x}}}{dt}=\mathsf{MedODE}_{\xi}(\bm{\mathrm{x}},\bm{\mathrm{u}})+\mathsf{NN}_{\xi}(\bm{\mathrm{x}},\bm{\mathrm{u}}).
\end{equation}
Here $\xi$ collects all patient-specific learnable quantities, namely the coefficients of $\mathsf{MedODE}_{\xi}$ and the weights of $\mathsf{NN}_{\xi}$; a separate model is fitted to each patient and no parameter is shared across patients.
Rather than integrating \eqref{eq:Med_NN} numerically at every interaction, we jointly train a network that predicts the state derivative directly from $(\bm{\mathrm{x}},\bm{\mathrm{u}})$, supervised both by finite differences of the recorded trajectory and by the right-hand side of \eqref{eq:Med_NN}.
The personalized transition model $\bm{\mathrm{F}}^{\mathsf{nn}}_{\xi}(\bm{\mathrm{x}},\bm{\mathrm{u}},\delta t)$ used in \eqref{eq:appendix_env_pinn} advances this field over the elapsed time $\delta t$ with a fixed number of equal sub-steps, so that an interaction of arbitrary duration is produced without an external ODE solver.

\paragraph{Training objective and optimization}
The loss is a weighted sum of four terms: a data term matching the predicted
derivative to finite differences of consecutive observations, a multi-step term
that unrolls the predicted field, a smoothness term on consecutive predicted
derivatives, and a physics term matching the predicted field to the right-hand
side of \eqref{eq:Med_NN}.
Their weights follow a fixed three-stage schedule, from $20/40/10/0.1$ to
$5/20/20/0.02$, which shifts emphasis from trajectory fitting to physical
consistency; the physics term is evaluated at the recorded observation times, and
intervals shorter than $0.05$ hours are excluded because their finite differences
are dominated by measurement noise.
The derivative network uses three hidden layers of width $128$ and the neural
residual two hidden layers of width $128$, about $5\times10^{4}$ parameters per
patient in total, trained jointly with Adam at learning rates $10^{-4}$ and
$5\times10^{-5}$ for at most $10^{4}$ epochs on the patient's full trajectory.
The simulator also provides an observation noise model, which is disabled in all
reported experiments.

Overall, the PINN construction in \textit{MedGym} should be interpreted as a clinically motivated continuous-time structural prior together with a data-driven correction term. 
This design preserves continuous-time disease evolution, accommodates irregular interaction intervals, and supports personalized dynamics through $\xi$, while remaining sufficiently simple and modular for standardized benchmark evaluation. 

\noindent
\paragraph{Scope of the current physical model.}
The current implementation of \textit{MedGym} uses MIMIC-III sepsis trajectories as a concrete use case, and the physiology-inspired ODE in Appendix~B is therefore designed to reflect major organ interactions relevant to critical-care sepsis management. 
However, this does not mean that \textit{MedGym} is restricted to sepsis as a benchmark concept. 
The main contribution of \textit{MedGym} is the environment construction pipeline itself, namely the combination of a continuous-time structured dynamics model with data-driven residual learning for building personalized simulators from clinical data. 
This pipeline is modular and can be re-instantiated with different state variables, treatment actions, reward definitions, and physiology-inspired priors for other medical domains. 
For example, the same framework can in principle be extended to continuous-time treatment problems such as glucose management in diabetes or medication scheduling in oncology. 
Thus, the current sepsis setting should be understood as an initial benchmark instantiation rather than as a restriction on the broader applicability of the \textit{MedGym} framework.

\paragraph{Interpretation of the individualized model.}
The individualized model (\textit{Ind}) in \textit{MedGym} should not be interpreted as a practically deployable personalized learning method for unseen patients. 
Rather, it serves as an \emph{upper-bound evaluation setting} that measures the performance achievable when patient-specific dynamics are fully available. 
In this sense, the individualized model plays the role of an oracle benchmark for personalization, allowing us to quantify how much performance can potentially be gained when the underlying patient dynamics are accurately identified. 
The separate problem of inferring patient-specific representations such as $\xi$ from a patient’s partial history, and then using them to generalize to unseen patients, remains an important open research challenge. 
We view \textit{MedGym} as a benchmark that can support future evaluation of such representation-learning and patient-adaptation methods once they become available.

\noindent
\textbf{Patient clustering method.}
Let $N_{\mathsf{psm}}$ and $N_{\mathsf{clu}}$ denote the number of
patient-specific models and the desired number of clusters, respectively.
For patient model $M_i$, we represent its transition dynamics by the
pretrained patient-specific PINN
$\bm{\mathrm{F}}^{\mathsf{nn}}_{\xi_i}
(\bm{\mathrm{x}},\bm{\mathrm{u}},\delta t)$.
Rather than clustering raw clinical variables, the proposed method compares
the predictive behavior of patient-specific PINNs when they are evaluated
under the same patient trajectories.

For each transition of patient $j$, we first use its own PINN to generate a
one-step teacher target:
\begin{equation}
\label{eq:clustering_teacher}
    \widetilde{\bm{\mathrm{x}}}_{j,t+1}
    =
    \bm{\mathrm{F}}^{\mathsf{nn}}_{\xi_j}
    \left(
        \bm{\mathrm{x}}_{j,t},
        \bm{\mathrm{u}}_{j,t},
        \delta t_{j,t}
    \right).
\end{equation}
Each prediction is re-anchored at the observed state
$\bm{\mathrm{x}}_{j,t}$ rather than recursively initialized from the
preceding model prediction, thereby avoiding the accumulation of
long-horizon rollout errors.

The PINN of another patient $i$ is then evaluated using the same state,
treatment action, and elapsed time:
\begin{equation}
\label{eq:clustering_transfer_prediction}
    \widehat{\bm{\mathrm{x}}}_{i\rightarrow j,t+1}
    =
    \bm{\mathrm{F}}^{\mathsf{nn}}_{\xi_i}
    \left(
        \bm{\mathrm{x}}_{j,t},
        \bm{\mathrm{u}}_{j,t},
        \delta t_{j,t}
    \right).
\end{equation}
The transfer error from patient model $M_i$ to patient model $M_j$ is
defined as
\begin{equation}
\label{eq:clustering_transfer_error}
    E_{ij}
    =
    \frac{1}{d_{\mathsf{x}}T_j}
    \sum_{m=1}^{d_{\mathsf{x}}}
    \sum_{t=1}^{T_j}
    \left(
        \frac{
            \widehat{x}_{i\rightarrow j,t+1}^{(m)}
            -
            \widetilde{x}_{j,t+1}^{(m)}
        }{s_m}
    \right)^2,
\end{equation}
where $d_{\mathsf{x}}$ is the dimension of the physiological state,
$T_j$ is the number of valid transitions for patient $j$, and $s_m$
is the population-level interquartile range of the $m$-th state variable.
A smaller $E_{ij}$ indicates that the two patient-specific PINNs produce
more similar dynamics under the same clinical inputs.

Because the transfer-error scale may differ across target patients, each
column of the error matrix is robustly normalized:
\begin{equation}
\label{eq:clustering_normalized_error}
    Z_{ij}
    =
    \frac{
        E_{ij}
        -
        \underset{r\neq j}{\operatorname{median}}(E_{rj})
    }{
        \max\left\{
            \underset{r\neq j}{\operatorname{IQR}}(E_{rj}),
            \epsilon
        \right\}
    },
\end{equation}
where $\epsilon>0$ prevents numerical instability when the interquartile
range is close to zero.
The $i$-th row of $\bm{\mathrm{Z}}$ forms the transfer profile of patient
model $M_i$.

The symmetric distance between patient models $M_i$ and $M_{\ell}$ is
defined as the root-mean-square difference between their transfer profiles:
\begin{equation}
\label{eq:clustering_distance}
    \Delta_{i\ell}
    =
    \sqrt{
        \frac{1}{N_{\mathsf{psm}}-2}
        \sum_{j\notin\{i,\ell\}}
        \left(
            Z_{ij}-Z_{\ell j}
        \right)^2
    }.
\end{equation}
The columns corresponding to patients $i$ and $\ell$ are excluded so that
their diagonal self-errors do not affect the comparison.
A small $\Delta_{i\ell}$ indicates that the two patient-specific PINNs
respond similarly across other patient trajectories and therefore represent
similar disease-progression and treatment-response dynamics.

The resulting distance matrix is used for capacity-constrained medoid
clustering.
For cluster $\mathcal{C}_k$, its medoid is defined as
\begin{equation}
\label{eq:clustering_medoid}
    m_k
    =
    \underset{u\in\mathcal{C}_k}{\arg\min}
    \sum_{v\in\mathcal{C}_k}\Delta_{uv}.
\end{equation}
The clustering objective is
\begin{equation}
\label{eq:clustering_objective}
    \min_{\{\mathcal{C}_k\}_{k=1}^{N_{\mathsf{clu}}}}
    \sum_{k=1}^{N_{\mathsf{clu}}}
    \sum_{i\in\mathcal{C}_k}
    \Delta_{i,m_k},
\end{equation}
subject to each patient model belonging to exactly one cluster and each
cluster satisfying its prescribed capacity.

After clustering, each $\mathcal{C}_k$ is partitioned into the disjoint
sets $\mathcal{C}_k^{\mathsf{train}}$ and
$\mathcal{C}_k^{\mathsf{test}}$.
Thus, the cluster membership of each test patient is determined by the
PINN transfer-based clustering procedure before policy evaluation.
A test patient assigned to $\mathcal{C}_k^{\mathsf{test}}$ is evaluated under the cluster-level policy of cluster $k$, but always on its own patient-specific
simulator, and it is excluded from the construction and training of the cluster-
and population-level PINNs and policies.
The assignment therefore depends on similarities in learned patient
dynamics rather than the $\mathsf{SOFA}$ score or reward subsequently
obtained during policy evaluation.

\noindent
\paragraph{Held-out evaluation of the patient-specific PINN.}
\label{app:heldout}
The patient-specific PINN in Figure~\ref{fig:pinn-case} is fitted to the full
trajectory of the test patient and therefore serves as an oracle reference,
quantifying the accuracy attainable when the individual dynamics are fully
available rather than the accuracy of predicting unseen dynamics.
To separate these two cases, we train the individual PINN under a temporal split: the first $70\%$ of the trajectory is used for training the PINN while the remaining $30\%$ is used to evaluate the errors.

\begin{figure}[htbp]
\centering
\includegraphics[width=\linewidth]
{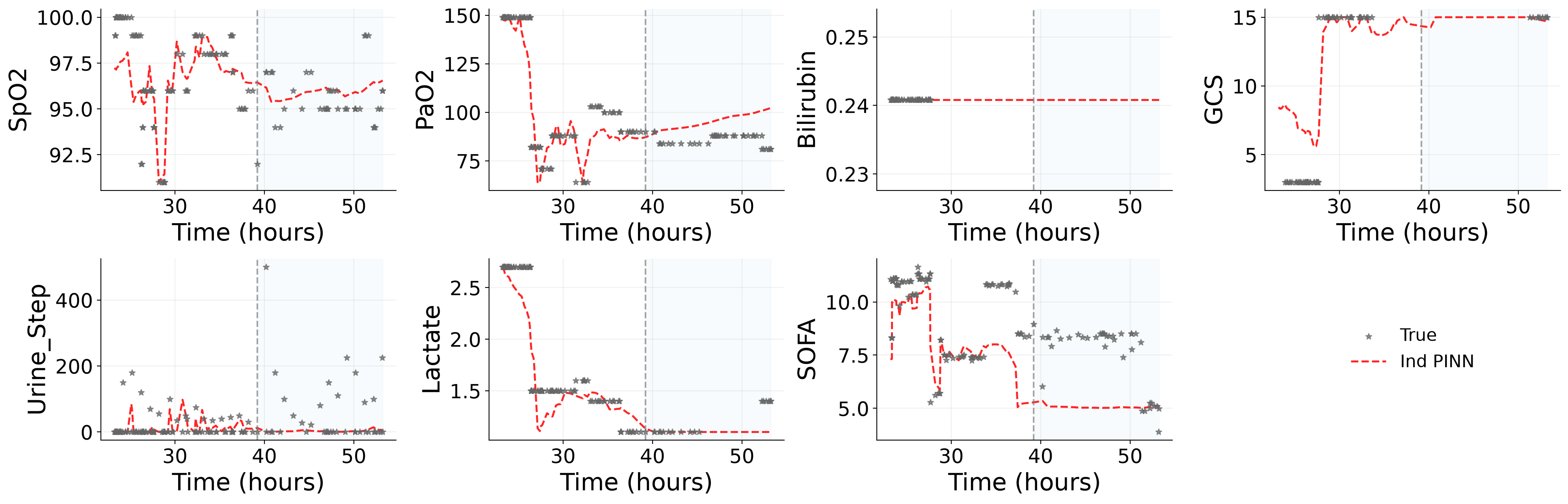}

\caption{
Held-out counterpart of Figure~\ref{fig:pinn-case} for the same representative test patient. The patient-specific PINN is trained only on the
first $70\%$ of the trajectory, indicated by the shaded region, and all rollouts
are evaluated on the remaining $30\%$. 
}
\label{fig:pinn-case-heldout}
\end{figure}

\subsection{Overview of the Synthetic Acute Hypotension Dataset}
\label{appendix:hypotension_dataset}

To evaluate whether \textit{MedGym} generalizes beyond sepsis management, we instantiate a
second environment from the Synthetic Acute Hypotension Dataset~\citep{healthgym2022}, a
GAN-generated corpus derived from MIMIC-III that contains $3{,}910$ ICU stays with
$187{,}680$ hourly state--action pairs over $48$ hours.
Acute hypotension (mean arterial pressure below $65$ mmHg) is a hemodynamic emergency with
physiological dynamics and clinical objectives distinct from the multi-organ pathophysiology
of sepsis: decisions are made hourly rather than on the irregular multi-hour scale of the
sepsis cohort, the reward is a continuous MAP-based signal rather than a discrete SOFA
surrogate, and the safety constraints differ.
The same continuous-time formulation and personalization mechanism are applied throughout,
with hyperparameters adjusted to the hourly resolution of this cohort.
 
\begin{table}[h]
\centering
\caption{%
  State variables, treatment actions, and reward design in the acute
  hypotension setting of \textit{MedGym}.
  Variables marked~$(\ast)$ are latent rather than observed.%
}
\renewcommand{\arraystretch}{1.3}
\begin{tabular}{|p{2.7cm}|p{4.8cm}|p{5.2cm}|}
\hline
\textbf{Category} & \textbf{Variables} & \textbf{Description} \\
\hline
\textbf{Differential States}
& $V^{(\ast)}$, $C_e^{(\ast)}$, MAP, PP, UO, Lac, Cr, Hep, PaO$_2$, GCS
& Effective circulating volume index, vasopressor effect-compartment
  concentration, mean arterial pressure, pulse pressure, urine output,
  lactate, creatinine, hepatic transaminase, arterial oxygen partial
  pressure, and Glasgow Coma Scale. \\
\hline
\textbf{Actions}
& $u_{\mathrm{fluid}}$ (mL/kg/h)\newline
$u_{\mathrm{vaso}}$ ($\mu$g/kg/min)\newline
  FiO$_2$
& Fluid resuscitation, vasopressor administration in noradrenaline
  equivalents, and supplemental oxygen. \\
\hline
\textbf{Reward}
& $r_{\mathrm{MAP}}$, $r_{\mathrm{vaso}}$, $r_{\mathrm{lac}}$
& Flat-top MAP reward on $[65,80]$ mmHg with quadratic vasopressor and
  linear lactate penalties. \\
\hline
\end{tabular}
\label{tab:hypo_state_selection}
\end{table}

\textbf{State Space.}
The state space comprises ten variables that evolve continuously in time: eight observed
physiological measurements and two latent quantities derived from the medication record, as
listed in Table~\ref{tab:hypo_state_selection}.
The two latent variables, the effective circulating volume index~$V$ and the vasopressor
effect-compartment concentration~$C_e$, are obtained by forward integration of
\eqref{eq:hypo_latent} along the recorded fluid and vasopressor history, and give the
simulator an explicit representation of volume
status and of the delay between an administered dose and its pressor effect.
Key differences from sepsis include the explicit separation of MAP and pulse pressure as
primary hemodynamic indicators, the hepatic marker ALT, and lactate as a direct indicator of
tissue perfusion adequacy.

\textbf{Action Space.}
The action space is three-dimensional, comprising the fluid rate, the vasopressor rate, and
the inspired oxygen fraction.
While fluid and vasopressor administration are originally categorical in the dataset, we
treat them as continuous to maintain consistency with the sepsis experiments and to enable
fine-grained policy optimization over arbitrary treatment intervals~$\delta t$.

\textbf{Reward Function.}
The clinical objective is to keep MAP within the target zone of $65$ to $80$ mmHg while
avoiding vasopressor overuse, which causes gut ischaemia and lactate elevation.
We therefore define the per-step reward as

\begin{equation}
b(\mathbf{x}_t, \mathbf{u}_t)
  = r_{\mathrm{MAP}}(\mathrm{MAP}_t)
  + r_{\mathrm{vaso}}(\tilde{u}_{\mathrm{vaso},t})
  + r_{\mathrm{lac}}(\mathrm{Lac}_t),
\label{eq:hypo_reward}
\end{equation}

where $\tilde{u}_{\mathrm{vaso}} \in [0,1]$ is the vasopressor dose normalized by its
maximum.
The MAP term is flat inside the target zone and decays quadratically outside it,

\begin{equation}
r_{\mathrm{MAP}}(\mathrm{MAP})=
\begin{cases}
+1, & 65 \le \mathrm{MAP} \le 80, \\[4pt]
\max\!\Bigl(-2,\;1 -
  \bigl(\tfrac{\mathrm{MAP}-65}{12.5}\bigr)^2\Bigr),
  & \mathrm{MAP} < 65, \\[4pt]
\max\!\Bigl(-2,\;1 -
  \bigl(\tfrac{\mathrm{MAP}-80}{12.5}\bigr)^2\Bigr),
  & \mathrm{MAP} > 80,
\end{cases}
\label{eq:r_map}
\end{equation}

and the penalty terms are

\begin{equation}
r_{\mathrm{vaso}}(\tilde{u}_{\mathrm{vaso}}) = -0.8\,\tilde{u}_{\mathrm{vaso}}^2,
\qquad
r_{\mathrm{lac}}(\mathrm{Lac}) = -0.5\max(0,\,\mathrm{Lac} - 2.0).
\label{eq:r_vaso_lac}
\end{equation}

The flat top removes the maximum-dose optimum that a monotone reward would admit, the
quadratic vasopressor penalty reflects the nonlinear toxicity of pressors, and the lactate
penalty applies only above the clinical threshold of $2.0$ mmol/L.
This continuous MAP-centerd design contrasts with the discrete multi-organ SOFA score used
in sepsis, and tests \textit{MedGym} under a fundamentally different reward structure. 

\textbf{Dynamics.}
As in \eqref{eq:Med_NN}, the mechanistic component is combined with a learned
residual and the two are trained jointly on each patient's record.
Volume status enters through a Frank--Starling relation and the vasopressor
through a first-order effect compartment, so that the two latent states follow
\begin{equation}
\frac{dV}{dt}
 = \frac{u_{\mathrm{fluid}}+\mathrm{UO}_0-\mathrm{UO}}{V_{\mathrm{ref}}}
   -\frac{(V-1)_+}{\tau_{\mathrm{leak}}},
\qquad
\frac{dC_e}{dt}
 = \frac{u_{\mathrm{vaso}}-C_e}{\tau_{c_e}},
\label{eq:hypo_latent}
\end{equation}
where $(\cdot)_+=\max(\cdot,0)$, and $u_{\mathrm{fluid}}$ and $u_{\mathrm{vaso}}$ are
the fluid and vasopressor components of $\mathbf{u}$.
The eight observed states are driven by these two: mean arterial pressure and
pulse pressure relax towards equilibria set by the Frank--Starling stroke volume
and by a saturating pressor effect that is attenuated under acidosis; urine output
relaxes towards a value set by autoregulated glomerular filtration, itself reduced
by venous congestion and by vasopressor-induced renal vasoconstriction; oxygen
tension relaxes towards the alveolar gas value corrected for a
congestion-dependent shunt; and lactate, creatinine, ALT and GCS follow
production--clearance balances driven by oxygen debt, perfusion deficit and gut
ischemia.
The complete system and the reference values of its parameters are provided in the
released code repository, and each rollout starts from the patient's first recorded
observation.

\textbf{Explicit Clinical Safety Constraints.}
We additionally monitor four physiological risks that reward optimization alone cannot
balance: tissue hypoperfusion indicated by elevated lactate, acute kidney injury indicated
by oliguria, hypoxaemia indicated by low PaO$_2$, and iatrogenic hypertension caused by
vasopressor overuse.
The per-step cost is the sum of four graded components,
\begin{equation}
c(\mathbf{x}_t) = c_{\mathrm{hyper}} + c_{\mathrm{lac}} + c_{\mathrm{UO}} + c_{\mathrm{PaO_2}},
\qquad c(\mathbf{x}_t) \in [0,\, 5],
\label{eq:hypo_cost}
\end{equation}
where
\begin{align}
c_{\mathrm{hyper}} &= \min\!\bigl(2,\,\max(0,\,(\mathrm{MAP}-85)/10)\bigr), \\
c_{\mathrm{lac}} &=
  \begin{cases}
    0, & \mathrm{Lac} < 2.0,\\
    \min(2,\,(\mathrm{Lac}-2.0)/2.0), & \mathrm{Lac} \ge 2.0,
  \end{cases} \\
c_{\mathrm{UO}} &= \min\!\bigl(1,\,1.5\max(0,\,(0.5 - \mathrm{UO})/0.5)\bigr), \\
c_{\mathrm{PaO_2}} &= \min\!\bigl(1,\,\max(0,\,(60 - \mathrm{PaO_2})/60)\bigr).
\end{align}
The urine threshold of $0.5$ mL/kg/h is the standard criterion for adequate renal perfusion,
and the lactate threshold of $2.0$ mmol/L is the established cutoff for tissue
hypoperfusion.
The Lagrangian-TRPO algorithm enforces $\mathbb{E}[c(\mathbf{x}_t)] \le d$ through an
adaptive multiplier~$\lambda$, with the budget set to $d = 0.8$ in our experiments.
Lactate therefore acts both as a reward penalty and as a constraint, mirroring the sepsis
setting, whereas the hypertension term is unique to this cohort and prevents the multiplier
from vanishing once MAP exceeds the upper bound of the target zone.

\section{Details of Policy Training}
\label{appendix:policy_training}


\subsection{Online Policy Learning}
\label{appendix:online_policy_learning}

This section describes the online policy training protocol used in \textit{MedGym}. Online reinforcement learning interacts directly with the corresponding simulators during training. Policies are optimized by repeatedly collecting rollouts from the simulator and updating the policy in a closed loop manner. We consider the major online reinforcement learning methods, such as Soft Actor Critic (SAC), Proximal Policy Optimization (PPO), and Trust Region Policy Optimization (TRPO), as well as their safety constrained variants, Lagrangian PPO and Lagrangian TRPO. Following the benchmarking pipeline in Figure~\ref{fig:benchmarking_pipeline}, each policy is trained through direct interaction with its corresponding MedGym environment.

\paragraph{Policy learning framework}

We formulate online policy learning as a continuous-time reinforcement learning problem on the MedGym environments. At each interaction step $k$, the policy $\pi$ observes the current physiological state $x_k$ and selects an action. For fixed-interval policies, the action corresponds to the treatment vector $u_k$, while for adaptive-time policies, the action is augmented as $\bar{u}_k=(u_k,\delta t_k)$, where $\delta t_k$ denotes the next intervention interval.

The state consists of the physiological variables listed in Table~\ref{tab:state_selection}, together with two additional components: the remaining time horizon and the remaining intervention budget.
These variables provide temporal context and allow the policy to adapt its strategy based on the remaining decision horizon.

For fixed-interval policies, the action corresponds to the treatment vector $u_k$ defined in Table~\ref{tab:state_selection}.
For adaptive-time policies, the action is augmented as $\bar{u}_k = (u_k, \delta t_k)$, where $\delta t_k$ denotes the next intervention interval.

Given $(x_k,\bar{u}_k)$, the patient state evolves according to the
PINN-based transition model.
For both fixed- and adaptive-timing policies, the reward associated with
an interaction is accumulated over the corresponding physical-time
interval.
The precise interval-reward definition and its relation to the
continuous-time objective are given below for the adaptive-timing
formulation.
Policies are optimized from closed-loop trajectories collected through
direct interaction with the simulator, in contrast to offline learning
from a fixed dataset.

\paragraph{Time-adaptive policy learning framework.}

We follow the bounded-number-of-interactions formulation of
TACOS \citep{Treven_NeurIPS2024}.
Let
\[
0=t_0<t_1<\cdots<t_K=T
\]
denote the $K$ intervention times over the fixed treatment horizon
$[0,T]$, and define
\[
\delta t_k:=t_{k+1}-t_k.
\]
At interaction $k$, the treatment action $\bm{\mathrm{u}}_k$ is applied
and held fixed over the interval $[t_k,t_{k+1})$.

Following TACOS, we denote the continuous-time state flow in
\textit{MedGym} by
\begin{equation}
\label{eq:medgym_state_flow}
\bm{\mathrm{x}}_{k+1}
=
\Xi_{\xi}
\left(
\bm{\mathrm{x}}_k,
\bm{\mathrm{u}}_k,
\delta t_k
\right),
\end{equation}
where
$\bm{\mathrm{x}}_k=\bm{\mathrm{x}}_{t_k}$ and
$\Xi_{\xi}$ is induced by the PINN-based continuous-time transition
model in \eqref{eq:pinn_ckd_approximate_solution_main}.

The corresponding reward flow is defined by accumulating the
continuous-time running reward over the same physical-time interval:
\begin{equation}
\label{eq:medgym_reward_flow}
\Xi_b
\left(
\bm{\mathrm{x}}_k,
\bm{\mathrm{u}}_k,
\delta t_k
\right)
:=
\int_{0}^{\delta t_k}
b
\left(
\Xi_{\xi}
(\bm{\mathrm{x}}_k,\bm{\mathrm{u}}_k,\tau),
\bm{\mathrm{u}}_k
\right)
\mathsf{d}\tau.
\end{equation}
We therefore define the discrete reward at interaction $k$ as
\begin{equation}
\label{eq:medgym_interval_reward}
r_k
=
\Xi_b
\left(
\bm{\mathrm{x}}_k,
\bm{\mathrm{u}}_k,
\delta t_k
\right).
\end{equation}

For the bounded-interaction setting, the policy uses the current patient
state, the remaining physical time, and the interaction index.
Accordingly, following TACOS, we consider
\begin{equation}
\pi:
\mathcal{X}\times\mathcal{T}\times\{0,\ldots,K-1\}
\rightarrow
\mathcal{U}\times[\delta t_{\min},\delta t_{\max}],
\end{equation}
where
\begin{equation}
\pi(\bm{\mathrm{x}}_k,T-t_k,k)
=
(\bm{\mathrm{u}}_k,\delta t_k).
\end{equation}
Thus, the adaptive policy determines both the treatment action and the
duration for which it is applied.

The resulting finite-horizon extended MDP optimizes
\begin{equation}
\label{eq:medgym_tacos_objective}
\max_{\pi}
\;
\mathbb{E}
\left[
\sum_{k=0}^{K-1}
r_k
\right],
\qquad
t_K=T.
\end{equation}
Since the intervals
$[t_k,t_{k+1})$ form a partition of the complete treatment horizon,
\begin{align}
\sum_{k=0}^{K-1}r_k
&=
\sum_{k=0}^{K-1}
\int_{t_k}^{t_{k+1}}
b(\bm{\mathrm{x}}_t,\bm{\mathrm{u}}_k)
\mathsf{d}t
\nonumber\\
&=
\int_{0}^{T}
b(\bm{\mathrm{x}}_t,\bm{\mathrm{u}}_t)
\mathsf{d}t.
\label{eq:medgym_tacos_equivalence}
\end{align}
Hence, the extended discrete-time objective
\eqref{eq:medgym_tacos_objective} is the TACOS reformulation of the
continuous-time treatment objective in
\eqref{eq:problem_original}.
In particular, increasing $\delta t_k$ does not reduce the accumulated
SOFA-based penalty, because treatment utility continues to accumulate
throughout the entire interval. 
Equation~\eqref{eq:medgym_tacos_equivalence} gives the exact
continuous-time-to-extended-MDP equivalence of the bounded-interaction
TACOS formulation when the finite-horizon return is undiscounted, i.e.,
$\gamma=1$ \citep{Treven_NeurIPS2024}.
In this case, changing the intervention intervals only changes how the
fixed physical horizon is partitioned and does not change the total amount
of time over which treatment utility is accumulated. 

\paragraph{Lagrangian constrained RL}
We further consider Lagrangian variants of PPO and TRPO to incorporate safety constraints during online policy learning. The unconstrained PPO and TRPO policies optimize the expected discounted return using the reward advantage $A^r_k$ estimated from on-policy rollouts. PPO performs this update through a clipped objective, while constrains by a KL-divergence trust region. 

Safety is enforced through an explicit per-interaction cost.
Let $c(\bm{\mathrm{x}},\bm{\mathrm{u}})$ denote the instantaneous cost of the
instantiation at hand, given in \eqref{eq:sepsis_cost} for sepsis and in
\eqref{eq:hypo_cost} for acute hypotension.
The cost of interaction $k$ is $C_k=c(\bm{\mathrm{x}}_k,\bm{\mathrm{u}}_k)$, evaluated
at the decision state, and the constraint is imposed on the average of $C_k$ over the
$K$ interactions of a rollout.
Because the interaction budget $K$ and the horizon $T$ are identical in the fixed and
the adaptive setting, both are subject to the same constraint budget, and the reported
safety quantities are in addition evaluated on a common physical-time grid over
$[0,T]$ rather than at decision times, so the comparison between timing settings does
not depend on where the interactions are placed.
The reward and cost advantages $A^r_k$ and $A^c_k$ are computed from $r_k$ and $C_k$,
respectively.

Given the Lagrange multiplier $\lambda \geq 0$, we define
$$
    A^\lambda_k
    =
    \frac{A^r_k - \lambda A^c_k}{1+\lambda}.
$$
Thus, actions with high reward advantage are encouraged, while actions that increase the expected safety cost are penalized.

For Lagrangian PPO, we use $A^\lambda_k$ in the standard PPO clipped surrogate objective:
$$
    L^{\mathrm{LagPPO}}(\theta)
    =
    \mathbb{E}_k
    \left[
        \min
        \left(
            \rho_k(\theta) A^\lambda_k,
            \mathrm{clip}
            \left(
                \rho_k(\theta),
                1-\epsilon,
                1+\epsilon
            \right)
            A^\lambda_k
        \right)
    \right],
$$
where
$$
    \rho_k(\theta)
    =
    \frac{\pi_\theta(a_k \mid x_k)}
         {\pi_{\theta_{\mathrm{old}}}(a_k \mid x_k)} .
$$

For Lagrangian TRPO, we use the same penalized advantage in the TRPO trust-region update:
$$
    \max_{\theta}
    \mathbb{E}_k
    \left[
        \rho_k(\theta) A^\lambda_k
    \right]
    \quad
    \mathrm{s.t.}
    \quad
    \mathbb{E}_k
    \left[
        D_{\mathrm{KL}}
        \left(
            \pi_{\theta_{\mathrm{old}}}(\cdot \mid x_k)
            \,\|\, 
            \pi_{\theta}(\cdot \mid x_k)
        \right)
    \right]
    \leq \delta .
$$
The multiplier is updated after each rollout batch according to the observed constraint violation:
$$
    \lambda
    \leftarrow
    \left[
        \lambda
        +
        \eta_\lambda
        \left(
            \bar{c}
            -
            c_{\mathrm{limit}}
        \right)
    \right]_+ ,
$$
where $\bar{c}$ is the mean rollout cost and $c_{\mathrm{limit}}$ is the cost threshold.

\paragraph{Detailed learning settings}
All policies are trained over the same fixed physical-time horizon
$T=96$ hours with a prescribed interaction budget $K$.
For adaptive-time policies, the policy outputs
\[
\bar{\bm{\mathrm{u}}}_k
=
(\bm{\mathrm{u}}_k,\delta t_k),
\]
where
$\delta t_k\in[\delta t_{\min},\delta t_{\max}]$.
Following the bounded-interaction formulation of
TACOS \citep{Treven_NeurIPS2024}, the selected durations satisfy
\begin{equation}
\sum_{k=0}^{K-1}\delta t_k=T,
\qquad
t_K=T.
\end{equation}
This constraint is enforced by the environment: each proposed duration is
clipped to $[\delta t_{\min},\delta t_{\max}]$ and to the remaining time
$T-t_k$, the last interaction takes the remainder, and an episode ends at
$t_k=T$ rather than when the budget is exhausted, so both timing settings
traverse the same physical horizon.
Hence, the adaptive policy allocates the available interactions across
the complete treatment horizon rather than changing the physical
duration of the episode.
For the fixed-timing baseline, the same horizon and interaction budget
are used, with equidistant intervals
$\delta t_k=T/K$.

The main difference among the algorithms is how rollout data are collected and reused.
SAC is trained as an off-policy method.
At each interaction step, the current policy generates a transition
$$
    (x_k, a_k, r_k, x_{k+1}, \mathrm{done}_k),
$$
which is stored in a replay buffer.
After the warm-up period, actor and critic networks are updated from mini-batches sampled from the replay buffer.
Thus, SAC can reuse past transitions collected by earlier policies.

In contrast, PPO and TRPO are trained as on-policy methods.
For each update, we first collect a rollout of length $L$ using the current policy.
The collected trajectory is then used to estimate returns and generalized advantages.
PPO updates the policy by optimizing the clipped surrogate objective over multiple epochs and mini-batches, whereas TRPO updates the policy by solving a trust-region step with a KL-divergence constraint.
After the update, the rollout data are discarded and new trajectories are collected with the updated policy.
The Lagrangian variants follow the same data-collection procedure as their base algorithms, but replace the reward advantage with the penalized advantage defined above and update the Lagrange multiplier using the rollout-level cost violation. 
Each online policy is trained on a single simulator: an individual-level policy
on the patient-specific PINN of its patient, a cluster-level policy on the
cluster PINN of its cluster, and a population-level policy on the population
PINN. The environment-step budget in Table~\ref{tab:hparam_rl} is therefore both
the total and the per-simulator budget, and it is identical across the three
personalization levels, so the comparison across levels is not confounded by
differences in training exposure.

\begin{table}[h]
\centering
\caption{RL training hyperparameters.}
\label{tab:hparam_rl}
\small
\begin{tabular}{llr}
\toprule
\textbf{Component} & \textbf{Hyperparameter} & \textbf{Value} \\
\midrule
\multirow{4}{*}{TACOS}
  & $\delta t_{\min}$                          & 0.5\,h               \\
  & $\delta t_{\max}$                          & 36.0\,h              \\
  & Interaction budget $K$                     & 20                   \\
  & Fixed-timing interval $T/K$                & 4.8\,h               \\
  & Total horizon $T$                          & 96\,h                \\
\midrule
\multirow{2}{*}{Shared (online)}
  & Hidden layer width                         & 256                  \\
  & Discount factor $\gamma$                   & 1                \\
\midrule
\multirow{7}{*}{SAC}
  & Total environment steps                    & 100{,}000            \\
  & Replay buffer capacity / warm-up           & 200{,}000 / 20{,}000 \\
  & Batch size                                 & 512                  \\
  & Learning rate (actor / critic / $\alpha$)  & $1\times10^{-4}$     \\
  & Soft-update rate $\tau$                    & 0.002                \\
  & Entropy temperature $\alpha$               & auto-tuned           \\
  & Target entropy                             & $-\dim(\mathcal{A})$ \\
\midrule
\multirow{9}{*}{PPO / PPO-Lagrangian}
  & Total environment steps                    & 200{,}000            \\
  & Rollout length                             & 2{,}048              \\
  & Update epochs per rollout                  & 10                   \\
  & Mini-batch size                            & 64                   \\
  & Actor / critic learning rate               & $3\times10^{-4}$ / $1\times10^{-3}$ \\
  & GAE $\lambda$                              & 0.95                 \\
  & Clip range $\epsilon$                      & 0.2                  \\
  & Entropy coefficient                        & 0.01                 \\
  & Cost limit $c_{\mathrm{limit}}$ (mean $C_k$) / lr $\lambda$ & 0.1 / $5\times10^{-2}$ \\
\midrule
\multirow{9}{*}{TRPO / TRPO-Lagrangian}
  & Total environment steps                    & 200{,}000            \\
  & Rollout length                             & 2{,}048              \\
  & Trust-region radius $\delta$               & 0.01                 \\
  & Critic learning rate                       & $1\times10^{-3}$     \\
  & Value-function update steps                & 5                    \\
  & Conjugate gradient iterations              & 10                   \\
  & Line search (max back-tracks, coeff.)      & 10, 0.5              \\
  & GAE $\lambda$                              & 0.95                 \\
  & Cost limit $c_{\mathrm{limit}}$ (mean $C_k$) / lr $\lambda$ & 0.1 / $5\times10^{-2}$ \\
  \midrule
\multirow{3}{*}{Offline data}
  & Behavior policy                            & TRPO-Lagrangian (fixed timing) \\
  & Rollout episodes per patient $S$            & 5                    \\
  & Action perturbation $\sigma_{\mathbf{u}}$    & 0.05                 \\
\midrule
\multirow{5}{*}{DQN / CQL / GCQL}
  & Gradient steps                              & 4{,}000              \\
  & Batch size                                  & 256                  \\
  & Actor / critic learning rate                & $3\times10^{-5}$     \\
  & Soft-update rate $\tau$                     & 0.001                \\
  & Discount factor $\gamma$                    & 0.99                 \\
\midrule
\multirow{2}{*}{CQL / GCQL}
  & Conservative coefficient $\alpha$           & 5.0                  \\
  & Candidate actions / sampling s.d.           & 12 / 0.1             \\
\midrule
\multirow{3}{*}{GCQL}
  & Support radius multiplier                   & 0.5                  \\
  & Guard weight  $\beta$ (critic / actor)               & 0.5 / 2.0            \\
  & Guard sigmoid sharpness                     & 20                   \\
\bottomrule

\end{tabular}
\end{table}

\subsection{Offline Policy Learning}
\label{appendix:offline_policy_learning}

This section describes the offline policy learning protocol used in MedGym.
Unlike the online policy learning methods in Appendix C.1, offline reinforcement learning does not interact with the simulator during training.
Instead, the policy is trained from a fixed dataset of transition tuples generated before policy optimization.
After training, the learned offline policy is deployed in MedGym and evaluated by closed-loop rollouts.
This follows the benchmark pipeline in Figure~2: MedGym is first used to generate offline training data, the offline RL algorithm learns a policy from the fixed dataset, and the learned policy is then evaluated online in the same continuous-time patient simulators.

Throughout the offline evaluation, both the state space
$\mathcal{X}\subseteq\mathbb{R}^{d_{\mathsf{s}}}$ and the treatment
action space $\mathcal{U}\subseteq\mathbb{R}^{d_{\mathsf{a}}}$ are
continuous, consistent with the environment definition in Section~\ref{sec:prior_works}.
Accordingly, action maximization and conservative regularization below
are defined over the continuous action space; in implementation, the
corresponding continuous-action operations are evaluated numerically
using sampled candidate actions from the admissible action domain.

\paragraph{MedGym environments.}
As stated in the main text, the offline setting uses a single cohort of $110$
patient-specific PINN simulators constructed in Appendix~B, partitioned into $10$
dynamics-based clusters.
For each cluster $k \in \{1,\ldots,10\}$, one patient, denoted by
$j_k^{\mathrm{test}}$, is held out for testing, and the remaining ten patients
form the training subset $\mathcal{C}_k^{\mathrm{train}}$, giving $100$ training
patients and $10$ held-out test patients.
The patient-specific MedGym environments remain the basic environments
for generating offline trajectories and evaluating learned policies.
Cluster-level and population-level offline datasets are constructed only
from the datasets of the training patients.

\paragraph{Patient-specific offline dataset.}
For each patient $i \in \{1,\dots,n\}$, we train a fixed-interval
Lagrangian TRPO behavior policy
$\pi^{(i)}_{\mathrm{LagrangianTRPO}}$ on the corresponding PINN
$F^{\mathrm{nn}}_{\xi_i}(\mathbf{x},\mathbf{u},\delta t)$,
using the online training protocol in
Appendix~\ref{appendix:online_policy_learning} with
$\delta t_k=T/K$.
Still on the same PINN $F^{\mathrm{nn}}_{\xi_i}(\mathbf x, \mathbf u, \delta t)$, we generate an offline dataset, consisting of $S$ rollout episodes,
\begin{align*}
    \mathcal{D}^{(i)}_\text{LagrangianTRPO}
    :=
    \left( \tau^{(i)}_s \right)_{s=1}^S,
    \quad \text{where} \quad
    \tau^{(i)}_s
    :=
    \left(
        \mathbf x^{(i)}_{s,0},
        \mathbf u^{(i)}_{s,0},
        r^{(i)}_{s,0},
        \mathbf x^{(i)}_{s,1},
        \dots,
        \mathbf x^{(i)}_{s,H_i^{(s)}}
    \right).
\end{align*}
To increase the diversity of the offline data and avoid generating a nearly deterministic dataset, we add small Gaussian perturbations to the behavior action and project the
result back onto the admissible action set, via
\begin{align}
    \mathbf u^{(i)}_{s,k}
    & =
    \Pi_{\mathcal U}
    \left(
      \mathbf u^{(i)}_k + \epsilon^{(i)}_{s,\mathbf u}
    \right),
    &
    \epsilon^{(i)}_{s,\mathbf u}
    & \sim
    \mathcal{N}(0,\sigma_{\mathbf u}^2 I),
    &
    \mathbf u^{(i)}_k
    =
    \pi^{(i)}_\text{LagrangianTRPO}(\mathbf x^{(i)}_{s,k}),
    \label{eq:perturbed_rollouts}
\end{align}
where $\Pi_{\mathcal U}$ denotes the projection onto the admissible action set and
$\sigma_{\mathbf u}=0.05$ in the normalized action space.
Every rollout starts from the patient's recorded initial condition, which is not
perturbed.
Only the patient-specific datasets associated with
$i \in \mathcal{I}_{\mathrm{train}}$ are used to construct the cluster-level
and population-level offline training datasets.
The dataset of each held-out test patient is used only to train the
corresponding individual oracle policy and is not included in cluster-level
or population-level policy training.

\paragraph{Deep Q-Network.}
The \emph{Deep Q-Network (DQN)} \citep{mnih2013playing} approximates the
action-value function by $Q_\theta:\mathcal{X}\times\mathcal{U}\rightarrow
\mathbb{R}$. For the continuous treatment action space, we write the
Bellman regression objective as
\begin{align}
\min_{\theta\in\Theta}
J_{\mathrm{DQN}}(\theta)
:=
\frac{1}{2}
\mathbb{E}_{(\mathbf{x},\mathbf{u},r,\mathbf{x}')
\sim\mathcal{D}}
\left[
r
+
\gamma
\sup_{\mathbf{u}'\in\mathcal{U}}
Q_\theta(\mathbf{x}',\mathbf{u}')
-
Q_\theta(\mathbf{x},\mathbf{u})
\right]^2 .
\end{align}

\paragraph{Conservative Q-Learning.}
Directly applying value-based learning to a fixed offline dataset may
assign erroneously high values to treatment actions insufficiently
supported by the behavior data.
CQL \citep{Kumar} mitigates this extrapolation error by penalizing high
values over the continuous action domain:
\begin{align}
\min_{\theta\in\Theta}
J_{\mathrm{CQL}}(\theta)
:=
J_{\mathrm{DQN}}(\theta)
+
\alpha
\Bigg(
\mathbb{E}_{\mathbf{x}\sim\mathcal{D}}
\left[
\log
\int_{\mathcal{U}}
\exp\!\left(
Q_\theta(\mathbf{x},\mathbf{u})
\right)
\,\mathrm{d}\nu(\mathbf{u})
\right]
-
\mathbb{E}_{(\mathbf{x},\mathbf{u})\sim\mathcal{D}}
\left[
Q_\theta(\mathbf{x},\mathbf{u})
\right]
\Bigg),
\end{align}
where $\nu$ denotes a reference distribution over the admissible
continuous action space. The integral is approximated by Monte Carlo
sampling in implementation.

\paragraph{Guarded CQL.}
Although CQL penalizes actions that may induce out-of-distribution (OOD)
behavior, the learned policy may still assign high values to
state--action pairs that are weakly supported by the offline dataset.
Guarded CQL (GCQL) \citep{Shen_NeurIPS2025,Tumay2025GuardianRL}
addresses this issue by introducing an additional support-aware
regularization mechanism.

Rather than explicitly estimating the density of the offline data, we use
the distance of a state--action pair from the center of the dataset as a
computationally simple and conceptually effective surrogate for data
support. This approximation does not reproduce the original density-based
criterion exactly, but retains its main purpose: state--action pairs
located farther from the region occupied by the behavior data are treated
as less supported and are therefore penalized more strongly.
Let
\[
\bm{\mathrm{z}}
=
(\bm{\mathrm{x}},\bm{\mathrm{u}})
\in
\mathcal{X}\times\mathcal{U},
\]
and define the dataset centroid as
\[
\overline{\bm{\mathrm{z}}}
=
\frac{1}{|\mathcal{D}|}
\sum_{\bm{\mathrm{z}}\in\mathcal{D}}
\bm{\mathrm{z}},
\]
together with the support threshold
\[
d_{\mathrm{lim}}
=
\tau
\max_{\bm{\mathrm{z}}\in\mathcal{D}}
\left\|
\bm{\mathrm{z}}-\overline{\bm{\mathrm{z}}}
\right\|_2,
\]
where $\tau=0.5$ in our experiments.
We define the corresponding guard indicator as
\[
g(\bm{\mathrm{z}})
=
\mathbf{1}
\left\{
\left\|
\bm{\mathrm{z}}-\overline{\bm{\mathrm{z}}}
\right\|_2
>
d_{\mathrm{lim}}
\right\},
\]
which identifies state--action pairs lying outside the prescribed
data-support region. In implementation the indicator is replaced by a sigmoid relaxation of the same threshold with sharpness $20$, and the guard is applied to the critic and the actor objectives with separate weights, $\beta=0.5$ and $\beta=2.0$. The resulting guarded objective takes the form
\begin{align}
\min_{\theta\in\Theta}
J_{\mathrm{GCQL}}(\theta)
=
J_{\mathrm{CQL}}(\theta)
+
\beta\,
\mathcal{R}_{\mathrm{guard}}(\theta).
\end{align}
We define the guarded regularizer as
\begin{equation}
\mathcal{R}_{\mathrm{guard}}(\theta)
=
\mathbb{E}_{
\bm{\mathrm{x}}\sim\mathcal{D},
\,\bm{\mathrm{u}}\sim\nu
}
\left[
g(\bm{\mathrm{x}},\bm{\mathrm{u}})
Q_\theta(\bm{\mathrm{x}},\bm{\mathrm{u}})
\right],
\end{equation}
where $\nu$ is the same reference distribution over admissible
continuous actions used in the CQL regularizer, and
$g(\bm{\mathrm{x}},\bm{\mathrm{u}})$ is the guard indicator defined
above. Thus, state--action pairs identified as weakly supported receive
an additional $Q_\theta$-dependent conservative penalty.

\paragraph{Patient-specific training.}
Given the fixed dataset $\mathcal{D}^{(i)}_\text{LagrangianTRPO}$, we train patient-specific DQN, CQL, and GCQL policies $\pi^{(i)}_\text{DQN}$, $\pi^{(i)}_\text{CQL}$, $\pi^{(i)}_\text{GCQL}$, respectively.

\paragraph{Patient-specific evaluation.}
After training, each $\pi^{(i)}_\text{DQN}$, $\pi^{(i)}_\text{CQL}$, and $\pi^{(i)}_\text{GCQL}$ is evaluated by closed-loop rollouts on the same PINN $F^{\mathrm{nn}}_{\xi_i}(\mathbf x, \mathbf u, \delta t)$ that they were trained on.
Similarly to \eqref{eq:perturbed_rollouts}, we produce the datasets $\mathcal{D}^{(i)}_\text{DQN}$, $\mathcal{D}^{(i)}_\text{CQL}$, and $\mathcal{D}^{(i)}_\text{GCQL}$. 
For each patient $i$, we compute treatment effectiveness and safety metrics.
The main effectiveness metrics include cumulative reward, mean SOFA, and SOFA at the terminal horizon.
The main safety metrics include mean lactate, lactate trajectory, and the lactate safety rate after the critical time threshold.
This gives one patient-specific DQN, CQL, and GCQL policy for each of the $10$
held-out test patients $\{j_k^{\mathrm{test}}\}_{k=1}^{10}$, each trained on that
patient's own offline dataset and evaluated on that patient's own simulator
$F^{\mathrm{nn}}_{\xi_{j_k^{\mathrm{test}}}}$

\paragraph{Cluster-level offline dataset, training and evaluation.}
For each cluster $k$, we construct a cluster-level offline dataset by
aggregating the patient-specific datasets of its training patients,
\begin{align*}
    \mathcal{D}^{(k)}_\text{clu}
    =
    \bigcup_{i \in \mathcal{C}_k^{\mathrm{train}}}
        \mathcal{D}^{(i)}_\text{LagrangianTRPO}.
\end{align*}
Using $\mathcal{D}^{(k)}_\text{clu}$, we train cluster-level policies
$\pi^{(k)}_\mathrm{DQN}$, $\pi^{(k)}_\mathrm{CQL}$ and
$\pi^{(k)}_\mathrm{GCQL}$ shared across the patients of cluster $k$.
Each cluster-level policy is then evaluated on the simulator of the held-out
test patient of the same cluster, i.e.\ on
$F^{\mathrm{nn}}_{\xi_{j_k^{\mathrm{test}}}}$ --- the same environment on which
the individual- and population-level policies of that cluster are evaluated ---
producing the datasets $\mathcal{D}^{(k)}_\text{DQN}$,
$\mathcal{D}^{(k)}_\text{CQL}$ and $\mathcal{D}^{(k)}_\text{GCQL}$.

\paragraph{Population wide offline dataset, training and evaluation.}
In addition to patient-specific offline learning, we construct a population-level offline dataset by aggregating all patient-specific datasets,
\begin{align*}
    \mathcal{D}^\text{pop}
    =
    \bigcup_{i=1}^{100}
        \mathcal{D}^{(i)}_\text{LagrangianTRPO}.
\end{align*}
Using $\mathcal{D}^{\mathrm{pop}}$, we train single population-level policies
$\pi^\text{pop}_\mathrm{DQN}$, $\pi^\text{pop}_\mathrm{CQL}$ and
$\pi^\text{pop}_\mathrm{GCQL}$ shared across all training patients.
These are then evaluated separately on each of the $10$ held-out
patient-specific environments
$\{F^{\mathrm{nn}}_{\xi_{j_k^{\mathrm{test}}}}\}_{k=1}^{10}$, i.e.\ on exactly
the same environments used to evaluate the individual- and cluster-level
policies, producing the datasets $\mathcal{D}^\text{pop}_\text{DQN}$,
$\mathcal{D}^\text{pop}_\text{CQL}$ and $\mathcal{D}^\text{pop}_\text{GCQL}$.

\paragraph{Evaluation question.}
This offline protocol is designed to answer the following question: how does
offline policy performance change as the training data become increasingly
patient-specific?
For each held-out test patient $j_k^{\mathrm{test}}$, the three policy levels
$\pi^{(j_k^{\mathrm{test}})}_{\diamond}$, $\pi^{(k)}_{\diamond}$ and
$\pi^\text{pop}_{\diamond}$ are compared on the same simulator
$F^{\mathrm{nn}}_{\xi_{j_k^{\mathrm{test}}}}$, where $\diamond$ can be DQN, CQL,
or GCQL.
They differ only in the offline dataset they are trained on: that patient's own
trajectories, the trajectories of its cluster's training patients, or the
trajectories of all training patients.
If $\pi^{(j_k^{\mathrm{test}})}_{\diamond}$ consistently outperforms
$\pi^{(k)}_{\diamond}$ and $\pi^\text{pop}_{\diamond}$, this indicates that
offline policy learning benefits from patient-specific behavior data.
If the cluster-level policy recovers part of that gap, the dynamics-based
grouping transfers useful treatment structure across similar patients.
If $\pi^\text{pop}_{\diamond}$ performs comparably, the aggregated offline
dataset already captures enough shared treatment structure to support
population-level generalization.

\section{Details of Evaluation Results}
\label{appendix:evaluation_results}

\subsection{Online Policy Methods Evaluations}
\label{appendix:online_policy_evaluations}

\begin{figure}[t!]
  \centering
  \captionsetup[subfigure]{skip=0pt}

  \begin{subfigure}[t]{0.24\linewidth}
    \centering
    \includegraphics[width=\linewidth]
    { 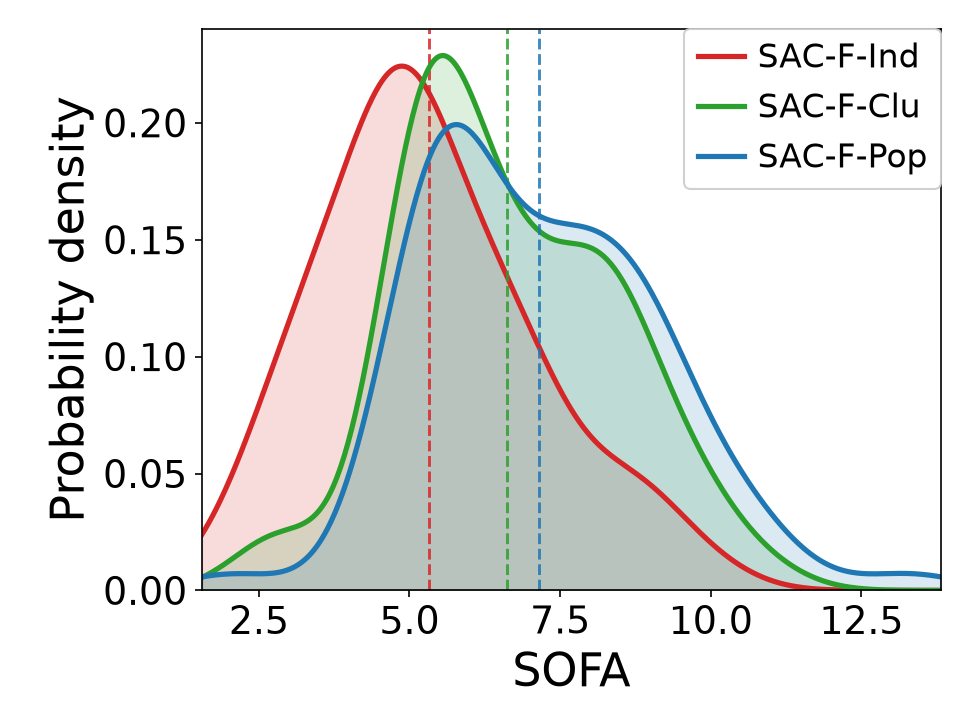}
    \caption{$\mathsf{SAC}$-F}
    \label{fig:sac_fixed_dist_appendix}
  \end{subfigure}\hfill
  \begin{subfigure}[t]{0.24\linewidth}
    \centering
    \includegraphics[width=\linewidth]
    { 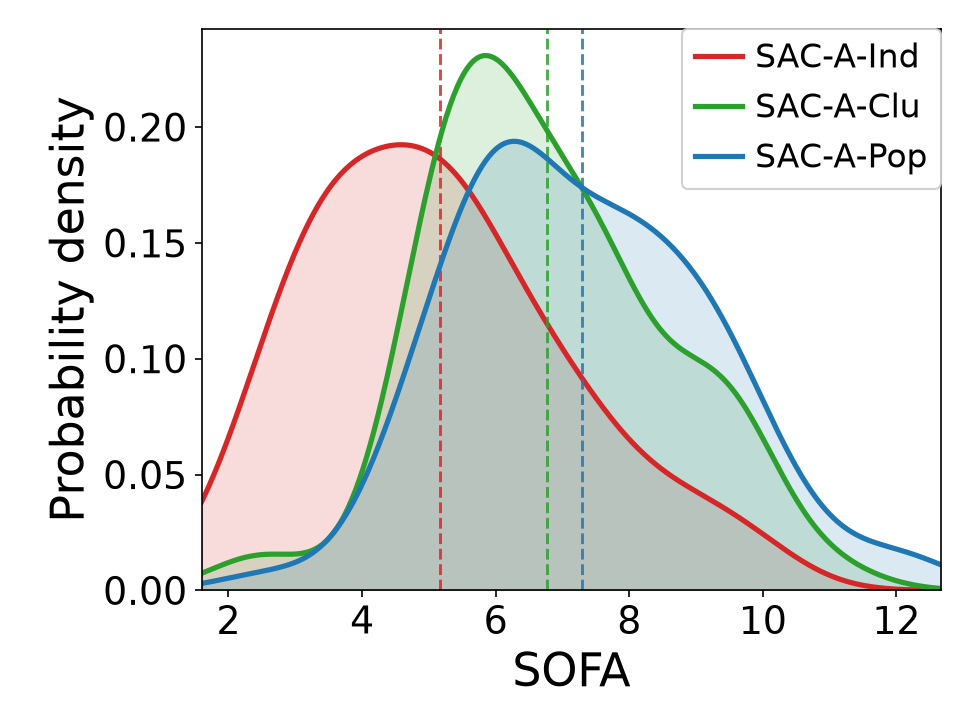}
    \caption{$\mathsf{SAC}$-A}
    \label{fig:sac_adaptive_dist_appendix}
  \end{subfigure}\hfill
  \begin{subfigure}[t]{0.24\linewidth}
    \centering
    \includegraphics[width=\linewidth]
    { 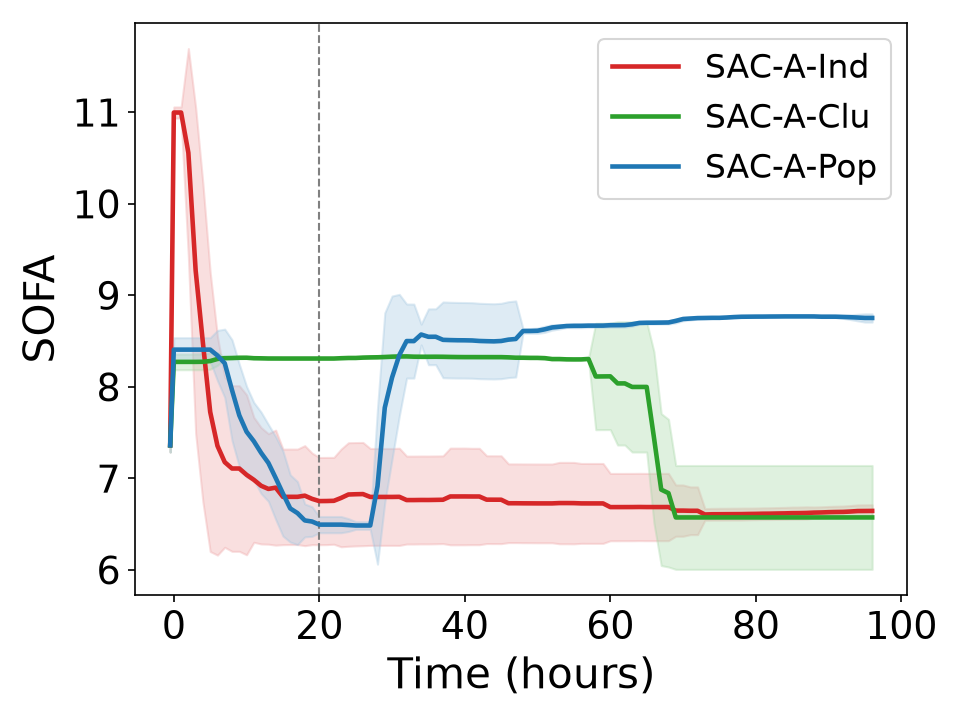}
    \caption{$\mathsf{SAC}$ SOFA}
    \label{fig:sac_sofa_appendix}
  \end{subfigure}\hfill
  \begin{subfigure}[t]{0.24\linewidth}
    \centering
    \includegraphics[width=\linewidth]
    { 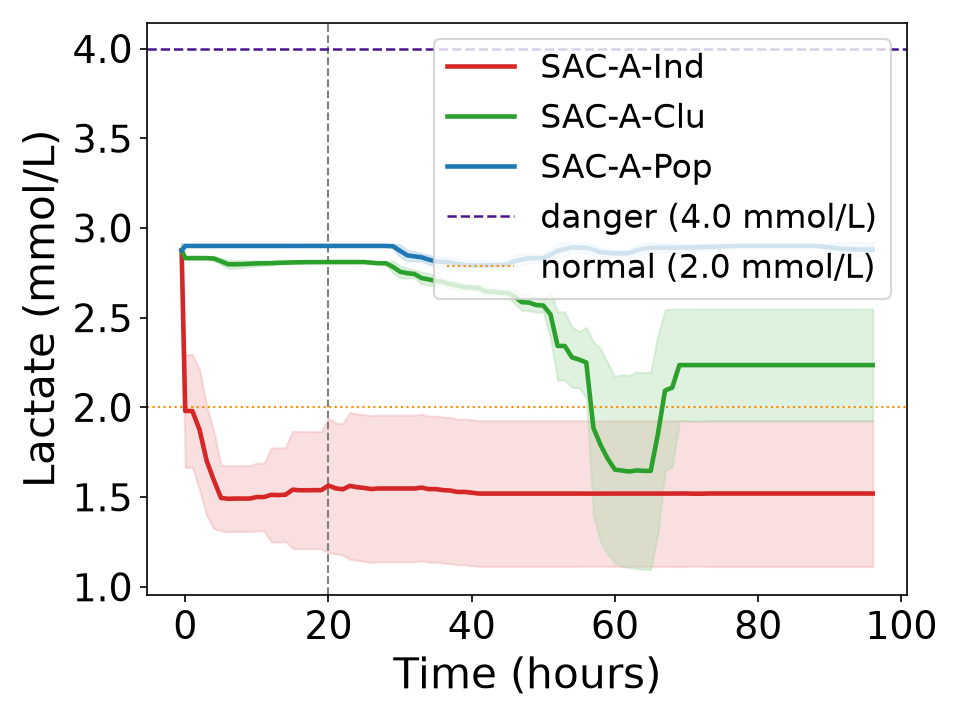}
    \caption{$\mathsf{SAC}$ Lactate}
    \label{fig:sac_lactate_appendix}
  \end{subfigure}

  \par\vspace{-0.2em}

  \begin{subfigure}[t]{0.24\linewidth}
    \centering
    \includegraphics[width=\linewidth]
    { 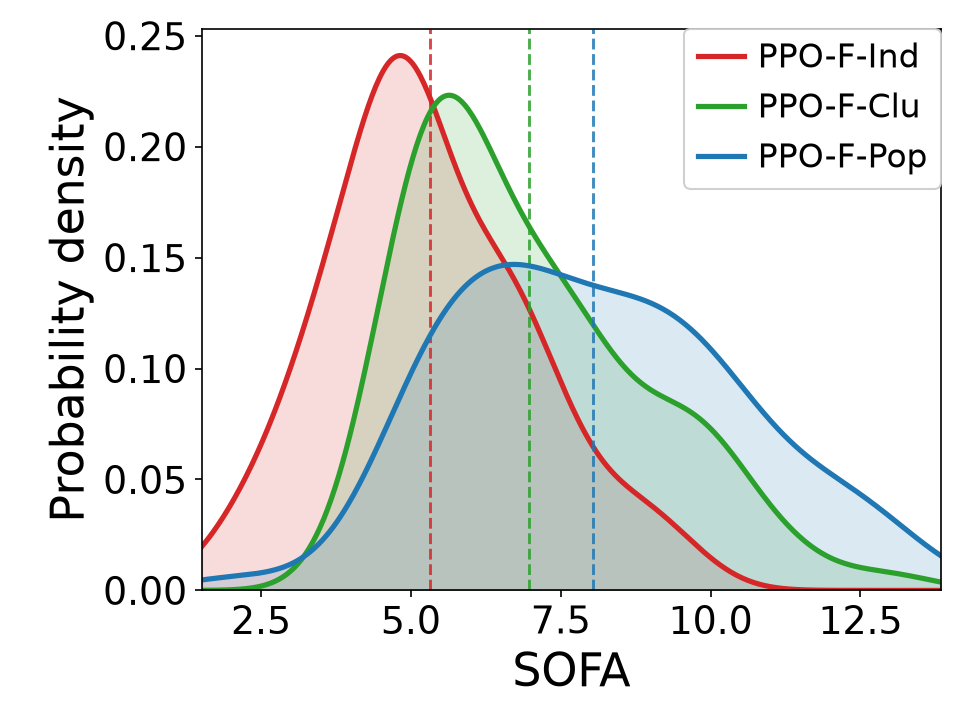}
    \caption{$\mathsf{PPO}$-F}
    \label{fig:ppo_fixed_dist_appendix}
  \end{subfigure}\hfill
  \begin{subfigure}[t]{0.24\linewidth}
    \centering
    \includegraphics[width=\linewidth]
    { 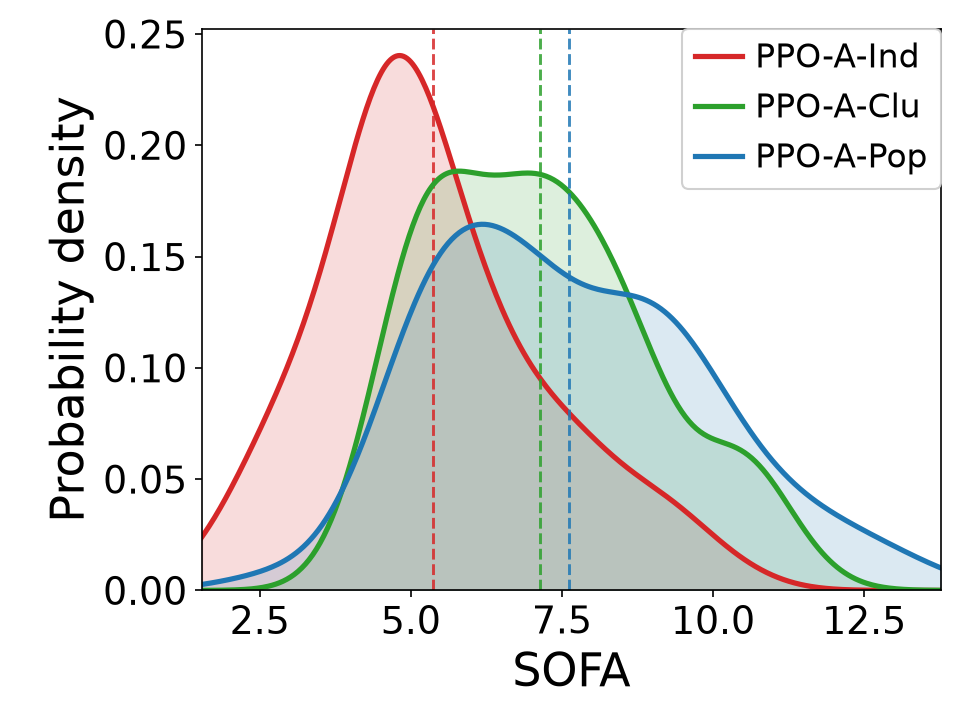}
    \caption{$\mathsf{PPO}$-A}
    \label{fig:ppo_adaptive_dist_appendix}
  \end{subfigure}\hfill
  \begin{subfigure}[t]{0.24\linewidth}
    \centering
    \includegraphics[width=\linewidth]
    { 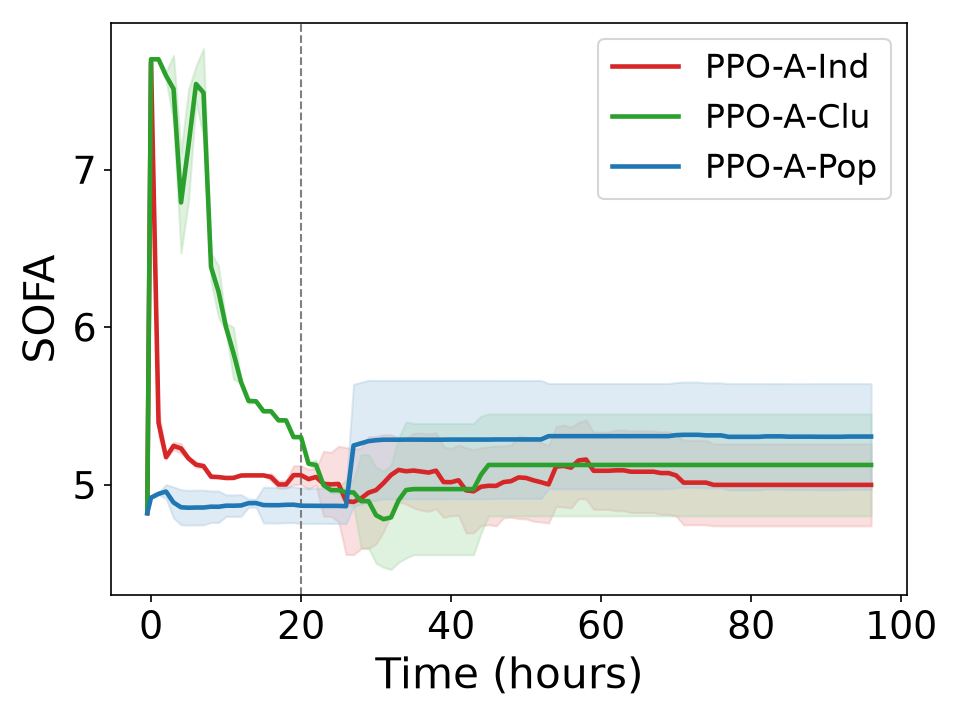}
    \caption{$\mathsf{PPO}$ SOFA}
    \label{fig:ppo_sofa_appendix}
  \end{subfigure}\hfill
  \begin{subfigure}[t]{0.24\linewidth}
    \centering
    \includegraphics[width=\linewidth]
    { 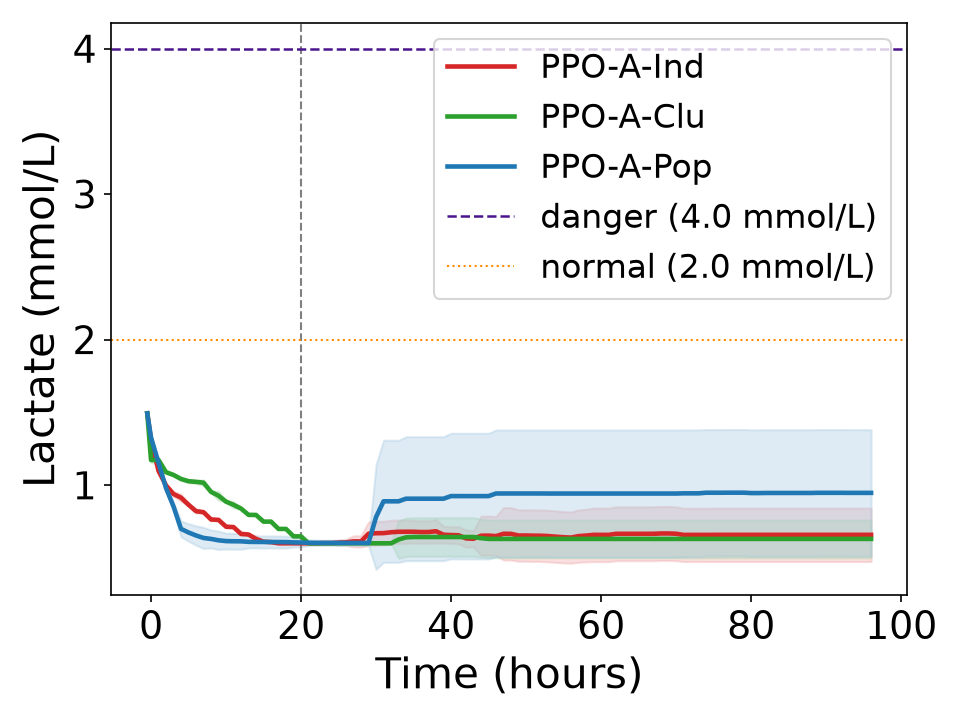}
    \caption{$\mathsf{PPO}$ Lactate}
    \label{fig:ppo_lactate_appendix}
  \end{subfigure}

  \par\vspace{-0.2em}

  \begin{subfigure}[t]{0.24\linewidth}
    \centering
    \includegraphics[width=\linewidth]
    { 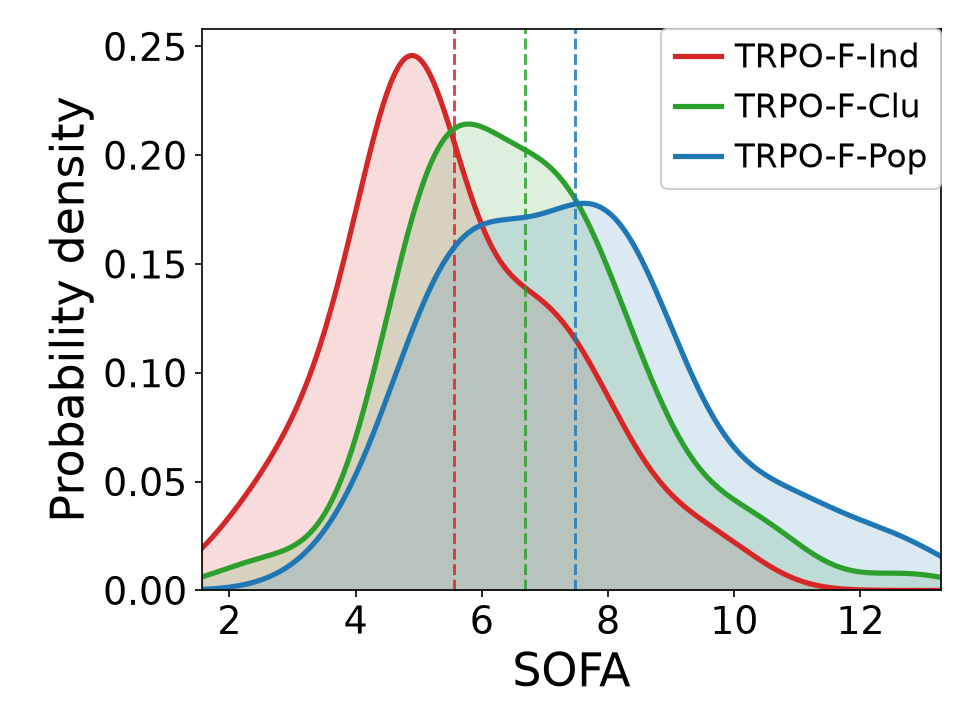}
    \caption{$\mathsf{TRPO}$-F}
    \label{fig:trpo_fixed_dist_appendix}
  \end{subfigure}\hfill
  \begin{subfigure}[t]{0.24\linewidth}
    \centering
    \includegraphics[width=\linewidth]
    { 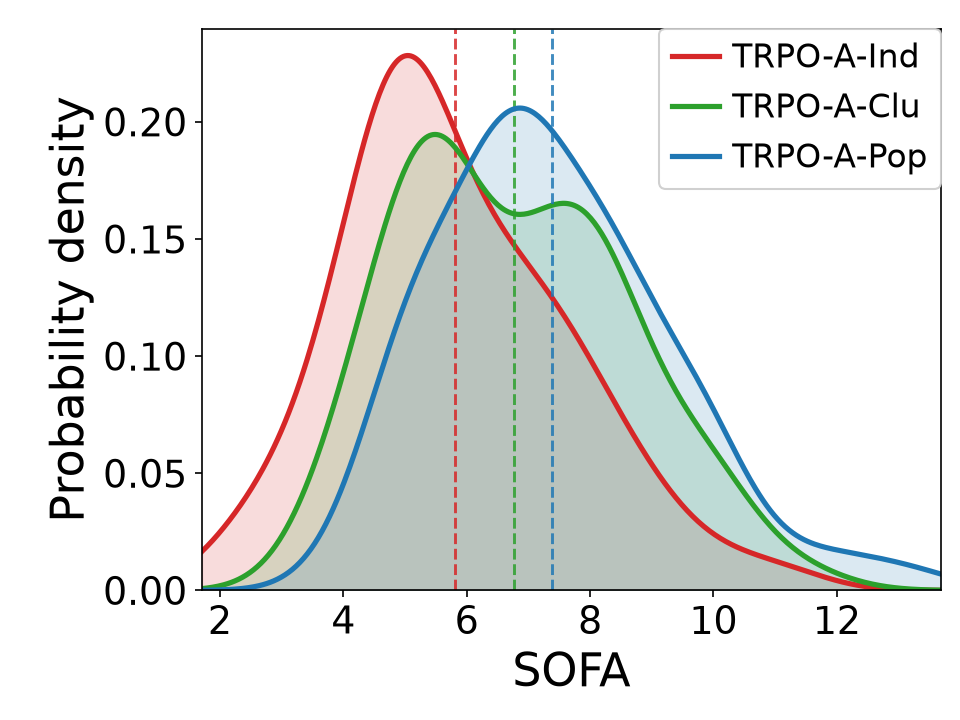}
    \caption{$\mathsf{TRPO}$-A}
    \label{fig:trpo_adaptive_dist_appendix}
  \end{subfigure}\hfill
  \begin{subfigure}[t]{0.24\linewidth}
    \centering
    \includegraphics[width=\linewidth]
    { 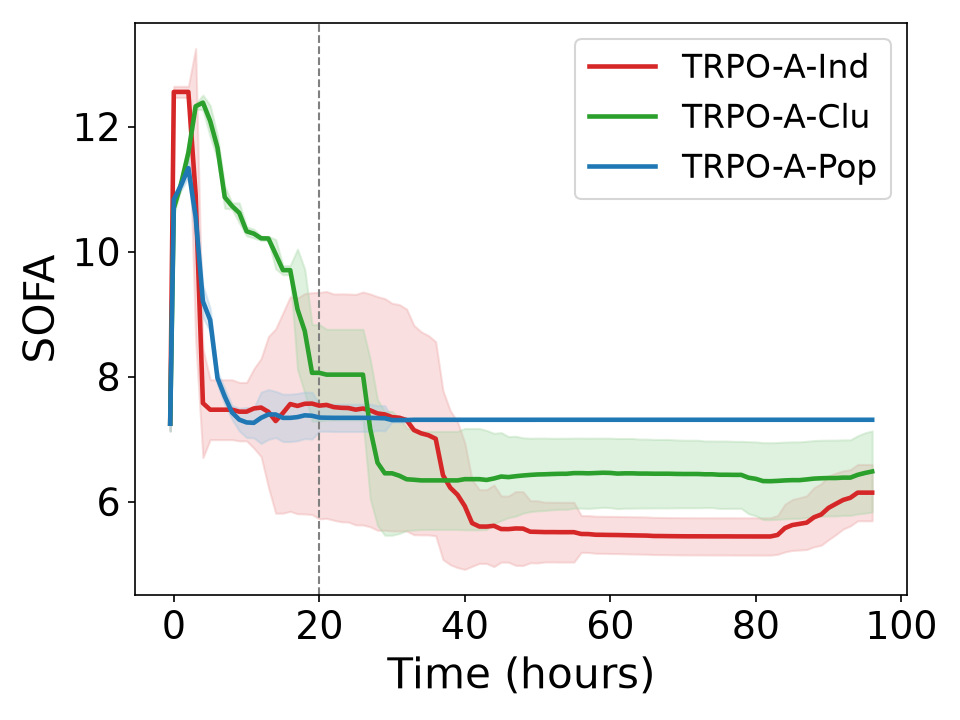}
    \caption{$\mathsf{TRPO}$ SOFA}
    \label{fig:trpo_sofa_appendix}
  \end{subfigure}\hfill
  \begin{subfigure}[t]{0.24\linewidth}
    \centering
    \includegraphics[width=\linewidth]
    { 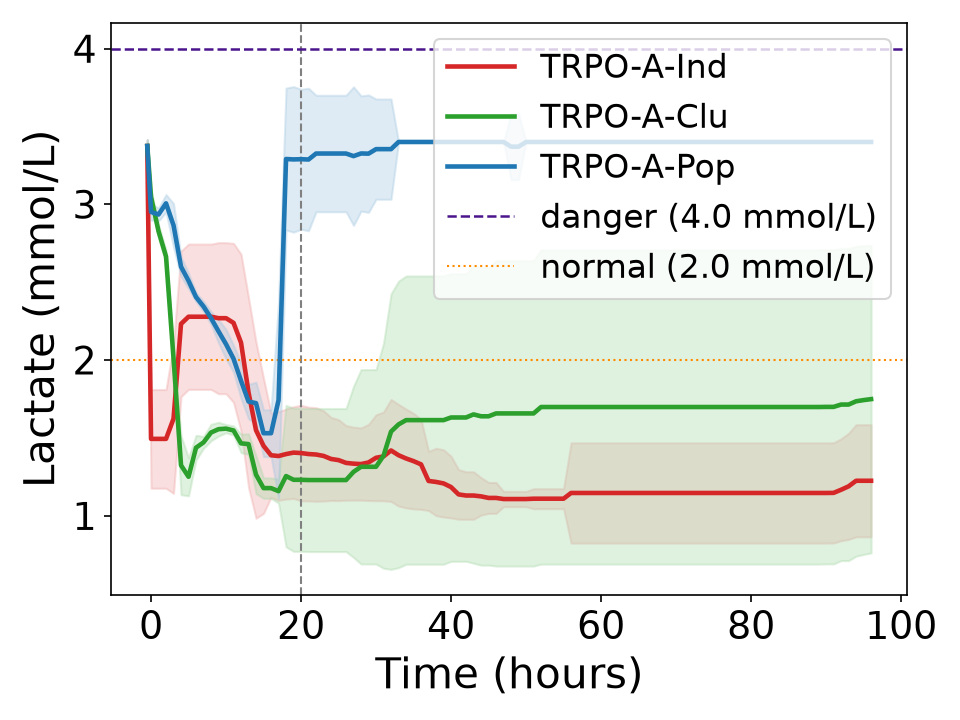}
    \caption{$\mathsf{TRPO}$ Lactate}
    \label{fig:trpo_lactate_appendix}
  \end{subfigure}

  \par\vspace{-0.2em}

  \begin{subfigure}[t]{0.24\linewidth}
    \centering
    \includegraphics[width=\linewidth]
    { 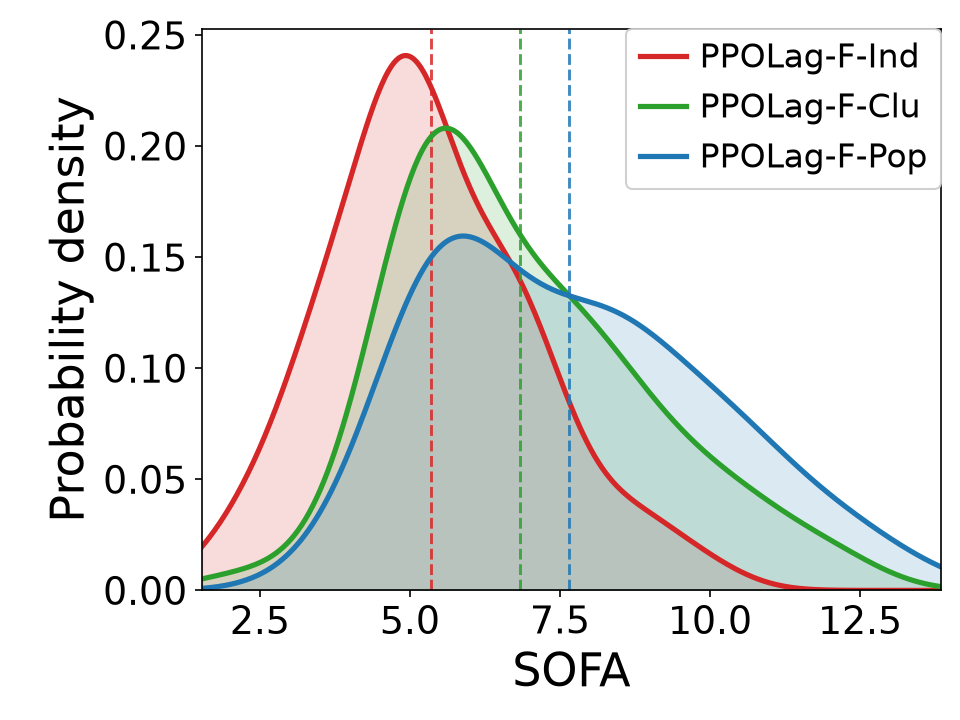}
    \caption{$\mathsf{PPOLag}$-F}
    \label{fig:ppolag_fixed_dist_appendix}
  \end{subfigure}\hfill
  \begin{subfigure}[t]{0.24\linewidth}
    \centering
    \includegraphics[width=\linewidth]
    { 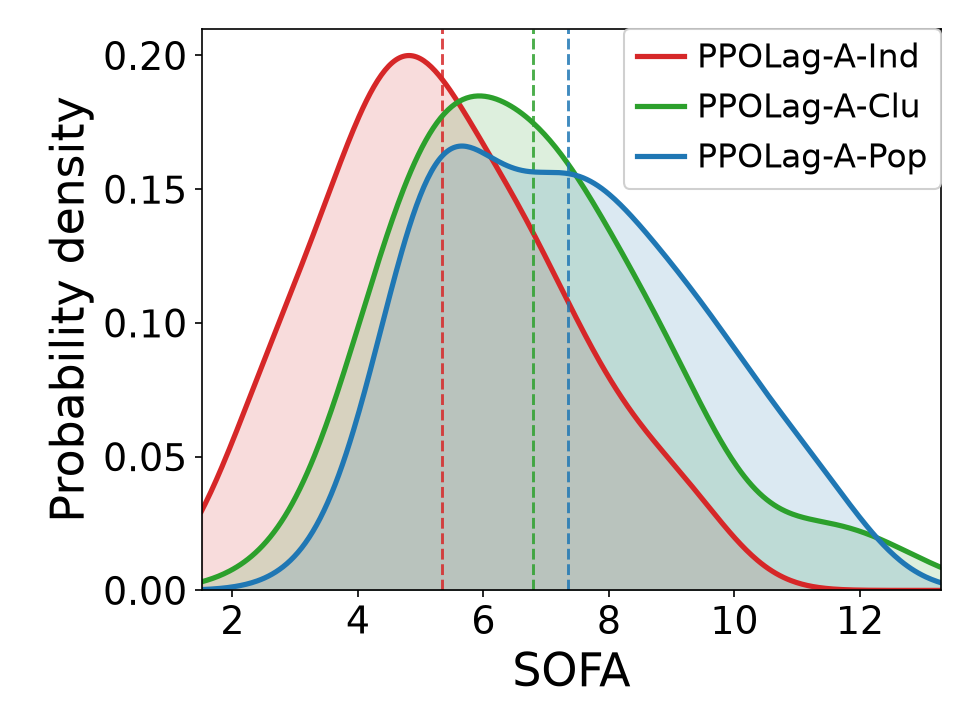}
    \caption{$\mathsf{PPOLag}$-A}
    \label{fig:ppolag_adaptive_dist_appendix}
  \end{subfigure}\hfill
  \begin{subfigure}[t]{0.24\linewidth}
    \centering
    \includegraphics[width=\linewidth]
    { 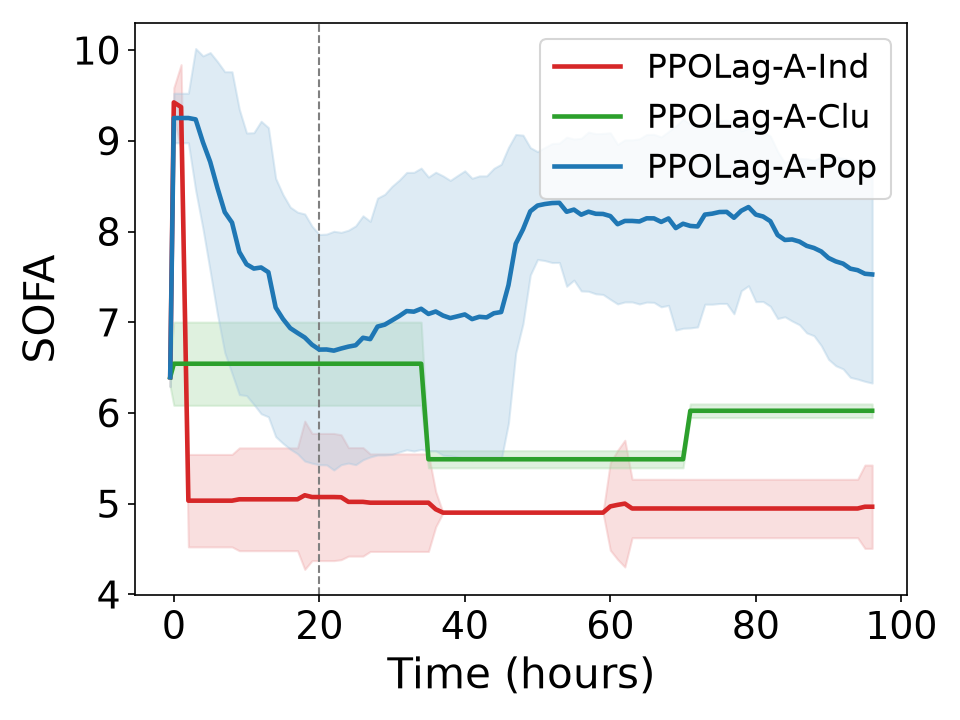}
    \caption{$\mathsf{PPOLag}$ SOFA}
    \label{fig:ppolag_sofa_appendix}
  \end{subfigure}\hfill
  \begin{subfigure}[t]{0.24\linewidth}
    \centering
    \includegraphics[width=\linewidth]
    { 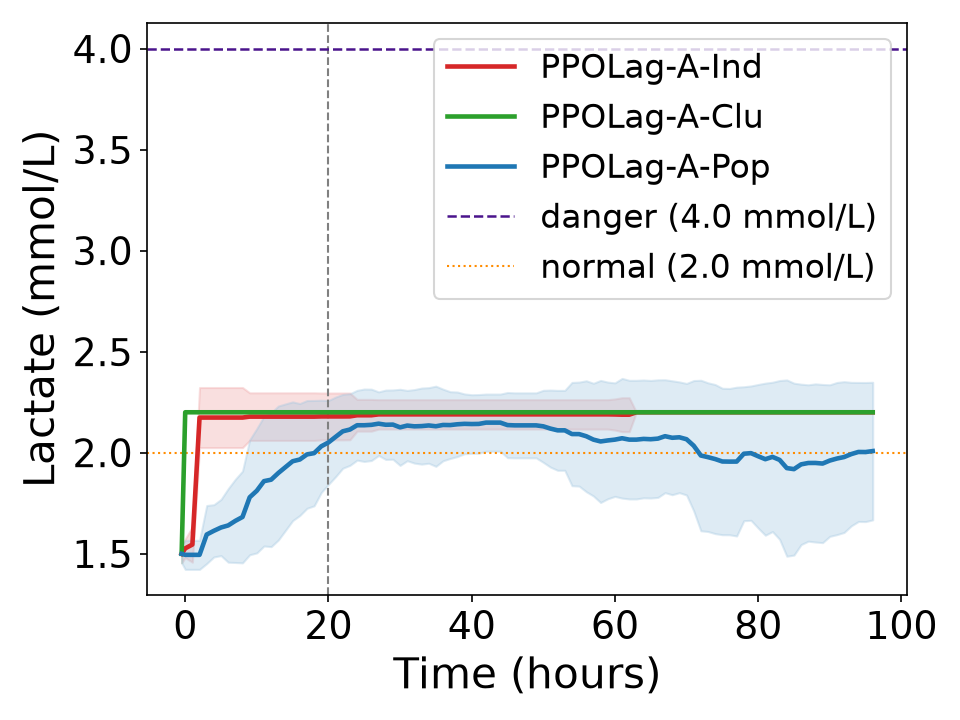}
    \caption{$\mathsf{PPOLag}$ Lactate}
    \label{fig:ppolag_lactate_appendix}
  \end{subfigure}

    \caption{
      Additional online evaluations of $\mathsf{SAC}$, $\mathsf{PPO}$,
      $\mathsf{TRPO}$, and $\mathsf{PPOLag}$. Each row corresponds to one
      algorithm; the columns show the final $\mathsf{SOFA}$ distributions
      under fixed and adaptive timing, followed by representative
      adaptive-timing $\mathsf{SOFA}$ and Lactate trajectories.
    }
  \label{fig:online_policy_evaluation_appendix}
\end{figure}

This subsection complements the main-text evaluation of
$\mathsf{TRPOLag}$ by reporting additional online results for
$\mathsf{SAC}$, $\mathsf{PPO}$, $\mathsf{TRPO}$, and
$\mathsf{PPOLag}$. Figure~\ref{fig:online_policy_evaluation_appendix}
presents the corresponding final $\mathsf{SOFA}$ distributions and
representative physiological trajectories, while
Table~\ref{tab:policy_evaluation_all_appendix} summarizes the
quantitative results. All methods use the same 70 held-out patients,
individual/cluster/population policy classes, and fixed/adaptive timing
settings as the main-text evaluation. The results exhibit the same
overall personalization pattern: the individual oracle generally
achieves the lowest $\mathsf{SOFA}$ scores, while the cluster policies
recover part of this benefit relative to the population policies.

\begin{table}[t!]
  \centering

  \fontsize{8pt}{8pt}\selectfont
  \setlength{\tabcolsep}{4pt}
  \renewcommand{\arraystretch}{0.82}
  \setlength{\aboverulesep}{0.3ex}
  \setlength{\belowrulesep}{0.3ex}
  \captionsetup{skip=2pt}

  \begin{tabular}{lcccc}
    \toprule
    Policy
    & SOFA(all) ($\downarrow$)
    & Lower SOFA
        vs. Pop (\%)
    & SOFA(96h) ($\downarrow$)
    & SafeTime\% ($\uparrow$) \\
    \midrule

    $\mathsf{SAC}$-$\mathsf{F}$-$\mathsf{Ind}$
    & $5.97 \pm 1.67$
    & $97.1$
    & $5.32 \pm 1.77$
    & $\mathbf{91.95 \pm 22.83}$ \\

    $\mathsf{SAC}$-$\mathsf{A}$-$\mathsf{Ind}$
    & $\mathbf{5.39 \pm 1.89}$
    & $100.0$
    & $\mathbf{5.16 \pm 1.90}$
    & $90.34 \pm 24.14$ \\

    $\mathsf{SAC}$-$\mathsf{F}$-$\mathsf{Clu}$
    & $7.05 \pm 1.74$
    & $70.0$
    & $6.62 \pm 1.75$
    & $90.87 \pm 24.90$ \\

    $\mathsf{SAC}$-$\mathsf{A}$-$\mathsf{Clu}$
    & $6.86 \pm 1.62$
    & $72.9$
    & $6.76 \pm 1.73$
    & $89.98 \pm 23.86$ \\

    $\mathsf{SAC}$-$\mathsf{F}$-$\mathsf{Pop}$
    & $7.31 \pm 1.87$
    & --
    & $7.15 \pm 1.92$
    & $90.48 \pm 26.80$ \\

    $\mathsf{SAC}$-$\mathsf{A}$-$\mathsf{Pop}$
    & $7.20 \pm 1.81$
    & --
    & $7.29 \pm 1.86$
    & $90.71 \pm 22.54$ \\

    \midrule

    $\mathsf{PPO}$-$\mathsf{F}$-$\mathsf{Ind}$
    & $5.75 \pm 1.66$
    & $100.0$
    & $\mathbf{5.32 \pm 1.67}$
    & $91.08 \pm 23.76$ \\

    $\mathsf{PPO}$-$\mathsf{A}$-$\mathsf{Ind}$
    & $\mathbf{5.45 \pm 1.80}$
    & $95.7$
    & $5.36 \pm 1.80$
    & $91.35 \pm 23.90$ \\

    $\mathsf{PPO}$-$\mathsf{F}$-$\mathsf{Clu}$
    & $7.85 \pm 1.85$
    & $72.9$
    & $6.97 \pm 1.92$
    & $89.04 \pm 28.76$ \\

    $\mathsf{PPO}$-$\mathsf{A}$-$\mathsf{Clu}$
    & $7.68 \pm 1.80$
    & $62.9$
    & $7.13 \pm 1.77$
    & $\mathbf{91.46 \pm 21.70}$ \\

    $\mathsf{PPO}$-$\mathsf{F}$-$\mathsf{Pop}$
    & $8.33 \pm 2.20$
    & --
    & $8.05 \pm 2.34$
    & $89.62 \pm 28.46$ \\

    $\mathsf{PPO}$-$\mathsf{A}$-$\mathsf{Pop}$
    & $8.00 \pm 2.12$
    & --
    & $7.61 \pm 2.19$
    & $89.06 \pm 27.34$ \\

    \midrule

    $\mathsf{TRPO}$-$\mathsf{F}$-$\mathsf{Ind}$
    & $6.46 \pm 1.78$
    & $94.3$
    & $\mathbf{5.55 \pm 1.73}$
    & $\mathbf{93.19 \pm 20.03}$ \\

    $\mathsf{TRPO}$-$\mathsf{A}$-$\mathsf{Ind}$
    & $\mathbf{6.27 \pm 1.79}$
    & $88.6$
    & $5.82 \pm 1.80$
    & $90.49 \pm 23.49$ \\

    $\mathsf{TRPO}$-$\mathsf{F}$-$\mathsf{Clu}$
    & $7.73 \pm 1.68$
    & $67.1$
    & $6.68 \pm 1.81$
    & $92.23 \pm 23.03$ \\

    $\mathsf{TRPO}$-$\mathsf{A}$-$\mathsf{Clu}$
    & $7.62 \pm 1.71$
    & $60.0$
    & $6.77 \pm 1.80$
    & $91.27 \pm 20.82$ \\

    $\mathsf{TRPO}$-$\mathsf{F}$-$\mathsf{Pop}$
    & $8.48 \pm 1.85$
    & --
    & $7.48 \pm 2.05$
    & $91.17 \pm 24.31$ \\

    $\mathsf{TRPO}$-$\mathsf{A}$-$\mathsf{Pop}$
    & $7.96 \pm 1.73$
    & --
    & $7.39 \pm 1.84$
    & $93.09 \pm 19.72$ \\

    \midrule

    $\mathsf{PPOLag}$-$\mathsf{F}$-$\mathsf{Ind}$
    & $5.74 \pm 1.65$
    & $98.6$
    & $5.35 \pm 1.67$
    & $\mathbf{97.72 \pm 12.05}$ \\

    $\mathsf{PPOLag}$-$\mathsf{A}$-$\mathsf{Ind}$
    & $\mathbf{5.55 \pm 1.87}$
    & $94.3$
    & $\mathbf{5.34 \pm 1.83}$
    & $96.60 \pm 13.25$ \\

    $\mathsf{PPOLag}$-$\mathsf{F}$-$\mathsf{Clu}$
    & $7.43 \pm 1.96$
    & $68.6$
    & $6.84 \pm 2.04$
    & $89.39 \pm 27.51$ \\

    $\mathsf{PPOLag}$-$\mathsf{A}$-$\mathsf{Clu}$
    & $7.23 \pm 1.89$
    & $65.7$
    & $6.78 \pm 2.05$
    & $89.27 \pm 25.85$ \\

    $\mathsf{PPOLag}$-$\mathsf{F}$-$\mathsf{Pop}$
    & $7.96 \pm 2.10$
    & --
    & $7.65 \pm 2.28$
    & $90.76 \pm 27.12$ \\

    $\mathsf{PPOLag}$-$\mathsf{A}$-$\mathsf{Pop}$
    & $7.62 \pm 1.96$
    & --
    & $7.36 \pm 2.00$
    & $90.24 \pm 25.01$ \\

    \bottomrule
  \end{tabular}

  \caption{
    Additional online-policy results on 70 held-out test patients.
    Metrics, comparisons, abbreviations, and boldface conventions follow Table~\ref{tab:policy_evaluation_selected}; dashes indicate that the population policy is not compared with itself. The corresponding trajectory-level safety metrics are reported in Table~\ref{tab:online_safety_evaluation}.
  }
  \label{tab:policy_evaluation_all_appendix}
\end{table}

\subsection{Evaluation Under an Alternative Clustering Scheme}
\label{appendix:alternative_clustering_evaluation}

\begin{figure}[!tbp]
  \centering
  \captionsetup[subfigure]{skip=0pt}

  \begin{subfigure}[t]{0.24\linewidth}
    \centering
    \includegraphics[width=\linewidth]
    { 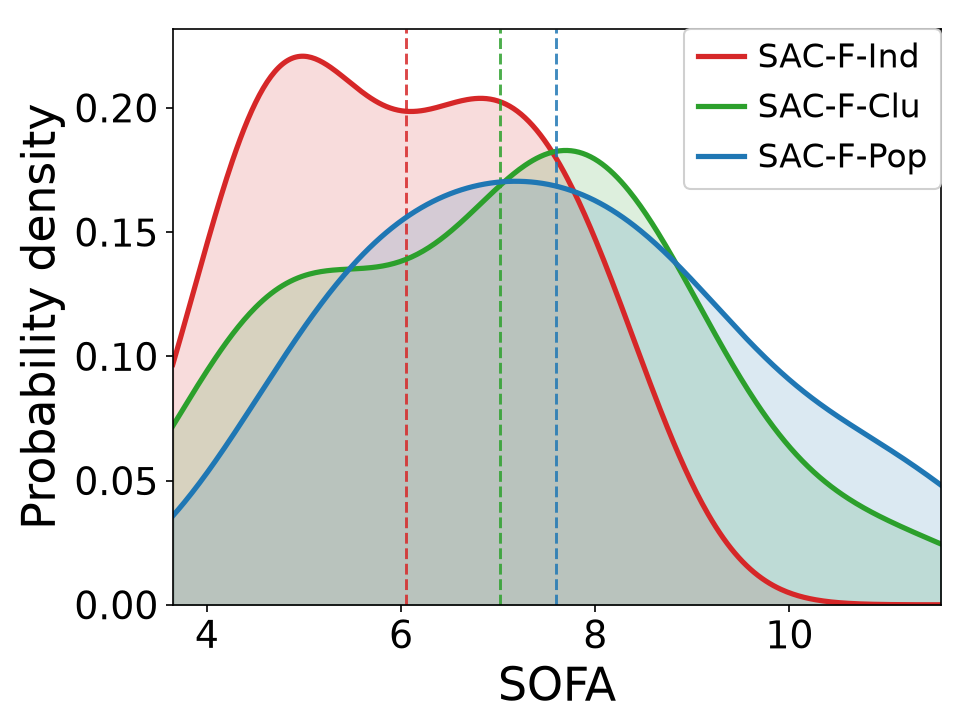}
    \caption{$\mathsf{SAC}$-F}
    \label{fig:alt_sac_fixed_dist}
  \end{subfigure}\hfill
  \begin{subfigure}[t]{0.24\linewidth}
    \centering
    \includegraphics[width=\linewidth]
    { 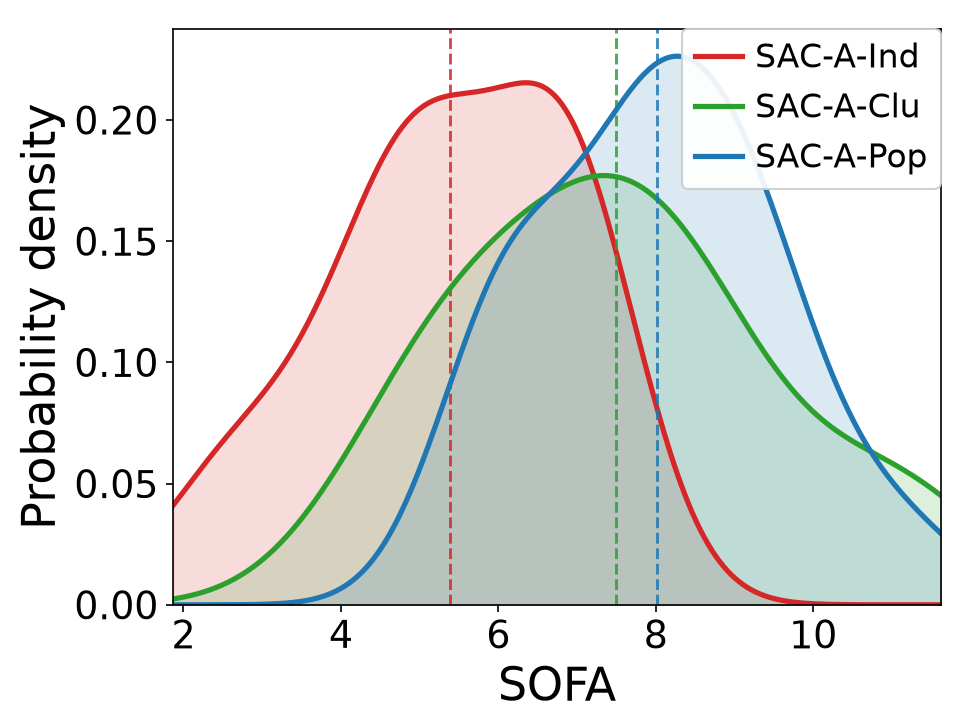}
    \caption{$\mathsf{SAC}$-A}
    \label{fig:alt_sac_adaptive_dist}
  \end{subfigure}\hfill
  \begin{subfigure}[t]{0.24\linewidth}
    \centering
    \includegraphics[width=\linewidth]
    { 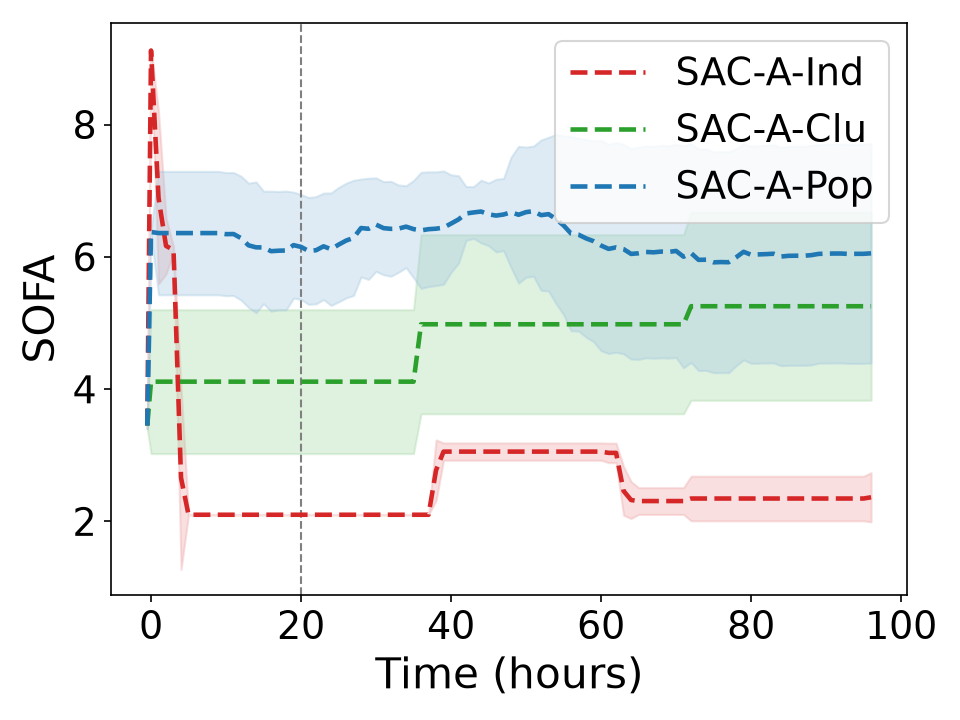}
    \caption{$\mathsf{SAC}$ SOFA}
    \label{fig:alt_sac_sofa}
  \end{subfigure}\hfill
  \begin{subfigure}[t]{0.24\linewidth}
    \centering
    \includegraphics[width=\linewidth]
    { 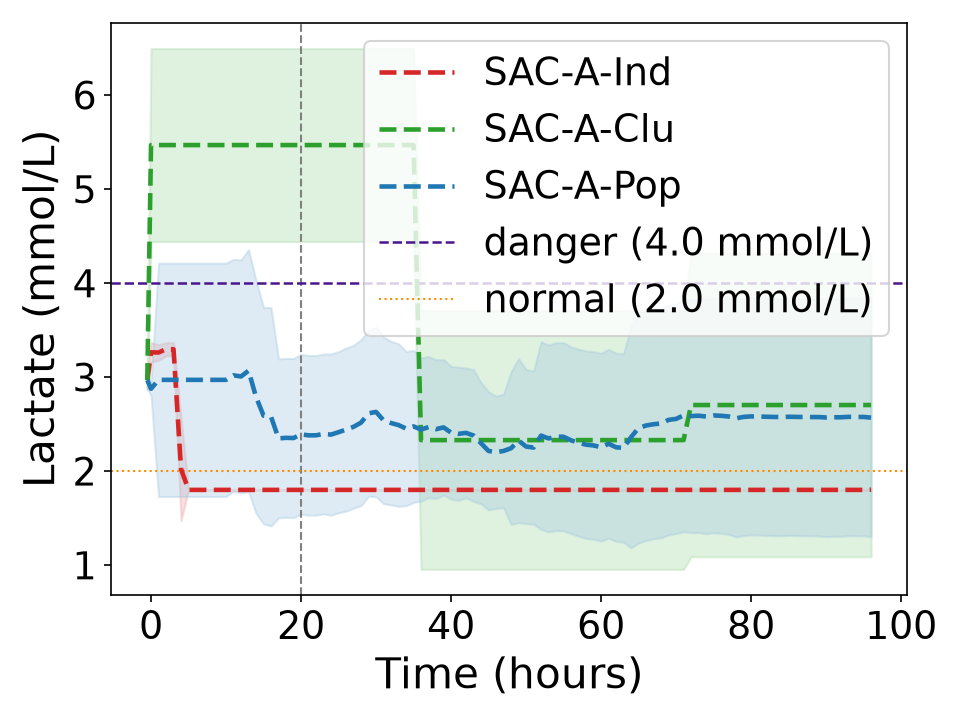}
    \caption{$\mathsf{SAC}$ Lactate}
    \label{fig:alt_sac_lactate}
  \end{subfigure}

  \par\vspace{-0.25em}

  \begin{subfigure}[t]{0.24\linewidth}
    \centering
    \includegraphics[width=\linewidth]
    { 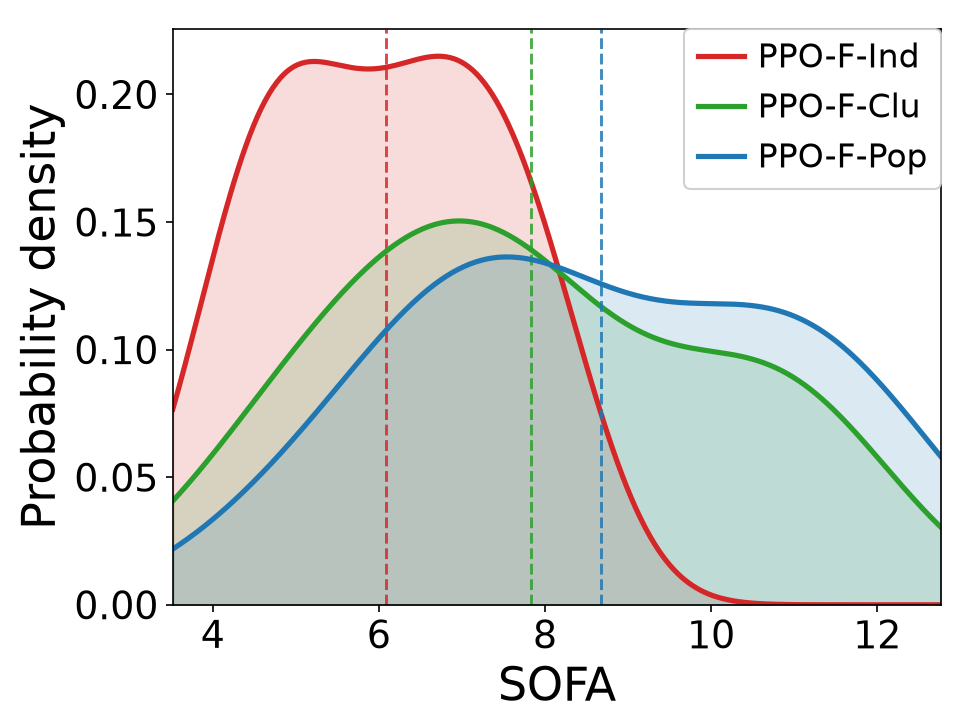}
    \caption{$\mathsf{PPO}$-F}
    \label{fig:alt_ppo_fixed_dist}
  \end{subfigure}\hfill
  \begin{subfigure}[t]{0.24\linewidth}
    \centering
    \includegraphics[width=\linewidth]
    { 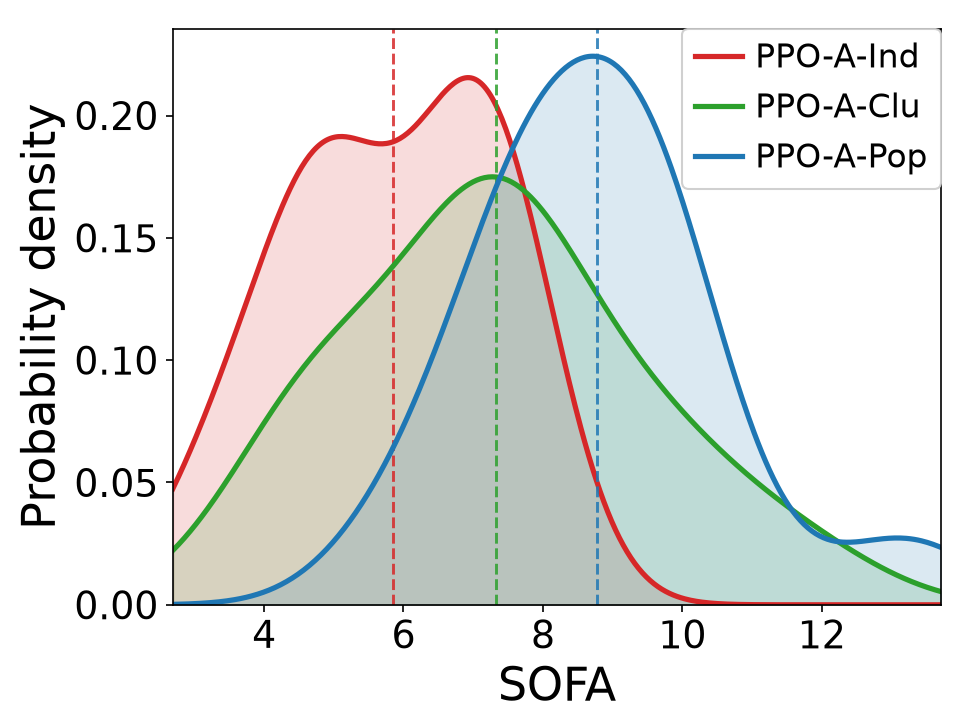}
    \caption{$\mathsf{PPO}$-A}
    \label{fig:alt_ppo_adaptive_dist}
  \end{subfigure}\hfill
  \begin{subfigure}[t]{0.24\linewidth}
    \centering
    \includegraphics[width=\linewidth]
    { 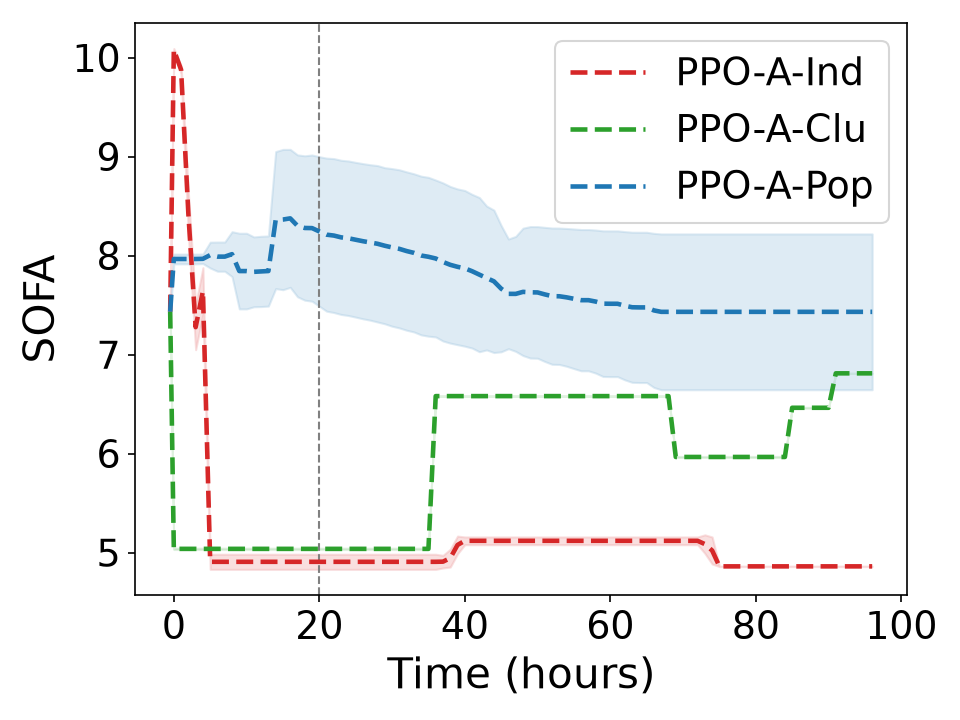}
    \caption{$\mathsf{PPO}$ SOFA}
    \label{fig:alt_ppo_sofa}
  \end{subfigure}\hfill
  \begin{subfigure}[t]{0.24\linewidth}
    \centering
    \includegraphics[width=\linewidth]
    { 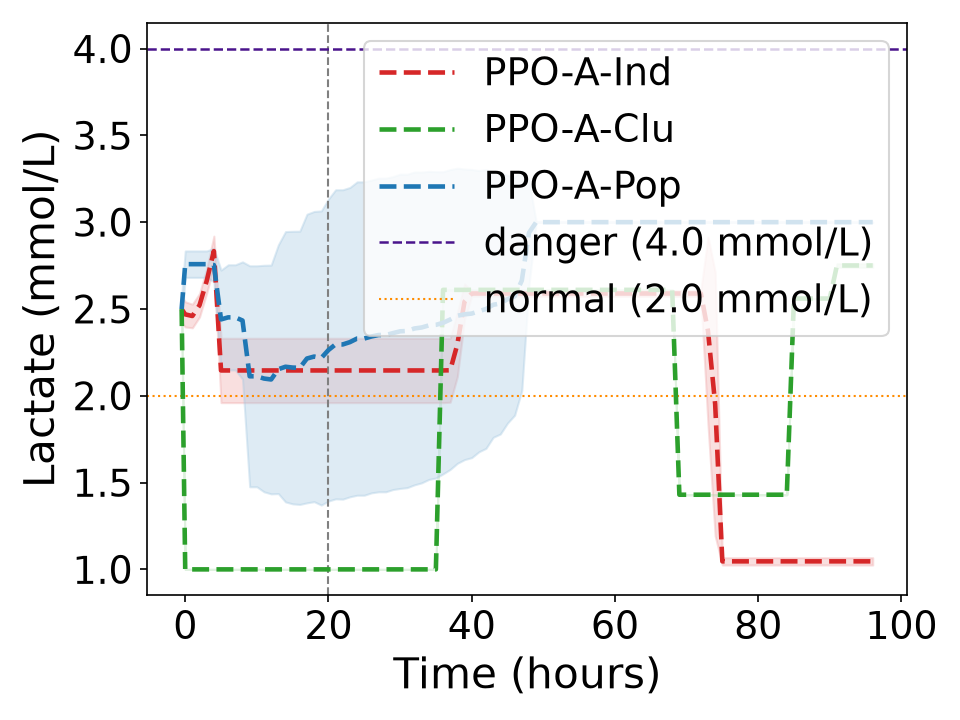}
    \caption{$\mathsf{PPO}$ Lactate}
    \label{fig:alt_ppo_lactate}
  \end{subfigure}

  \par\vspace{-0.25em}

  \begin{subfigure}[t]{0.24\linewidth}
    \centering
    \includegraphics[width=\linewidth]
    { 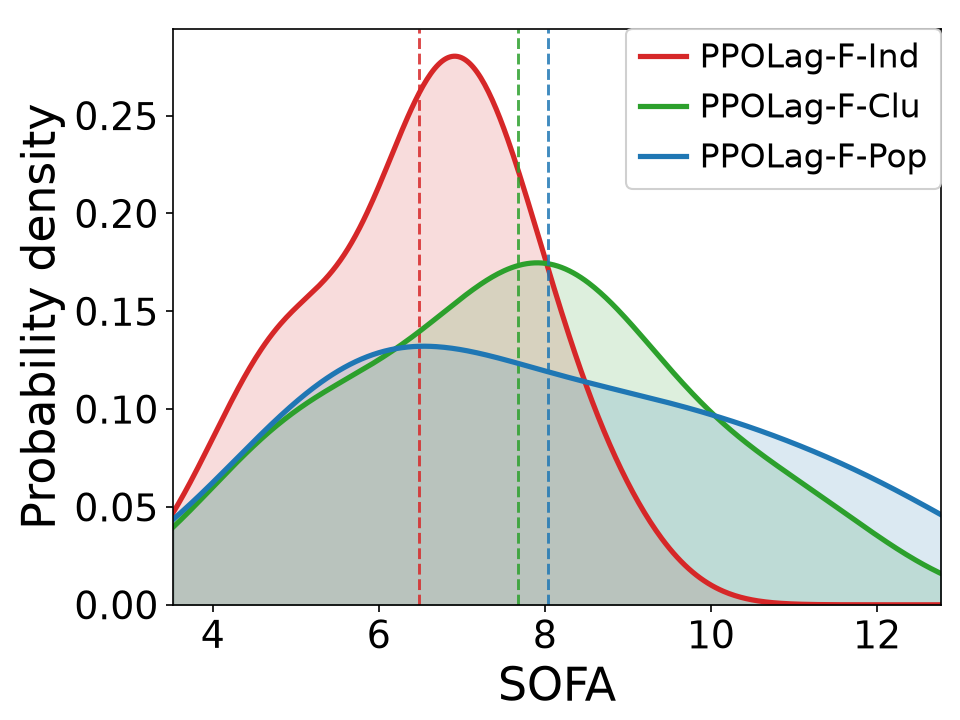}
    \caption{$\mathsf{PPOLag}$-F}
    \label{fig:alt_ppolag_fixed_dist}
  \end{subfigure}\hfill
  \begin{subfigure}[t]{0.24\linewidth}
    \centering
    \includegraphics[width=\linewidth]
    { 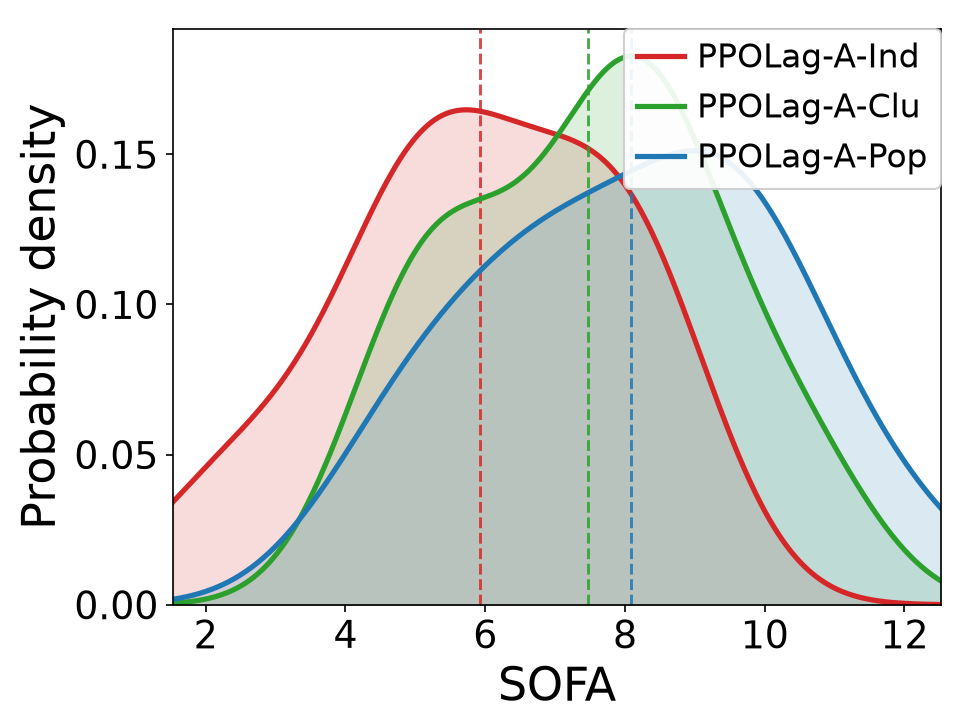}
    \caption{$\mathsf{PPOLag}$-A}
    \label{fig:alt_ppolag_adaptive_dist}
  \end{subfigure}\hfill
  \begin{subfigure}[t]{0.24\linewidth}
    \centering
    \includegraphics[width=\linewidth]
    { 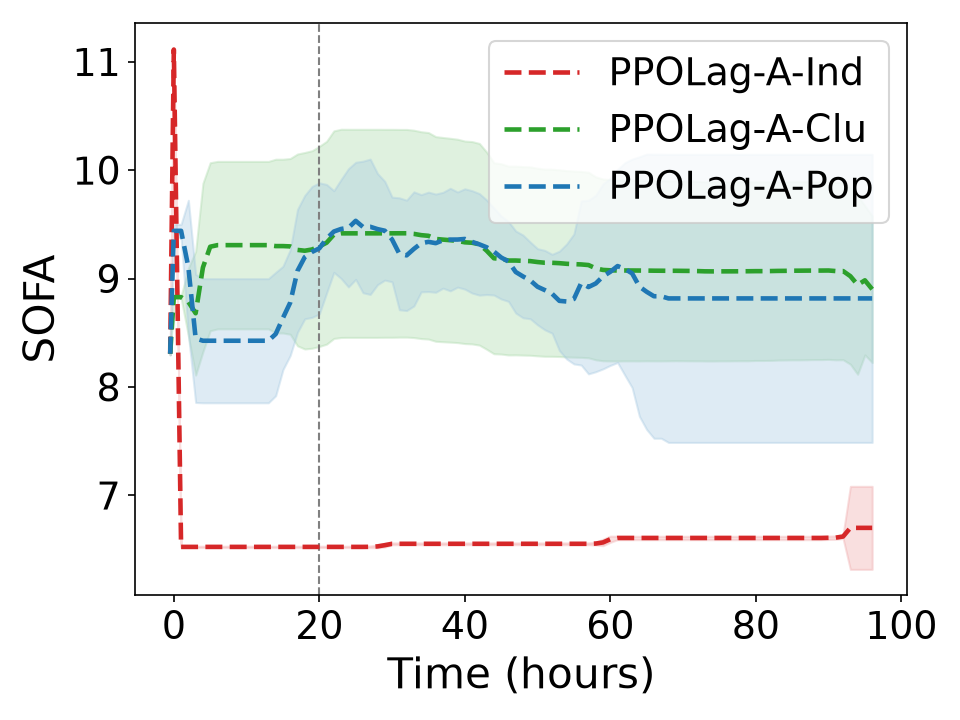}
    \caption{$\mathsf{PPOLag}$ SOFA}
    \label{fig:alt_ppolag_sofa}
  \end{subfigure}\hfill
  \begin{subfigure}[t]{0.24\linewidth}
    \centering
    \includegraphics[width=\linewidth]
    { 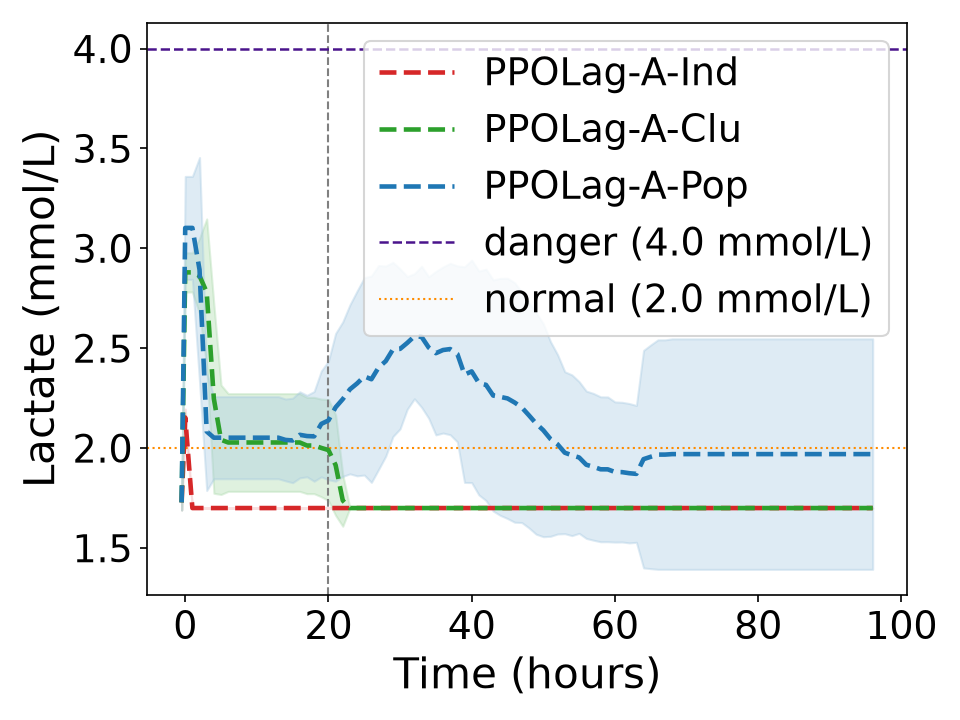}
    \caption{$\mathsf{PPOLag}$ Lactate}
    \label{fig:alt_ppolag_lactate}
  \end{subfigure}

  \par\vspace{-0.25em}

  \begin{subfigure}[t]{0.24\linewidth}
    \centering
    \includegraphics[width=\linewidth]
    { 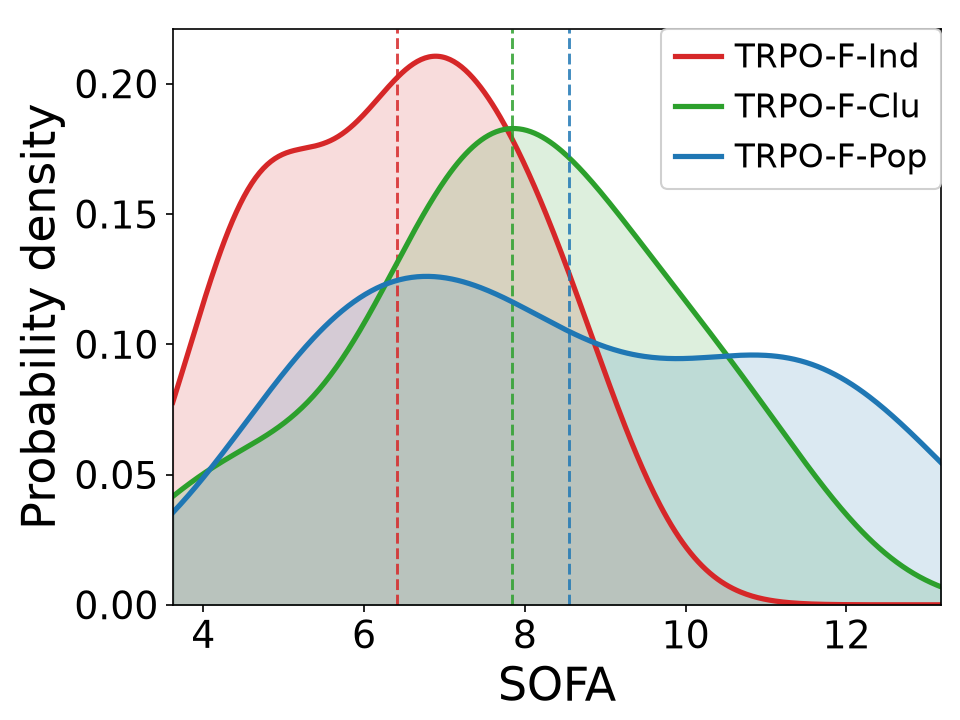}
    \caption{$\mathsf{TRPO}$-F}
    \label{fig:alt_trpo_fixed_dist}
  \end{subfigure}\hfill
  \begin{subfigure}[t]{0.24\linewidth}
    \centering
    \includegraphics[width=\linewidth]
    { 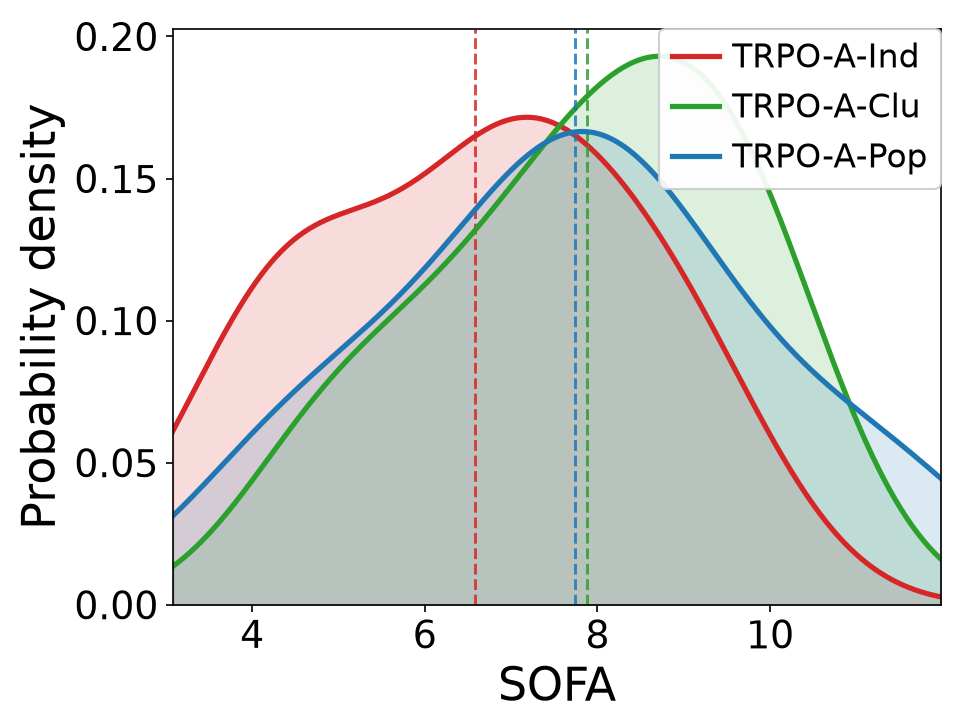}
    \caption{$\mathsf{TRPO}$-A}
    \label{fig:alt_trpo_adaptive_dist}
  \end{subfigure}\hfill
  \begin{subfigure}[t]{0.24\linewidth}
    \centering
    \includegraphics[width=\linewidth]
    { 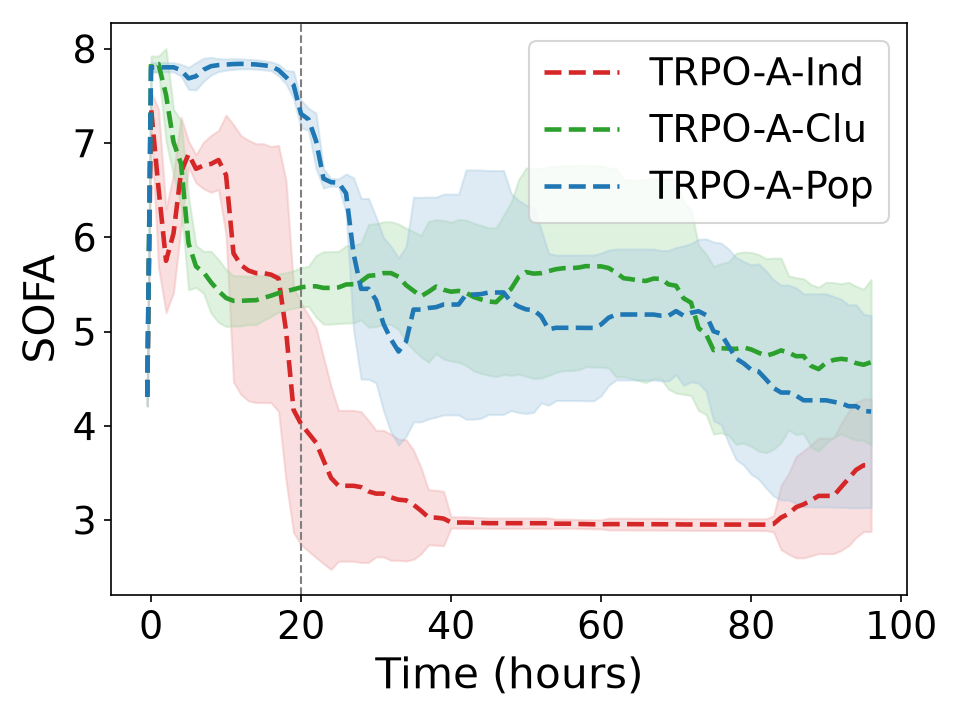}
    \caption{$\mathsf{TRPO}$ SOFA}
    \label{fig:alt_trpo_sofa}
  \end{subfigure}\hfill
  \begin{subfigure}[t]{0.24\linewidth}
    \centering
    \includegraphics[width=\linewidth]
    { 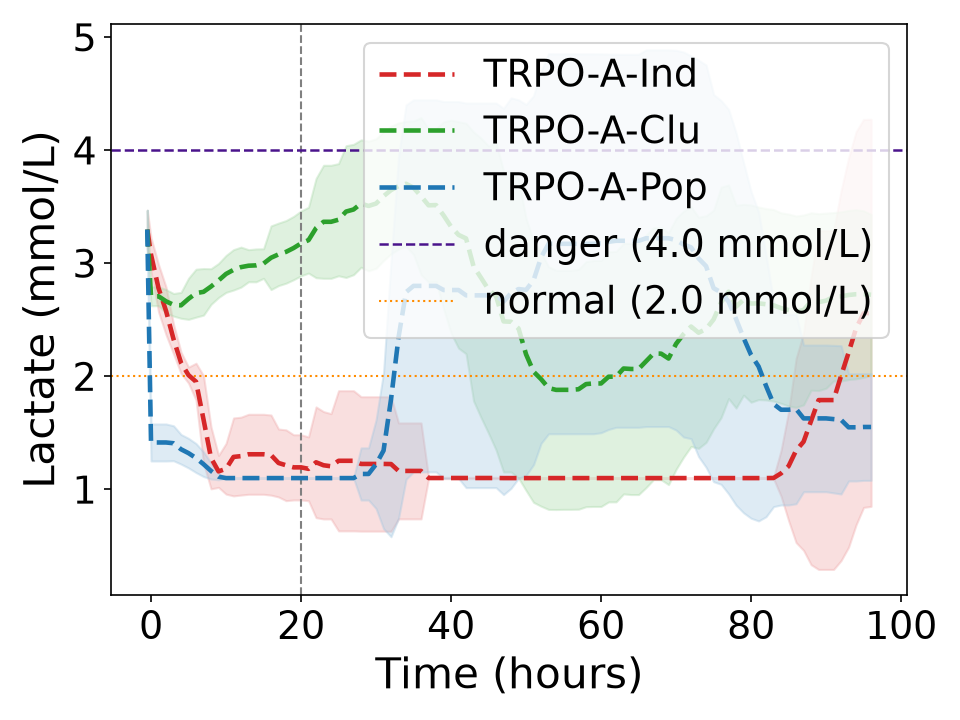}
    \caption{$\mathsf{TRPO}$ Lactate}
    \label{fig:alt_trpo_lactate}
  \end{subfigure}

  \par\vspace{-0.25em}

  \begin{subfigure}[t]{0.24\linewidth}
    \centering
    \includegraphics[width=\linewidth]
    { 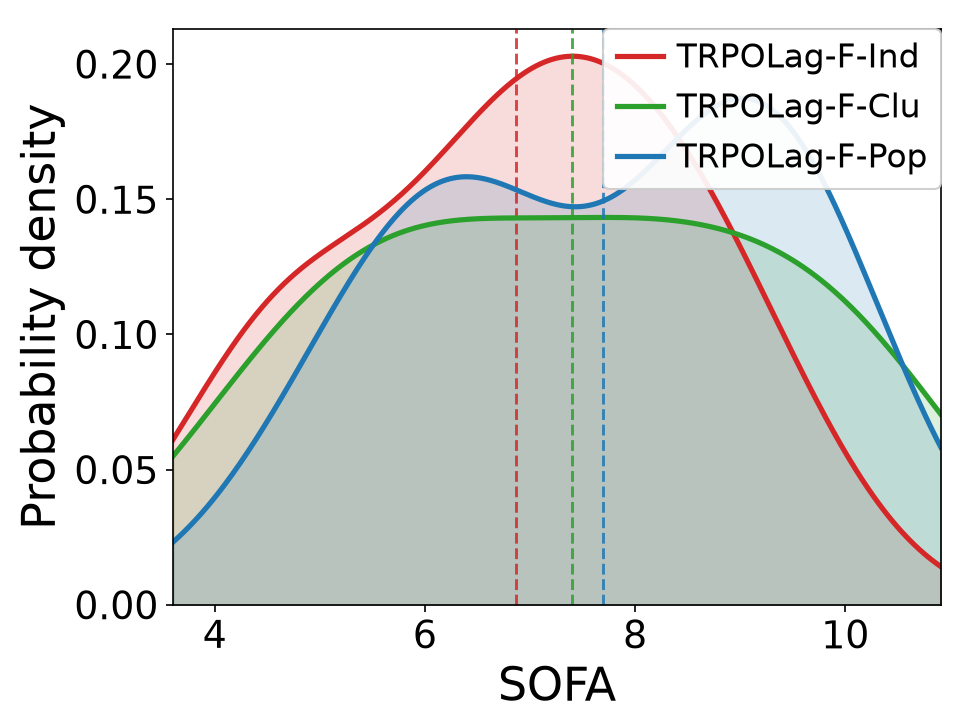}
    \caption{$\mathsf{TRPOLag}$-F}
    \label{fig:alt_trpolag_fixed_dist}
  \end{subfigure}\hfill
  \begin{subfigure}[t]{0.24\linewidth}
    \centering
    \includegraphics[width=\linewidth]
    { 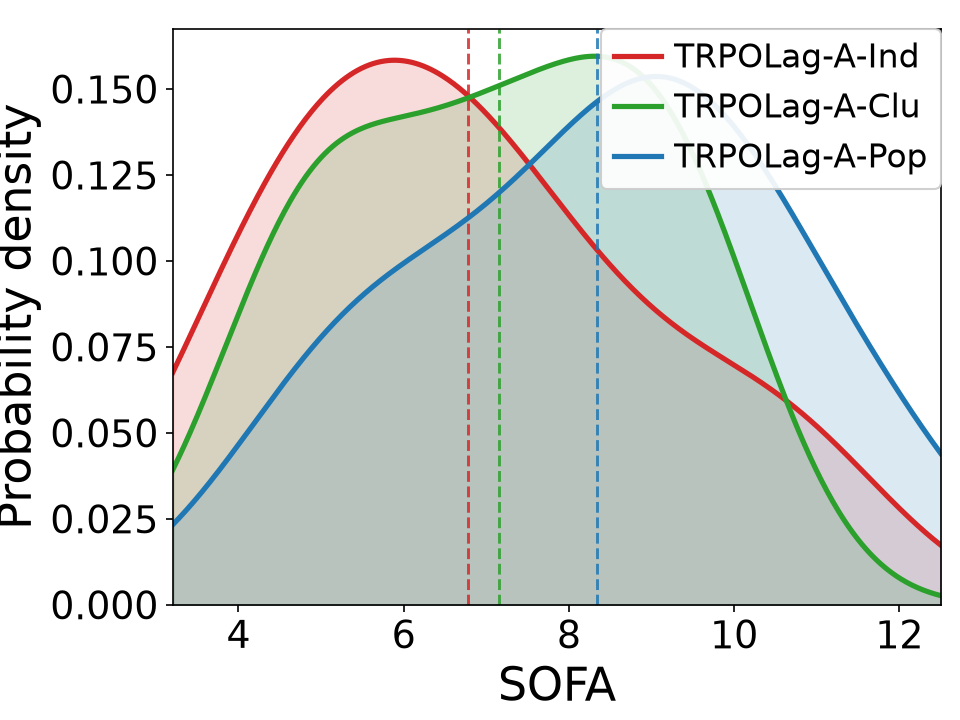}
    \caption{$\mathsf{TRPOLag}$-A}
    \label{fig:alt_trpolag_adaptive_dist}
  \end{subfigure}\hfill
  \begin{subfigure}[t]{0.24\linewidth}
    \centering
    \includegraphics[width=\linewidth]
    { 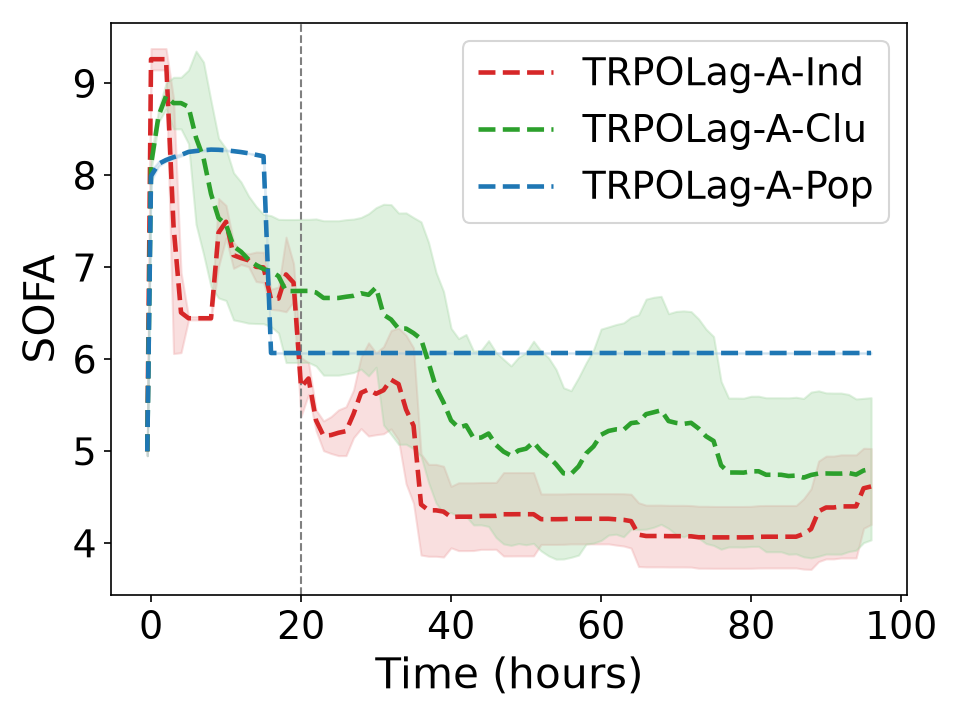}
    \caption{$\mathsf{TRPOLag}$ SOFA}
    \label{fig:alt_trpolag_sofa}
  \end{subfigure}\hfill
  \begin{subfigure}[t]{0.24\linewidth}
    \centering
    \includegraphics[width=\linewidth]
    { 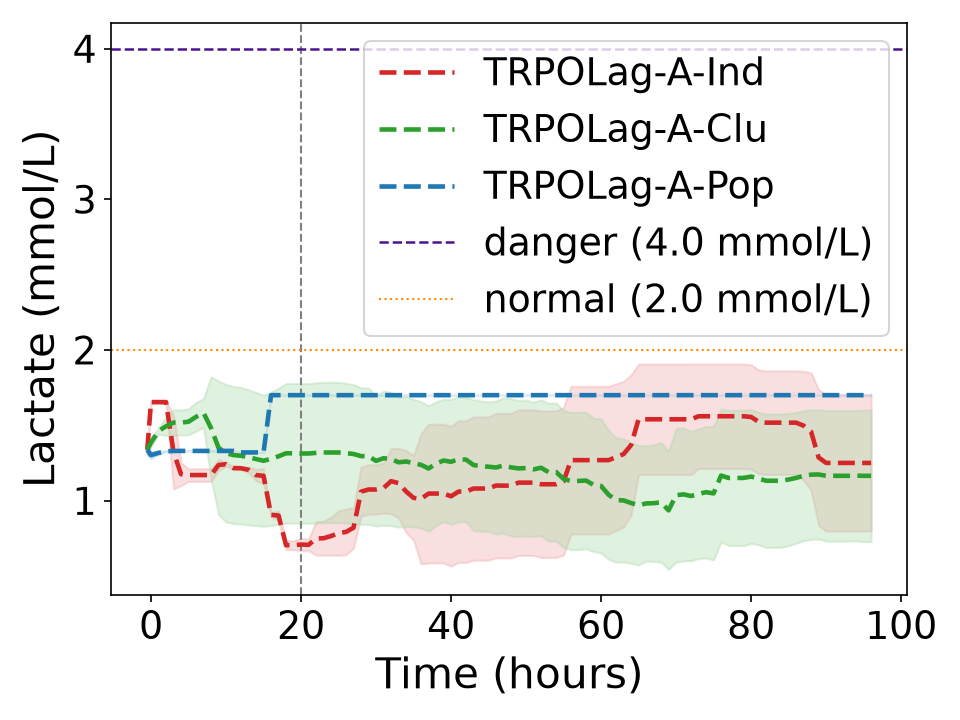}
    \caption{$\mathsf{TRPOLag}$ Lactate}
    \label{fig:alt_trpolag_lactate}
  \end{subfigure}

  \caption{
    Online evaluation under the alternative 15-cluster partition.
    Each row corresponds to one RL algorithm. From left to right, the
    columns show the distributions of final $\mathsf{SOFA}$ scores under
    fixed and adaptive timing, followed by adaptive-timing
    $\mathsf{SOFA}$ and Lactate trajectories on a representative test
    patient. The distributional results are computed over 15 held-out
    patients, with each patient-level score averaged over ten noise
    realizations.
  }
  \label{fig:alternative_clustering_evaluation}
\end{figure}

\textbf{Model and Dataset Construction.}
\noindent
To evaluate the sensitivity of the results to the patient-grouping
strategy, we additionally consider an alternative setting with
$N_{\mathsf{clu}}=15$ patient groups. Each group contains one
representative training patient and one matched held-out test patient,
such that $|\mathcal{C}_k^{\mathsf{train}}|=1$ and
$|\mathcal{C}_k^{\mathsf{test}}|=1$. The cluster-level model and policy
for each group are constructed using its representative training
patient. The population-level model and policy aggregate all training
patients, giving
$\sum_{k=1}^{N_{\mathsf{clu}}}
|\mathcal{C}_k^{\mathsf{train}}|=15$ training patients in total.
Evaluation is conducted on the corresponding
$\sum_{k=1}^{N_{\mathsf{clu}}}
|\mathcal{C}_k^{\mathsf{test}}|=15$ held-out test patients.
These test patients are excluded from the construction and training of
the cluster- and population-level models and policies. The evaluated RL
algorithms, interaction-timing settings, and evaluation metrics remain
the same as those used in the primary evaluation.
\noindent
Based on this alternative patient-grouping setting,
Figure~\ref{fig:alternative_clustering_evaluation} and
Table~\ref{tab:alternative_clustering_evaluation} report the results
for $\mathsf{SAC}$, $\mathsf{PPO}$, $\mathsf{TRPO}$,
$\mathsf{PPOLag}$, and $\mathsf{TRPOLag}$ under both fixed and adaptive
timing. The figure presents the final $\mathsf{SOFA}$ distributions and
representative physiological trajectories, while the table summarizes
the corresponding treatment-effectiveness and safety metrics.

The results are consistent with those obtained under the primary
70-patient partition. The individual oracle generally achieves the
lowest $\mathsf{SOFA}$ scores, while the cluster policies recover part
of this personalization benefit relative to the population policies.
Under the alternative partition, the cluster policy achieves a lower
mean final $\mathsf{SOFA}$ score than the population policy in nine of
the ten algorithm--timing combinations, with
$\mathsf{TRPO}$-$\mathsf{A}$ being the only exception. This result
suggests that the benefit of cluster-based transfer persists across
different patient-grouping strategies, although its magnitude depends
on the selected partition.

\begin{table}[t!]
  \centering

  \fontsize{8pt}{8pt}\selectfont
  \setlength{\tabcolsep}{4pt}
  \renewcommand{\arraystretch}{0.82}
  \setlength{\aboverulesep}{0.3ex}
  \setlength{\belowrulesep}{0.3ex}
  \captionsetup{skip=2pt}

  \begin{tabular}{lcccc}
    \toprule
    Policy
    & SOFA(all) ($\downarrow$)
    & Lower SOFA vs. Pop (\%)
    & SOFA(96h) ($\downarrow$)
    & SafeTime\% ($\uparrow$) \\
    \midrule

    $\mathsf{SAC}$-$\mathsf{F}$-$\mathsf{Ind}$
    & $6.73 \pm 1.35$
    & $100.0$
    & $6.06 \pm 1.33$
    & $\mathbf{76.03 \pm 38.46}$ \\

    $\mathsf{SAC}$-$\mathsf{A}$-$\mathsf{Ind}$
    & $\mathbf{5.79 \pm 1.46}$
    & $100.0$
    & $\mathbf{5.39 \pm 1.46}$
    & $70.52 \pm 41.71$ \\

    $\mathsf{SAC}$-$\mathsf{F}$-$\mathsf{Clu}$
    & $7.28 \pm 1.70$
    & $73.3$
    & $7.02 \pm 1.86$
    & $74.02 \pm 37.24$ \\

    $\mathsf{SAC}$-$\mathsf{A}$-$\mathsf{Clu}$
    & $7.51 \pm 1.84$
    & $80.0$
    & $7.50 \pm 1.91$
    & $70.10 \pm 34.31$ \\

    $\mathsf{SAC}$-$\mathsf{F}$-$\mathsf{Pop}$
    & $7.80 \pm 1.81$
    & --
    & $7.60 \pm 1.89$
    & $70.25 \pm 41.38$ \\

    $\mathsf{SAC}$-$\mathsf{A}$-$\mathsf{Pop}$
    & $7.89 \pm 1.49$
    & --
    & $8.02 \pm 1.45$
    & $70.94 \pm 34.44$ \\

    \midrule

    $\mathsf{PPO}$-$\mathsf{F}$-$\mathsf{Ind}$
    & $6.56 \pm 1.39$
    & $100.0$
    & $6.08 \pm 1.31$
    & $78.64 \pm 38.06$ \\

    $\mathsf{PPO}$-$\mathsf{A}$-$\mathsf{Ind}$
    & $\mathbf{5.92 \pm 1.42}$
    & $100.0$
    & $\mathbf{5.85 \pm 1.42}$
    & $65.15 \pm 41.45$ \\

    $\mathsf{PPO}$-$\mathsf{F}$-$\mathsf{Clu}$
    & $8.11 \pm 1.90$
    & $60.0$
    & $7.83 \pm 2.18$
    & $78.47 \pm 34.36$ \\

    $\mathsf{PPO}$-$\mathsf{A}$-$\mathsf{Clu}$
    & $7.48 \pm 1.88$
    & $86.7$
    & $7.33 \pm 2.01$
    & $69.87 \pm 37.24$ \\

    $\mathsf{PPO}$-$\mathsf{F}$-$\mathsf{Pop}$
    & $8.88 \pm 2.03$
    & --
    & $8.68 \pm 2.24$
    & $73.44 \pm 37.28$ \\

    $\mathsf{PPO}$-$\mathsf{A}$-$\mathsf{Pop}$
    & $9.05 \pm 1.72$
    & --
    & $8.78 \pm 1.67$
    & $\mathbf{85.89 \pm 16.47}$ \\

    \midrule

    $\mathsf{PPOLag}$-$\mathsf{F}$-$\mathsf{Ind}$
    & $6.82 \pm 1.32$
    & $86.7$
    & $6.48 \pm 1.26$
    & $\mathbf{92.64 \pm 20.40}$ \\

    $\mathsf{PPOLag}$-$\mathsf{A}$-$\mathsf{Ind}$
    & $\mathbf{6.23 \pm 1.98}$
    & $100.0$
    & $\mathbf{5.93 \pm 1.90}$
    & $78.41 \pm 29.20$ \\

    $\mathsf{PPOLag}$-$\mathsf{F}$-$\mathsf{Clu}$
    & $7.80 \pm 1.70$
    & $46.7$
    & $7.67 \pm 1.95$
    & $78.93 \pm 35.80$ \\

    $\mathsf{PPOLag}$-$\mathsf{A}$-$\mathsf{Clu}$
    & $7.60 \pm 1.79$
    & $73.3$
    & $7.47 \pm 1.79$
    & $63.76 \pm 36.93$ \\

    $\mathsf{PPOLag}$-$\mathsf{F}$-$\mathsf{Pop}$
    & $8.14 \pm 2.31$
    & --
    & $8.04 \pm 2.42$
    & $65.51 \pm 42.73$ \\

    $\mathsf{PPOLag}$-$\mathsf{A}$-$\mathsf{Pop}$
    & $8.63 \pm 1.60$
    & --
    & $8.09 \pm 2.09$
    & $76.86 \pm 30.76$ \\

    \midrule

    $\mathsf{TRPO}$-$\mathsf{F}$-$\mathsf{Ind}$
    & $7.36 \pm 1.51$
    & $100.0$
    & $\mathbf{6.42 \pm 1.45}$
    & $74.34 \pm 38.30$ \\

    $\mathsf{TRPO}$-$\mathsf{A}$-$\mathsf{Ind}$
    & $\mathbf{7.00 \pm 1.87}$
    & $66.7$
    & $6.58 \pm 1.80$
    & $74.05 \pm 37.81$ \\

    $\mathsf{TRPO}$-$\mathsf{F}$-$\mathsf{Clu}$
    & $8.28 \pm 1.54$
    & $80.0$
    & $7.84 \pm 1.93$
    & $\mathbf{79.61 \pm 39.03}$ \\

    $\mathsf{TRPO}$-$\mathsf{A}$-$\mathsf{Clu}$
    & $8.35 \pm 1.61$
    & $46.7$
    & $7.88 \pm 1.72$
    & $70.57 \pm 31.74$ \\

    $\mathsf{TRPO}$-$\mathsf{F}$-$\mathsf{Pop}$
    & $8.82 \pm 2.29$
    & --
    & $8.55 \pm 2.52$
    & $75.65 \pm 34.35$ \\

    $\mathsf{TRPO}$-$\mathsf{A}$-$\mathsf{Pop}$
    & $8.23 \pm 1.62$
    & --
    & $7.74 \pm 2.09$
    & $76.88 \pm 33.03$ \\

    \midrule

    $\mathsf{TRPOLag}$-$\mathsf{F}$-$\mathsf{Ind}$
    & $7.59 \pm 1.41$
    & $93.3$
    & $6.87 \pm 1.59$
    & $\mathbf{91.82 \pm 21.52}$ \\

    $\mathsf{TRPOLag}$-$\mathsf{A}$-$\mathsf{Ind}$
    & $\mathbf{7.09 \pm 2.05}$
    & $73.3$
    & $\mathbf{6.78 \pm 2.16}$
    & $82.79 \pm 25.84$ \\

    $\mathsf{TRPOLag}$-$\mathsf{F}$-$\mathsf{Clu}$
    & $7.70 \pm 1.79$
    & $73.3$
    & $7.40 \pm 1.97$
    & $77.96 \pm 34.42$ \\

    $\mathsf{TRPOLag}$-$\mathsf{A}$-$\mathsf{Clu}$
    & $7.53 \pm 1.74$
    & $73.3$
    & $7.16 \pm 1.83$
    & $71.67 \pm 36.45$ \\

    $\mathsf{TRPOLag}$-$\mathsf{F}$-$\mathsf{Pop}$
    & $8.46 \pm 1.35$
    & --
    & $7.70 \pm 1.68$
    & $78.37 \pm 32.61$ \\

    $\mathsf{TRPOLag}$-$\mathsf{A}$-$\mathsf{Pop}$
    & $8.57 \pm 1.75$
    & --
    & $8.34 \pm 2.15$
    & $76.50 \pm 34.60$ \\

    \bottomrule
  \end{tabular}

  \caption{
  Online-policy results under the alternative clustering scheme on
  15 held-out test patients. Metrics, comparisons, abbreviations, and
  boldface conventions follow
  Table~\ref{tab:policy_evaluation_selected}; dashes indicate that the
  population policy is not compared with itself.
  }
  \label{tab:alternative_clustering_evaluation}
\end{table}

\subsection{Trajectory-Level Safety Evaluation for Online RL}
\label{appendix:online_safety_evaluation}

To evaluate safety independently of the number and spacing of intervention
times, we assess all policies on a common physical-time grid over the
complete treatment horizon $[0,T]$. Let $\mathrm{Lac}(t)$ denote the
simulated lactate trajectory and let $\ell_{\mathrm{safe}}=4.0$ mmol/L
denote the safety threshold. We report four complementary metrics:
\emph{SafeTime\%}, the fraction of physical time for which
$\mathrm{Lac}(t)<\ell_{\mathrm{safe}}$; \emph{ViolDur}, the total duration
for which $\mathrm{Lac}(t)\geq\ell_{\mathrm{safe}}$; \emph{PeakLac}, the
maximum lactate value over the trajectory; and \emph{ExcessAUC}, the time
integral of the threshold exceedance
$[\mathrm{Lac}(t)-\ell_{\mathrm{safe}}]_+$. These metrics respectively
capture the amount of safe physical time, duration of unsafe exposure,
worst-case excursion, and the combined duration and severity of threshold
violation.
\begin{align}
\mathrm{SafeTime}\%
&=
\frac{100}{T}
\int_0^T
\mathbf{1}\!\left\{
\mathrm{Lac}(t)<\ell_{\mathrm{safe}}
\right\}
\,\mathsf{d}t,
\\
\mathrm{ViolDur}
&=
\int_0^T
\mathbf{1}\!\left\{
\mathrm{Lac}(t)\geq\ell_{\mathrm{safe}}
\right\}
\,\mathsf{d}t,
\\
\mathrm{PeakLac}
&=
\max_{t\in[0,T]}
\mathrm{Lac}(t),
\\
\mathrm{ExcessAUC}
&=
\int_0^T
\left[
\mathrm{Lac}(t)-\ell_{\mathrm{safe}}
\right]_+
\,\mathsf{d}t.
\end{align}

\begin{table}[t]
\centering
\fontsize{7.5pt}{8pt}\selectfont
\setlength{\tabcolsep}{3pt}
\renewcommand{\arraystretch}{0.82}
\setlength{\aboverulesep}{0.3ex}
\setlength{\belowrulesep}{0.3ex}
\captionsetup{skip=2pt}

\begin{tabular}{lcccc}
\toprule
Policy
& SafeTime\% ($\uparrow$)
& ViolDur (h) ($\downarrow$)
& PeakLac ($\downarrow$)
& ExcessAUC ($\downarrow$)
\\
\midrule

$\mathsf{SAC}$-$\mathsf{F}$-$\mathsf{Ind}$
    & $91.95 \pm 22.83$ & $7.71 \pm 21.92$ & $2.83 \pm 2.57$ & $12.88 \pm 48.72$ \\

$\mathsf{SAC}$-$\mathsf{A}$-$\mathsf{Ind}$
    & $90.34 \pm 24.14$ & $9.26 \pm 23.20$ & $2.89 \pm 2.08$ & $17.26 \pm 72.36$ \\

$\mathsf{SAC}$-$\mathsf{F}$-$\mathsf{Clu}$
    & $90.87 \pm 24.90$ & $8.77 \pm 23.92$ & $2.80 \pm 2.36$ & $25.84 \pm 143.59$ \\

$\mathsf{SAC}$-$\mathsf{A}$-$\mathsf{Clu}$
    & $89.98 \pm 23.86$ & $9.61 \pm 22.90$ & $2.95 \pm 2.34$ & $24.35 \pm 101.26$ \\

$\mathsf{SAC}$-$\mathsf{F}$-$\mathsf{Pop}$
    & $90.48 \pm 26.80$ & $9.14 \pm 25.73$ & $2.87 \pm 2.62$ & $36.29 \pm 189.17$ \\

$\mathsf{SAC}$-$\mathsf{A}$-$\mathsf{Pop}$
    & $90.71 \pm 22.54$ & $8.90 \pm 21.61$ & $2.99 \pm 2.53$ & $26.96 \pm 151.06$ \\

\midrule

$\mathsf{PPO}$-$\mathsf{F}$-$\mathsf{Ind}$
    & $91.08 \pm 23.76$ & $8.55 \pm 22.80$ & $2.82 \pm 2.46$ & $12.73 \pm 40.91$ \\

$\mathsf{PPO}$-$\mathsf{A}$-$\mathsf{Ind}$
    & $91.35 \pm 23.90$ & $8.27 \pm 22.95$ & $2.97 \pm 2.01$ & $13.48 \pm 50.35$ \\

$\mathsf{PPO}$-$\mathsf{F}$-$\mathsf{Clu}$
    & $89.04 \pm 28.76$ & $10.53 \pm 27.63$ & $2.80 \pm 2.63$ & $31.65 \pm 176.28$ \\

$\mathsf{PPO}$-$\mathsf{A}$-$\mathsf{Clu}$
    & $91.46 \pm 21.70$ & $8.19 \pm 20.83$ & $2.99 \pm 2.59$ & $27.23 \pm 145.41$ \\

$\mathsf{PPO}$-$\mathsf{F}$-$\mathsf{Pop}$
    & $89.62 \pm 28.46$ & $9.96 \pm 27.33$ & $2.84 \pm 2.54$ & $16.24 \pm 69.55$ \\

$\mathsf{PPO}$-$\mathsf{A}$-$\mathsf{Pop}$
    & $89.06 \pm 27.34$ & $10.48 \pm 26.26$ & $2.99 \pm 2.28$ & $21.35 \pm 83.77$ \\

\midrule

$\mathsf{TRPO}$-$\mathsf{F}$-$\mathsf{Ind}$
    & $93.19 \pm 20.03$ & $6.51 \pm 19.19$ & $2.84 \pm 2.41$ & $9.02 \pm 35.38$ \\

$\mathsf{TRPO}$-$\mathsf{A}$-$\mathsf{Ind}$
    & $90.49 \pm 23.49$ & $9.11 \pm 22.55$ & $3.03 \pm 2.36$ & $13.19 \pm 37.85$ \\

$\mathsf{TRPO}$-$\mathsf{F}$-$\mathsf{Clu}$
    & $92.23 \pm 23.03$ & $7.45 \pm 22.10$ & $2.82 \pm 2.54$ & $15.54 \pm 58.15$ \\

$\mathsf{TRPO}$-$\mathsf{A}$-$\mathsf{Clu}$
    & $91.27 \pm 20.82$ & $8.37 \pm 19.99$ & $3.04 \pm 2.63$ & $25.96 \pm 149.66$ \\

$\mathsf{TRPO}$-$\mathsf{F}$-$\mathsf{Pop}$
    & $91.17 \pm 24.31$ & $8.47 \pm 23.38$ & $2.83 \pm 2.45$ & $16.32 \pm 71.68$ \\

$\mathsf{TRPO}$-$\mathsf{A}$-$\mathsf{Pop}$
    & $93.09 \pm 19.72$ & $6.60 \pm 18.91$ & $3.04 \pm 2.60$ & $12.20 \pm 39.79$ \\

\midrule

$\mathsf{PPOLag}$-$\mathsf{F}$-$\mathsf{Ind}$
    & $97.72 \pm 12.05$ & $2.15 \pm 11.55$ & $2.72 \pm 2.14$ & $3.50 \pm 14.92$ \\

$\mathsf{PPOLag}$-$\mathsf{A}$-$\mathsf{Ind}$
    & $96.60 \pm 13.25$ & $3.24 \pm 12.70$ & $2.72 \pm 1.37$ & $3.61 \pm 11.81$ \\

$\mathsf{PPOLag}$-$\mathsf{F}$-$\mathsf{Clu}$
    & $89.39 \pm 27.51$ & $10.20 \pm 26.43$ & $2.85 \pm 2.64$ & $35.60 \pm 186.08$ \\

$\mathsf{PPOLag}$-$\mathsf{A}$-$\mathsf{Clu}$
    & $89.27 \pm 25.85$ & $10.29 \pm 24.81$ & $2.89 \pm 2.52$ & $34.23 \pm 175.37$ \\

$\mathsf{PPOLag}$-$\mathsf{F}$-$\mathsf{Pop}$
    & $90.76 \pm 27.12$ & $8.86 \pm 26.05$ & $2.84 \pm 2.53$ & $14.47 \pm 55.30$ \\

$\mathsf{PPOLag}$-$\mathsf{A}$-$\mathsf{Pop}$
    & $90.24 \pm 25.01$ & $9.36 \pm 24.01$ & $2.76 \pm 1.71$ & $14.50 \pm 51.28$ \\

\midrule

$\mathsf{TRPOLag}$-$\mathsf{F}$-$\mathsf{Ind}$
    & $97.46 \pm 12.65$ & $2.41 \pm 12.12$ & $2.66 \pm 2.40$ & $4.42 \pm 23.64$ \\

$\mathsf{TRPOLag}$-$\mathsf{A}$-$\mathsf{Ind}$
    & $95.99 \pm 13.76$ & $3.84 \pm 13.21$ & $2.54 \pm 1.10$ & $4.02 \pm 12.12$ \\

$\mathsf{TRPOLag}$-$\mathsf{F}$-$\mathsf{Clu}$
    & $91.45 \pm 23.62$ & $8.22 \pm 22.71$ & $2.79 \pm 2.21$ & $25.86 \pm 127.96$ \\

$\mathsf{TRPOLag}$-$\mathsf{A}$-$\mathsf{Clu}$
    & $88.37 \pm 25.61$ & $11.16 \pm 24.60$ & $2.85 \pm 2.12$ & $23.64 \pm 97.95$ \\

$\mathsf{TRPOLag}$-$\mathsf{F}$-$\mathsf{Pop}$
    & $89.15 \pm 29.45$ & $10.42 \pm 28.31$ & $2.76 \pm 2.37$ & $31.88 \pm 150.16$ \\

$\mathsf{TRPOLag}$-$\mathsf{A}$-$\mathsf{Pop}$
    & $89.40 \pm 24.14$ & $10.18 \pm 23.17$ & $2.93 \pm 2.55$ & $33.57 \pm 161.28$ \\

\bottomrule
\end{tabular}

\caption{
Schedule-independent trajectory-level safety evaluation for all online
RL policies. All metrics are computed over the common physical-time
horizon $[0,T]$, independently of the number and spacing of intervention
times. SafeTime\% denotes the fraction of physical time for which lactate
remains below $4.0$ mmol/L; ViolDur denotes the total duration of lactate
violation; PeakLac denotes the maximum lactate value over the trajectory;
and ExcessAUC denotes the time integral of lactate above the
$4.0$ mmol/L threshold. Results are reported as mean and standard
deviation across held-out patients.
The large standard deviations of ExcessAUC reflect a zero-inflated
distribution: the majority of patients maintain lactate below the
threshold throughout and thus contribute zero excess, while a small
number of patients with high initial lactate dominate the variance.
}
\label{tab:online_safety_evaluation}
\end{table}

Table~\ref{tab:online_safety_evaluation} reports these metrics for all
online policies. The Lagrangian individual policies are the safest
overall, whereas the Lagrangian cluster and population policies show no
consistent safety advantage over their unconstrained counterparts. This
indicates that treatment effectiveness and trajectory-level safety can
lead to different rankings of the same policies.

\subsection{Offline Policy Methods Evaluations}
\label{appendix:offline_policy_evaluations}
This section reports two sets of offline results that complement
Table~\ref{tab:offline_evaluation_dqn_cql_gcql} in the main text: the
$\mathsf{CQL}$ results under the primary grouping, and the results of all three
offline learners under the alternative patient grouping of
Appendix~\ref{appendix:alternative_clustering_evaluation}.
Following the offline learning protocol in
Appendix~\ref{appendix:offline_policy_learning}, we first generate fixed-dt
offline datasets from the Lagrangian TRPO behavior policy and then train DQN,
CQL, and GCQL policies without further simulator interaction during policy
optimization.
After training, all offline policies are deployed in the same MedGym
environments and evaluated by closed-loop rollouts.
This evaluation is intended to assess whether offline learning can recover
useful patient-specific treatment policies from fixed behavior data, and whether
the resulting individual policies improve over their corresponding
population-level policies.

\textbf{CQL under the primary grouping.}
$\mathsf{CQL}$ is reported here rather than in the main text because it behaves
differently from the other two offline learners. Its individual policy attains
the lowest trajectory-wide $\mathsf{SOFA}$ score of the three levels, but its
cluster policy falls behind the population policy rather than sitting between
the two. Since all three algorithms are trained on the same datasets and
evaluated on the same simulators, the difference lies in how each controls
extrapolation away from the behavior data.
\begin{table}[h]
\centering
\fontsize{8pt}{8pt}\selectfont
\setlength{\tabcolsep}{4pt}
\renewcommand{\arraystretch}{0.82}
\setlength{\aboverulesep}{0.3ex}
\setlength{\belowrulesep}{0.3ex}
\captionsetup{skip=2pt}
\begin{tabular}{lcccc}
\toprule
Policy & SOFA(all) ($\downarrow$) & Lower SOFA vs. Pop (\%) & SOFA(96h) ($\downarrow$) & SafeTime\% ($\uparrow$) \\
\midrule
$\mathsf{CQL}$-$\mathsf{F}$-$\mathsf{Ind}$ & $6.83 \pm 0.61$ & $50.0$ & $6.55 \pm 0.90$ & $98.95 \pm 2.99$ \\
$\mathsf{CQL}$-$\mathsf{F}$-$\mathsf{Clu}$ & $7.59 \pm 1.65$ & $30.0$ & $7.57 \pm 1.74$ & $87.34 \pm 30.17$ \\
$\mathsf{CQL}$-$\mathsf{F}$-$\mathsf{Pop}$ & $7.05 \pm 1.15$ & --     & $6.98 \pm 1.18$ & $84.14 \pm 32.87$ \\
\bottomrule
\end{tabular}
\caption{Offline-policy results for CQL under the primary grouping, evaluated on
the same $10$ held-out test patients and with the same metrics as
Table~\ref{tab:offline_evaluation_dqn_cql_gcql}.}
\label{tab:offline_evaluation_cql}
\end{table}

\textbf{Alternative patient grouping.}
Table~\ref{tab:offline_evaluation_dqn_cql_gcql_N15} repeats the offline
evaluation under the alternative grouping of
Appendix~\ref{appendix:alternative_clustering_evaluation}, in which each group
contains one training and one test patient. $\mathsf{GCQL}$ again shows the
clearest individual-over-population advantage, winning on $93.3\%$ of matched
patients, whereas for $\mathsf{DQN}$ the individual policy attains a slightly
lower mean score but wins on fewer than half of the patients.
\begin{table}[h]
\centering
\fontsize{8pt}{8pt}\selectfont
\setlength{\tabcolsep}{4pt}
\renewcommand{\arraystretch}{0.82}
\setlength{\aboverulesep}{0.3ex}
\setlength{\belowrulesep}{0.3ex}
\captionsetup{skip=2pt}
\begin{tabular}{lcccc}
\toprule
Policy & SOFA(all) ($\downarrow$) & Lower SOFA vs. Pop (\%) & SOFA(96h) ($\downarrow$) & SafeTime\% ($\uparrow$) \\
\midrule
$\mathsf{TRPOLag}$-$\mathsf{F}$-$\mathsf{Ind}$ (behavior) & $7.53 \pm 1.38$ & $93.3$ & $6.85 \pm 1.56$ & $92.13 \pm 20.99$ \\
\midrule
$\mathsf{DQN}$-$\mathsf{F}$-$\mathsf{Ind}$ & $9.75 \pm 1.54$ & $46.7$ & $9.72 \pm 1.82$ & $57.63 \pm 38.95$ \\
$\mathsf{DQN}$-$\mathsf{F}$-$\mathsf{Clu}$ & $9.01 \pm 1.98$ & $66.7$ & $9.10 \pm 2.27$ & $81.61 \pm 31.05$ \\
$\mathsf{DQN}$-$\mathsf{F}$-$\mathsf{Pop}$ & $9.80 \pm 1.09$ & --    & $9.77 \pm 1.13$ & $69.22 \pm 36.92$ \\
\midrule
$\mathsf{CQL}$-$\mathsf{F}$-$\mathsf{Ind}$ & $8.78 \pm 1.86$ & $66.7$ & $8.65 \pm 1.98$ & $86.60 \pm 26.70$ \\
$\mathsf{CQL}$-$\mathsf{F}$-$\mathsf{Clu}$ & $8.84 \pm 1.52$ & $53.3$ & $8.82 \pm 1.60$ & $80.37 \pm 36.15$ \\
$\mathsf{CQL}$-$\mathsf{F}$-$\mathsf{Pop}$ & $9.18 \pm 1.25$ & --    & $9.08 \pm 1.06$ & $80.02 \pm 32.80$ \\
\midrule
$\mathsf{GCQL}$-$\mathsf{F}$-$\mathsf{Ind}$ & $\mathbf{7.98 \pm 1.57}$ & $\mathbf{93.3}$ & $\mathbf{7.73 \pm 1.74}$ & $\mathbf{91.21 \pm 19.38}$ \\
$\mathsf{GCQL}$-$\mathsf{F}$-$\mathsf{Clu}$ & $8.36 \pm 1.55$ & $66.7$ & $8.37 \pm 1.61$ & $80.71 \pm 33.13$ \\
$\mathsf{GCQL}$-$\mathsf{F}$-$\mathsf{Pop}$ & $8.80 \pm 1.46$ & --    & $8.87 \pm 1.56$ & $75.38 \pm 39.61$ \\
\bottomrule
\end{tabular}
\caption{Offline-policy results for DQN, CQL, and GCQL under the alternative
grouping of Appendix~\ref{appendix:alternative_clustering_evaluation}, evaluated on the patient-specific simulators of the corresponding $15$ held-out test patients. Metrics and abbreviations follow Table~\ref{tab:policy_evaluation_selected}; dashes indicate that the population policy is not compared with itself.}
\label{tab:offline_evaluation_dqn_cql_gcql_N15}
\end{table}


To further examine why GCQL yields a clearer personalization pattern, we quantify the degree to which each evaluated offline policy departs from the behavior support. 
For each patient $i$, let $\mathcal{D}^{(i)}_{\mathrm{LagrangianTRPO}}$ denote the offline dataset generated by the TRPOLag behavior policy. 
We split it into a reference subset $\mathcal{D}^{(i)}_{\mathrm{ref}}$ and a held-out query subset $\mathcal{D}^{(i)}_{\mathrm{qry}}$. 
For a state-action pair $(\bm{\mathrm{x}},\bm{\mathrm{u})}$ generated during evaluation, we define its nearest-neighbor support distance as
\begin{equation}
d_{\mathrm{OOD}}^{(i)}(\bm{\mathrm{x}},\bm{\mathrm{u}})
=
\min_{(\bm{\mathrm{x}}',\bm{\mathrm{u}}') \in \mathcal{D}^{(i)}_{\mathrm{ref}}}
\left\|
\begin{bmatrix}
\bar{\bm{\mathrm{x}}} \\
\bar{\bm{\mathrm{u}}}
\end{bmatrix}
-
\begin{bmatrix}
\bar{\bm{\mathrm{x}}}' \\
\bar{\bm{\mathrm{u}}}'
\end{bmatrix}
\right\|_2 ,
\label{eq:offline_ood_distance}
\end{equation}
where $(\bar{\bm{\mathrm{x}}},\bar{\bm{\mathrm{u}}})$ denotes the coordinate-wise normalized state-action vector, with normalization statistics estimated from the offline behavior dataset.
For an evaluation dataset $\mathcal{D}^{(i)}_{\pi}$ generated by policy $\pi$, we plot the empirical distribution of $d_{\mathrm{OOD}}^{(i)}(\bm{\mathrm{x}},\bm{\mathrm{u}})$ over all state-action pairs $(\bm{\mathrm{x}},\bm{\mathrm{u}})\in \mathcal{D}^{(i)}_{\pi}$ and all patients. 
For the TRPOLag curve, we compute the same distance using the held-out query subset $\mathcal{D}^{(i)}_{\mathrm{qry}}$ rather than the reference subset itself, which avoids the trivial zero-distance comparison. 
Smaller values indicate that the evaluated policy remains closer to the support of the behavior data.

\begin{figure}[t]
\centering
\includegraphics[width=0.82\linewidth]{ 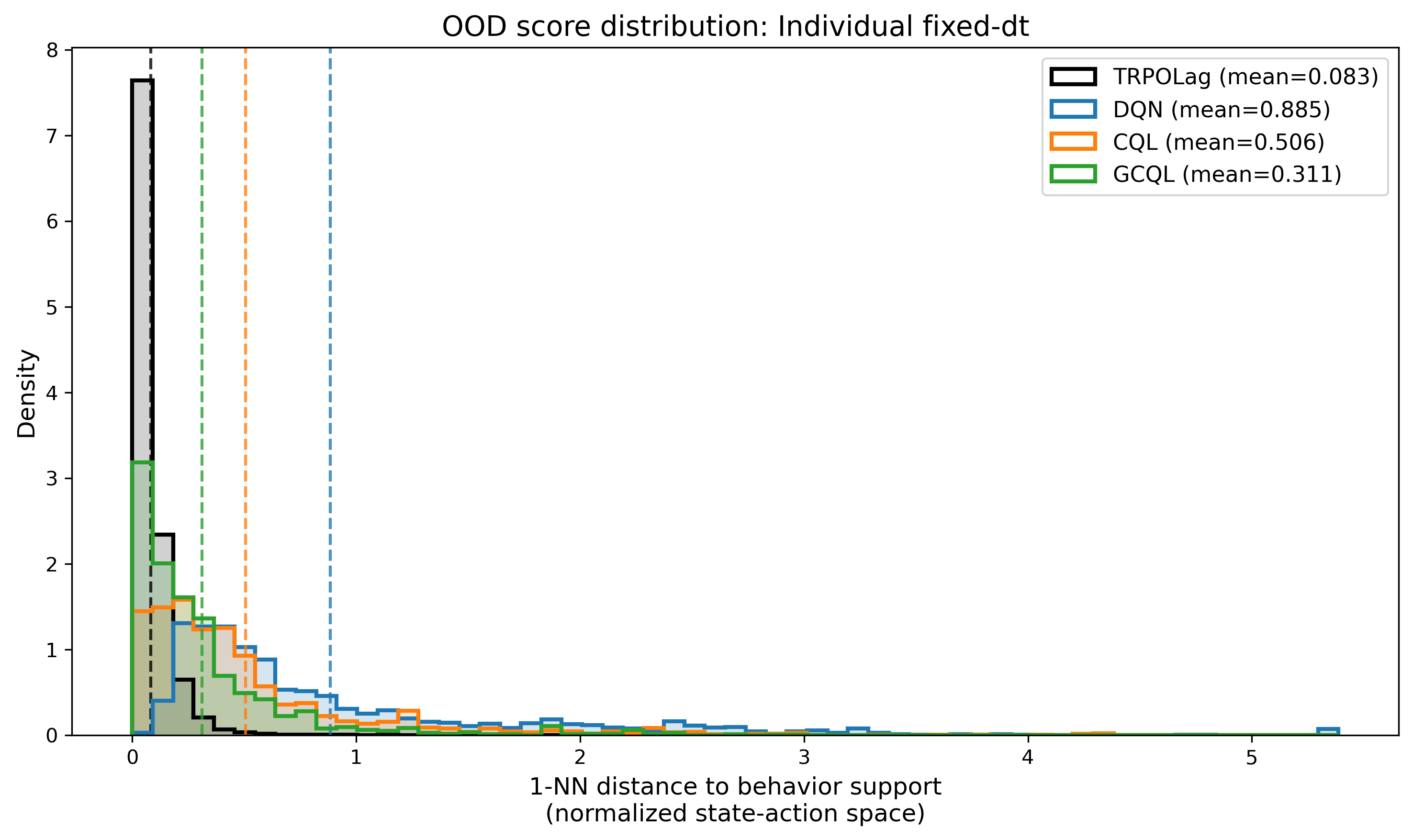}
\caption{
OOD support-distance distribution for fixed-dt individual offline policies.
For each evaluated state-action pair, we compute its 1-nearest-neighbor distance to the TRPOLag behavior dataset in normalized state-action space.
Smaller values indicate closer adherence to the behavior support.
GCQL stays closer to the behavior support than DQN and CQL, while DQN exhibits the largest OOD deviation.
}
\label{fig:offline_ood_distance}
\end{figure}

Figure~\ref{fig:offline_ood_distance} relates the preceding offline evaluation results to how far each method departs from the behavior-data support.
DQN has the largest mean support distance and a long right tail, indicating that it frequently visits state-action regions poorly supported by the behavior dataset. 
CQL reduces this deviation relative to DQN, but it still remains farther from the behavior support than GCQL. 
GCQL has the smallest OOD distance among the offline methods, suggesting that its guarded conservative objective better restricts the learned policy to supported state-action regions. 
This helps explain why GCQL is more stable in the personalization evaluation: by reducing OOD state-action visitation, GCQL avoids the unreliable value extrapolation that can occur when offline policies move outside the behavior-data support. 
Consequently, GCQL does not surpass the online TRPOLag behavior policy, but it narrows the offline performance gap relative to DQN and CQL and recovers a clearer individual-over-population advantage.

\section{Guidance for Researchers}
\label{appendix:guidance}

\textit{MedGym} is designed first and foremost as a benchmark for dynamic treatment recommendation under continuous-time patient evolution, irregular intervention timing, and patient heterogeneity. 
Accordingly, methods that claim relevance to medical treatment RL should be evaluated on \textit{MedGym} before drawing conclusions about their suitability for longitudinal healthcare settings. 
For fair comparison, researchers should begin with the default benchmark configuration and report results under the standard evaluation protocol, including treatment effectiveness, safety, and personalization-related metrics. 
The discount factor is configurable: setting $\gamma=1$ recovers the
finite-horizon continuous-time objective under the TACOS reformulation,
whereas $\gamma<1$ can be used to study discounted treatment objectives.

For offline RL, policies should be trained on datasets generated from the \textit{MedGym} environment or on the provided benchmark datasets, and then evaluated through online rollouts in the same environment. 
For online RL, methods should be trained directly in \textit{MedGym} until convergence and then assessed under the standard test settings. 
When applicable, researchers should compare both fixed-interval and adaptive-time variants, and should evaluate both population-level and individualized policies. 
In particular, methods that outperform the standard-of-care baseline or the population-level policy while maintaining strong safety performance provide evidence that they can exploit patient-specific and continuous-time structure rather than only fit average-case treatment patterns.

Although \textit{MedGym} is configurable, we caution against interpreting it as a disease-specific simulator for direct sim-to-real claims. 
The configurable parameters in \textit{MedGym} primarily define a family of medically grounded benchmark environments rather than an exact digital twin of a particular disease. 
Therefore, simply matching one dataset or one clinical cohort does not guarantee that conclusions drawn in \textit{MedGym} will transfer directly to real deployment. 
Researchers interested in a specific disease should treat \textit{MedGym} as a structured stress-test environment and, when necessary, complement it with additional domain-specific validation.

A more appropriate use of \textit{MedGym} in applied medical RL is to tune the benchmark configuration so that it reflects the qualitative properties of the target treatment scenario. 
Examples include the degree of patient heterogeneity, the irregularity of measurement and intervention times, the typical treatment horizon, the sparsity of observations, and the severity of safety constraints. 
Such choices allow researchers to ask practically meaningful questions, such as whether a method remains effective when interaction opportunities are limited, whether adaptive timing is beneficial, or whether individualized policies outperform population-level ones under realistic patient variation.

Finally, we recommend that results in \textit{MedGym} always be interpreted from multiple perspectives rather than from reward alone. 
At a minimum, researchers should report treatment effectiveness, trajectory-level safety, and the relative performance of individualized versus population-level policies. 
Whenever possible, evaluations should also include multiple patient models or environment instantiations in order to assess robustness and generalizability. 
In this way, \textit{MedGym} can serve as a standardized yet flexible benchmark for identifying which RL methods are most promising for future dynamic treatment applications, while avoiding overly strong conclusions about direct clinical deployment.


\section{Broader Impact}
\label{appendix:broader_impact}

\textit{MedGym} is developed to support more realistic and informative evaluation of reinforcement learning methods for dynamic medical treatment. 
A potential positive impact of this work is that it provides a benchmark in which researchers can study clinically important properties that are often simplified or ignored in existing settings, including continuous-time disease progression, irregular intervention timing, patient heterogeneity, and trajectory-level safety. 
By enabling more faithful pre-deployment evaluation, \textit{MedGym} may help the community identify which classes of methods are better suited to safety-critical treatment recommendation and may reduce over-optimistic conclusions drawn from overly simplified benchmarks. 
More broadly, the benchmark may encourage the development of learning algorithms that better reflect the temporal and personalized nature of real clinical decision-making.

The benchmark may also have practical value for comparing offline and online policy learning methods under a common simulation framework. 
In healthcare, direct online experimentation is often costly, ethically constrained, or infeasible. 
A configurable benchmark such as \textit{MedGym} can therefore provide a useful intermediate step for stress-testing algorithms before any consideration of real-world translation. 
In addition, the emphasis on individualized dynamics may motivate future work on patient-specific modeling and treatment design, which could ultimately contribute to more personalized and data-efficient decision support tools.

At the same time, this work has several potential negative societal implications if misused or over-interpreted. 
First, \textit{MedGym} is still a learned simulator rather than a validated clinical deployment system. 
If benchmark results were mistakenly interpreted as direct evidence of clinical effectiveness, this could encourage premature or unsafe real-world use of RL-based treatment policies. 
Second, because the benchmark is derived from historical clinical data, any biases, omissions, or cohort-specific patterns present in the source data may be inherited by the learned environment. 
As a result, methods that perform well in \textit{MedGym} may still behave unevenly across patient populations that are underrepresented or insufficiently modeled in the data. 
Third, like other benchmark environments, \textit{MedGym} may incentivize optimization toward benchmark-specific metrics at the expense of broader clinical validity, interpretability, or robustness.

There is also a risk that personalized modeling could be misunderstood as fully solving the personalization problem in medicine. 
In the current version of \textit{MedGym}, patient-specificity is operationalized through individualized models rather than through a clinically validated latent representation of patient factors. 
Accordingly, good benchmark performance should not be interpreted as evidence that a method has discovered causal patient structure or that it will generalize safely across hospitals, cohorts, or diseases. 
In addition, although the benchmark includes safety-oriented evaluation, it does not eliminate the broader ethical challenges associated with deploying automated decision systems in healthcare, including accountability, transparency, fairness, and clinician oversight.

To mitigate these risks, \textit{MedGym} should be used as a research benchmark rather than as a stand-alone basis for clinical decision-making. 
Results obtained in this environment should be interpreted together with external validation, domain expertise, and careful analysis of failure modes, subgroup behavior, and safety trade-offs. 
We hope that, when used appropriately, \textit{MedGym} will have a net positive societal impact by improving the rigor with which medical RL methods are evaluated before any future translational use.

\end{document}